\documentclass[sn-mathphys]{sn-jnl}

\usepackage{caption}
\usepackage{subcaption}
\usepackage{multirow}
\usepackage{siunitx}

\jyear{2022}%

\begin{document}

\title[Vision and Tactile Robotic System to Grasp Litter in Outdoor Environments]{Vision and Tactile Robotic System to Grasp Litter in Outdoor Environments}


\author*[1,2]{\fnm{Ignacio} \sur{de Loyola Páez-Ubieta}}\email{ignacio.paez@ua.es}

\author*[1]{\fnm{Julio} \sur{Castaño-Amorós}}\email{julio.ca@ua.es}

\author[1]{\fnm{Santiago} \sur{T. Puente}}\email{santiago.puente@ua.es}

\author[1]{\fnm{Pablo} \sur{Gil}}\email{pablo.gil@ua.es}

\affil[1]{\orgdiv{AUtomatics, RObotics and Artificial Vision Lab}, \orgname{University of Alicante}, \state{Alicante}, \country{Spain}}

\affil[2]{\orgdiv{University Institute for Engineering Research}, \orgname{Miguel Hernandez University}, \city{Elche}, \country{Spain}}


\abstract{The accumulation of litter is increasing in many places and is consequently becoming a problem that must be dealt with. In this paper, we present a 
manipulator robotic system to collect litter in outdoor environments. This system has three functionalities.
Firstly, it uses colour images to detect and recognise litter comprising different materials.
Secondly, depth data are combined with pixels of waste objects to compute a 3D location and segment three-dimensional point clouds of the litter items in the scene. The grasp in 3 Degrees of Freedom (DoFs) is then estimated for a robot arm with a gripper for the segmented cloud of each instance of waste. Finally, two tactile-based algorithms are implemented and then employed in order to provide the gripper with a sense of touch. This work uses two low-cost visual-based tactile sensors at the fingertips. One of them addresses the detection of contact (which is obtained from tactile images) between the gripper and solid waste, while another has been designed to detect slippage in order to prevent the objects grasped from falling. Our proposal was successfully tested by carrying out extensive experimentation with different objects varying in size, texture, geometry and materials in different outdoor environments (a tiled pavement, a surface of stone/soil, and grass). Our system achieved an average score of 94\% for the detection and Collection Success Rate (CSR) as regards its overall performance, and of 80\% for the collection of items of litter at the first attempt.}




\maketitle

\section{Introduction}
\label{sec:intro}

Several environmental problems currently harm our planet, one of which is the accumulation of waste such as plastic bottles, metal cans, drink cardboard or glass that is clearly visible in the streets and parks of cities. The mean estimated degradation time for cardboard and glass ranges from 5 to 4,000 years, respectively. In order to help avoid the contamination of soil and the environment, this waste should be selectively picked up in an automated manner for its subsequent recycling. We propose to solve the problem of collecting this kind of waste outdoors by providing a 
robotic system that incorporates several visual-tactile perception systems. 

Several solutions that use robots for cleaning purposes already exist in literature. For instance, \cite{chiang2015vision} solve the aforementioned problem in a simulated indoor environment, and \cite{muthugala2020tradeoff} presents a re-configurable cleaning robot that works in a real environment, but these solutions do not provide object recognition or manipulation skills that enable refuse to be picked. In this line, \cite{zapata2018autotrans} shows an outdoor solution in which a robot detects and picks up refuse bags, but without tactile perception or variability in the scenarios. At present, robot learning techniques make it possible to address interactions with the environment during navigation and manipulation tasks, as shown in \cite{pmlr-v164-sun22a}. Although its results are promising, this kind of approach requires a lot of training data and is usually limited to controlled settings, which are usually indoors.  

The technique presented herein has been tested in real-world scenarios by performing not only detection, as occurs in \cite{sultana2020trash}, but also the recognition of a wider number of waste elements than occurs in \citep{lin2015robot}. This is done by extracting the best grasping points of the segmented point cloud, signifying that there are fewer points to process, thus lightening the computation time when compared to that of \cite{article}. Unlike the aforementioned works, our approach also includes object handling skills thanks to new tactile perception algorithms, which make it possible to accomplish stable grasping when picking up litter. 


Our main contributions are:
\begin{itemize}
    \item We propose a tactile-based grasping estimation method for the picking of waste objects. We specifically use low-cost visual-based sensors for tactile manipulation in order to carry out the task of litter collection. As these tactile sensors do not have a mathematical equation with which to map tactile images (intensity) onto force in N. and they also do not contain visual markers to estimate the movement variation, then our data-based methods are, therefore, crucial as regards performing this task correctly.
    \item We generate two datasets, one containing tactile images for the contact and slip detection tasks, and another containing color images of household waste in a wide variety of outdoor environments, for the object detection and localisation tasks.
    \item We present comparative studies of object recognition and contact-slip detection during the grasping task, first using our datasets and later grasping litter in outdoor environments such as in our university campus. Therefore, another contribution is based on the design, implementation and communication of the different perception modules on a real robot system applied to the task of litter collection in outdoor environments.
\end{itemize}

\begin{figure}[!htb]
     \centering
     \begin{subfigure}[b]{0.45\textwidth}
         \centering
         \includegraphics[width=\textwidth, height=4.3cm]{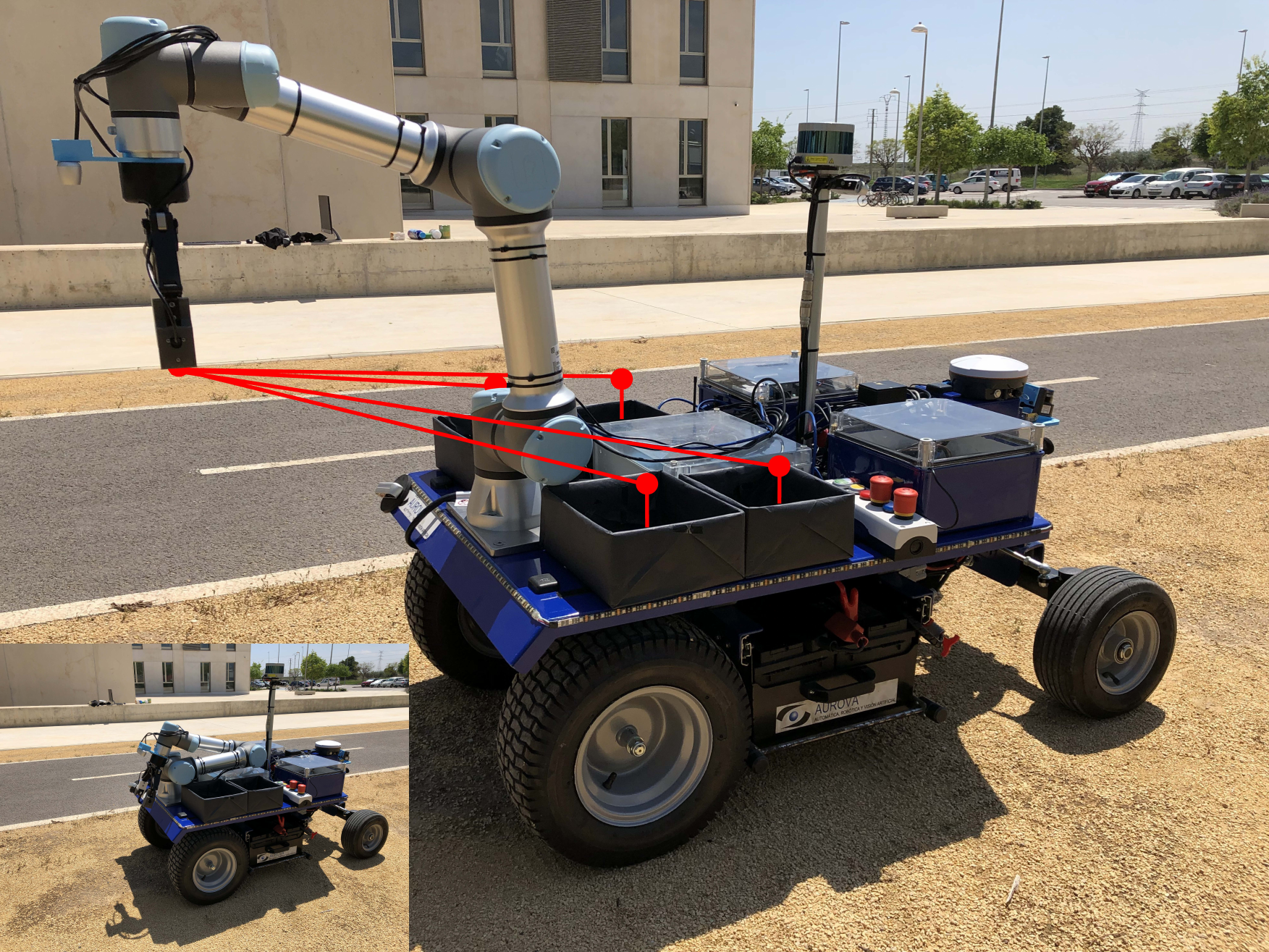}
     \end{subfigure}
     \begin{subfigure}[b]{0.45\textwidth}
         \centering
         \includegraphics[width=\textwidth, height=4.3cm]{FIG1-RIGHT.pdf}
     \end{subfigure}
     \caption{Home, navigation and detection poses of BLUE robot. (Left) UR5e pose when BLUE is in home pose. (Bottom-left) Navigation pose when BLUE is moving around. (Right) Detection pose when BLUE is near an item of litter}
     \label{fig:blue_poses}
\end{figure}

Our robotic system is divided into two main modules: the vision module and the tactile module. The vision module is able to detect and recognise litter from images and to calculate grasping points from the point cloud of the object. The tactile module performs grasping detection and control on the basis of tactile feedback. Our solution has been integrated and tested using an UR5e commercial robotic arm with a 2F-140 ROBOTIQ gripper. The arm was installed on our mobile robotic platform with autonomous navigation, called BLUE \citep{delPino2019}, and can be seen in Fig. \ref{fig:blue_poses}. BLUE has several sensors to which we have added IMU and RGBD cameras ({Intel\textregistered} {RealSense\texttrademark} D435i) for detection and recognition, along with two DIGIT sensors \citep{Lambeta2020} attached to the fingertips of the gripper for the tactile operations.

This paper is organised as follows: Section \ref{sec:works} provides an explanation of related works regarding each part of the pipeline., while Section \ref{sec:system} shows a description of the proposed methods of which our visual-tactile perception system is composed. A description of how each of the methods is trained and validated separately is provided in Section \ref{sec:expe}, along with the preliminary results obtained after carrying out tests with previously unseen items of waste. The results obtained by the whole visual-tactile perception system in real environments are then presented, and finally, the paper concludes with a discussion of the results obtained and the performance of the proposal.


\section{Related works}
\label{sec:works}

\subsection{Visual perception}
\label{sec:bdetection}
The rapid development of deep learning has led to an avalanche of object detection methods \citep{chandra2021review}. For example, \cite{sultana2020trash} used an AlexNet Convolutional Neural Network (CNN) to perform image classification tasks in both indoor and outdoor environments. However, this work presents solely a detection system for household waste rather than a system with which to pick waste up.

Other works attempt to solve more complex problems, such as object detection and segmentation. In \cite{bai2018deep}, the author uses two Neural Networks (NNs) to eliminate the ground and classify six kinds of waste items. We, however, use a single NN to locate and classify the objects, thus speeding up the process. Going one step further,  \cite{liu2021garbage} trained a YOLACT NN to classify items of domestic waste in indoor environments. They tested the model with only three categories (``plastic", ``metal" and ``cardboard"). We, in contrast, add a new class of glass objects, thus making the detection more difficult owing to their transparency. Moreover, our system works in outdoor environments in which lighting conditions cannot be controlled. 

The waste collection task makes it necessary to address not only its detection and location but also the grasping of objects \citep{mnyussiwalla2022evaluation}, as already implemented in the robotic field. For instance, \cite{article} proposed a mathematical method with which to calculate the best pair of grasping points from a 3D-scene point cloud. This method extracts the object and obtains the grasping pose using the curvature of the object. In our work, we use a segmented point cloud of the 3D-object for the grasping task. Reducing the number of candidate points makes the process faster. Another NN approach, which is shown in \citep{ten2017grasp}, generates grasping proposals from voxelised 3D-object point clouds. But it has some limitations as regards complete outdoor point clouds rather than segmented object point clouds and working in real time. Our proposal is an improvement in this respect, since it is faster in outdoor scenarios.

\subsection{Tactile perception}
\label{sec:btactileandgraspingcontrol}

Several types of tactile sensors (capacitive, resistive, barometric, optical-based, etc.) have been developed in recent years in order to assist in robotic manipulation tasks such as contact or slip detection, which are essential for a safe grasp when picking up objects. In previous works, \cite{dong2017improved} calculated contact by comparing the intensity of the colour of two tactile images using an optical tactile sensor called Gelsight \citep{yuan2017gelsight}. However, this method requires the readjustment of its parameters and is less robust to uncertainty owing to the use of traditional computer vision techniques. In \cite{pmlr-v100-zhang20b}, the authors trained a Recurrent Neural Network (RNN) to detect contact events in image sequences from another optical tactile sensor called FingerVision \citep{zhang2018fingervision}. Nonetheless, this sensor contains markers that are complex to generate and require high-cost machinery. The DIGIT sensors do not contain markers, while our contact detection CNN-based method is more robust than traditional computer vision techniques.

When employed in the literature related to this field, the term slip usually refers to the normal component of the grasping force going outside of the grasping cone. Detecting and reacting to this phenomenon is, therefore, fundamental if an object is to be grasped correctly. In \citep{james2018slip} a Support Vector Machine (SVM) algorithm is trained in order to classify whether images obtained from a TacTip sensor are stable or slipping events. However, they do not grasp the object with a gripper or hand, but rather push the object against the wall in order to stop the slip movement. This application is, therefore, limited. The slip or stable classification task has also been studied in \cite{li2018slip}, in which visual (eye-to-hand camera) and tactile (Gelsight sensor) information is combined. However, it is necessary to process a lot of data, and real-time execution is not guaranteed. Conversely, our slip detector works extremely fast and allows the implementation of our controller. In a major advance, \cite{sliptmo} researched the slip detection problem using a multi-fingered hand. Nonetheless, detecting slip is more complex as the number of sensors increase. Our approach of using a two-fingered robotic gripper is, therefore, more optimal for the litter collection task.


\section{Our approach: Visual-tactile perception for 
robotic manipulation}
\label{sec:system}

\begin{figure}[htpb]
    \centering
    \includegraphics[width=1\textwidth, height=13cm]{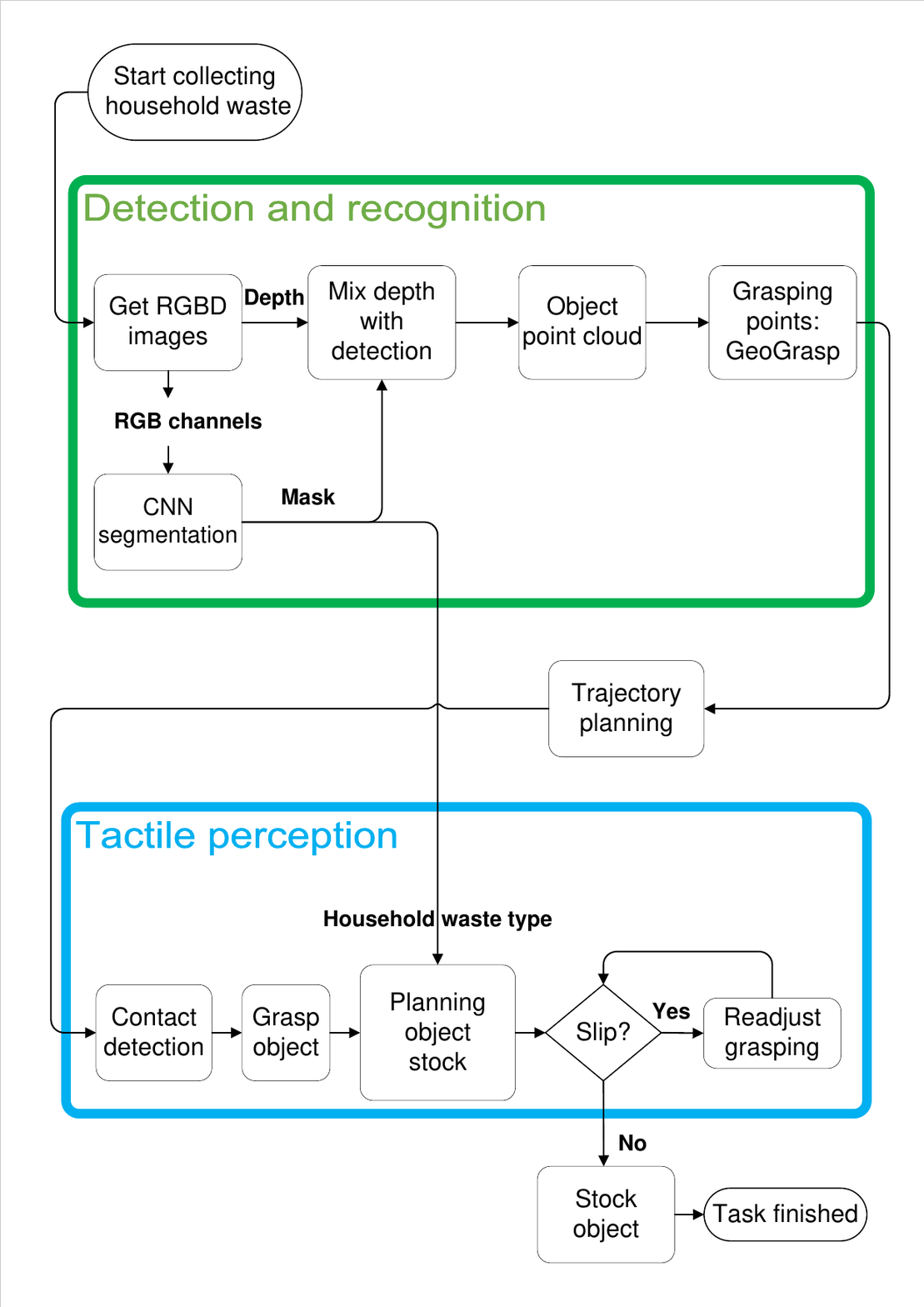}
    \caption{Scheme of our tactile-visual system for robotic grasping. It is made up of two main parts: waste detection and recognition (green part) and tactile perception in order to manipulate the item of waste  (blue part).}
    \label{fig:flow_chart}
\end{figure}

In this work, we propose a visual-tactile perception system for robotic manipulation, whose architecture is shown in Fig.  \ref{fig:flow_chart}. Only the six DoFs manipulation arm of the mobile manipulator robotic system described in Section \ref{sec:intro} is employed for the task of litter collection, assuming that the navigation task towards the litter objects is already carried out, thus obtaining an optimal localization of the mobile platform. Detailed descriptions of the three main modules of our system are provided in the following sections.

\subsection{Object detection and localisation}
\label{sec:detection}
In this section, the first module is explained, which consists of detecting and locating the item of litter using a CNN.

The first task carried out by our system is detecting and classifying litter in outdoor environments. In order to accomplish it, our BLUE robot performs the exploration task using several navigation algorithms. While the exploration is taking place, BLUE is able to annotate objects as possible instances of refuse and compute their spatial location in the world \citep{Tornero2022}, thus making it possible to plan trajectories towards them. The visual sub-system, which is shown in the green scheme in Fig. \ref{fig:flow_chart}, captures images while the robot is navigating and then processes them using a CNN to obtain the object position and category of the objects. After analysing several CNNs available in the state of the art \cite{hafiz2020survey}, \cite{gu2022review}, we chose Mask R-CNN \citep{He_2017_ICCV}, YOLACT \citep{yolact-iccv2019} and YOLACT++ \citep{yolact-plus-tpami2020}. These CNNs appear to work well in a wide variety of fields that require segmentation tasks, such as understanding natural scenes or intelligent driving. In the former, it improves object detection since it avoids some cases of occlusion by providing a more detailed analysis of the image, while in the latter it is used to determine the localisation of major categories of objects such as street lights or people running.

\begin{figure}[!htb]
     \centering
     \includegraphics[width=0.98\textwidth, height=4.3cm]{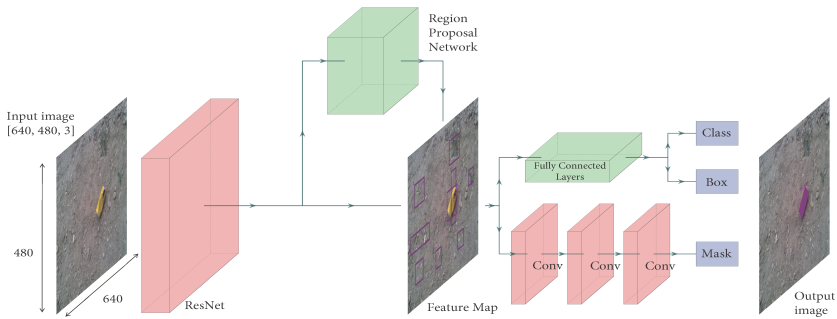}
     \caption{Mask R-CNN architecture with an example of the input and output from the litter detection task}
     \label{fig:maskrcnn_scheme_image}
\end{figure}

On the one hand, Mask R-CNN is considered to be a two-stage detector, since it has two parts. The first generates Regions of Interest (RoIs), and the second classifies and segments those RoIs. These detectors have drawbacks, such as low performance and a dependency on feature localisation. Its architecture is similar to that of the Faster-RCNN \citep{DBLP:journals/corr/RenHG015}, since it adds a new layer to the Faster-RCNN in order to predict a segmented mask. In this case, there are three output layers: the class label, the Bounding Box (BB), and the aforementioned segmented mask. These changes comprise an improved RoI Polling layer (RoI Align) and the addition of a segmented mask output layer in Mask R-CNN (see Fig. \ref{fig:maskrcnn_scheme_image}). 

\begin{figure}[!htb]
     \centering
     \includegraphics[width=0.95\textwidth, height=4.5cm]{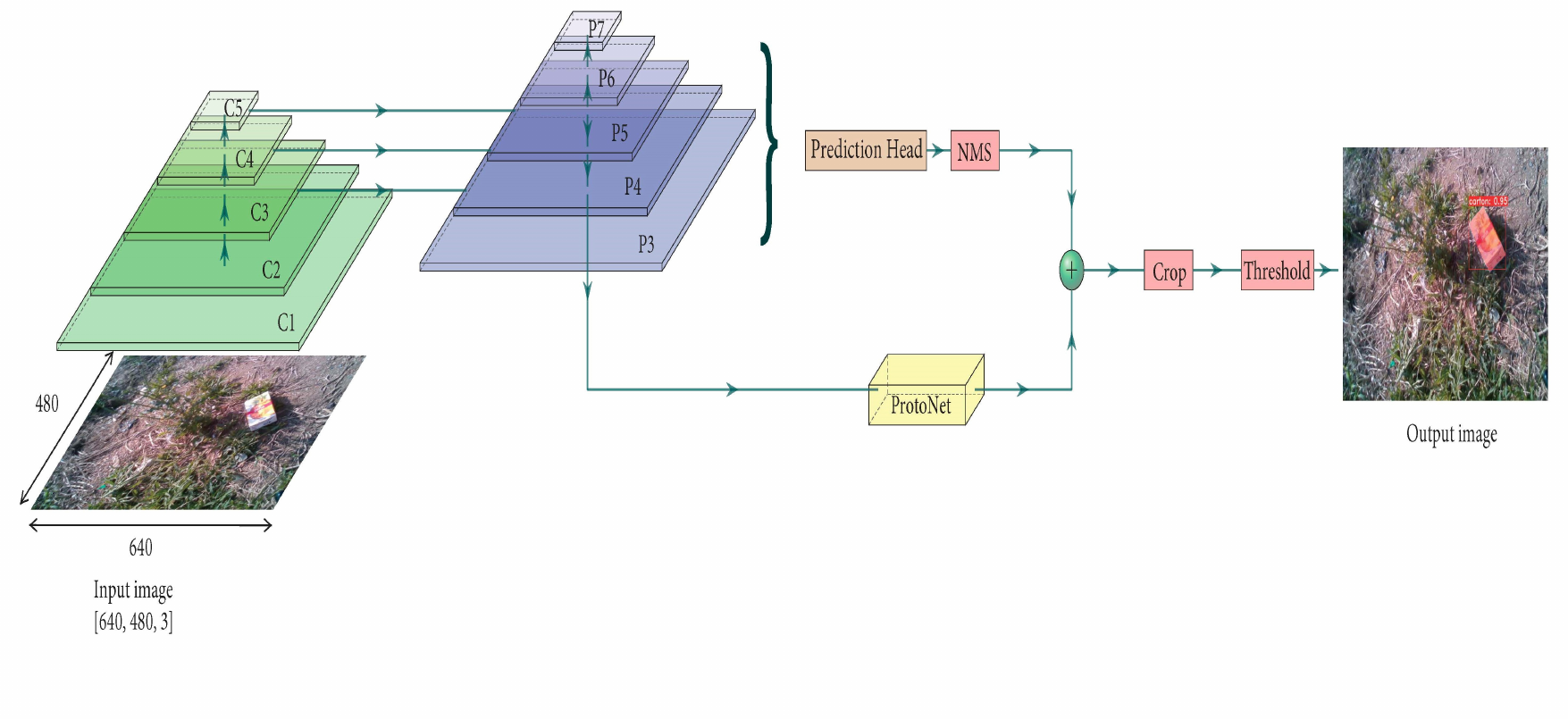}
     \caption{YOLACT architecture with an example of the input and output from the litter detection task}
     \label{fig:yolact}
\end{figure}

On the other hand, YOLACT is considered to be a single-stage detector that performs the instance segmentation in one step. This makes it really fast and allows it to achieve real-time inference. The detection task  is divided  into two simpler parallel sub-tasks. The first consists of the prototype generation branch, which predicts a $k$-set of prototype masks without loss (this loss exists in the traditional methods). The second involves the mask coefficient branch, which is a vector of mask coefficients for each prototype. These result in $k$-mask coefficients (one per prototype), $c$ class confidences, and $4$ BB regressors, producing a total of $4+c+k$ coefficients per anchor. Finally, the two sub-tasks come together in the mask assembly process. This is done by applying a linear combination to both sub-tasks and a sigmoid non-linearity process (see Fig. \ref{fig:yolact}). YOLACT++ was created later by making some minor changes to YOLACT. These changes range from a fast mask re-scoring network to deformable convolution in layers, including an optimised prediction head. 

One of the configurable parameters in both CNNs is the backbone. Our visual sub-system is implemented in order to choose between the ResNet50-FPN and the ResNet101-FPN. Mask R-CNN includes the DarkNet53-FPN as an extra backbone. The number that accompanies each backbone determines the number of layers. In fact, all of these CNNs include a Feature Pyramid Network (FPN) as part of their architecture.  

ResNet \citep{He_2016_CVPR} stands for Residual Neural Network. In 2015, it was still believed that adding additional layers to a NN would make it work better. This worked in theory but not in practice since there was the problem of the vanishing gradient. This new architecture solved that problem by incorporating residual blocks with skip connections.

The FPN consists of reducing the size of an image step by step. Image features are then extracted from both the original and the scaled images in each step. FPN later combines all the features that have been extracted, mixing both low-resolution semantically strong and high-resolution semantically weak features. This combination can be achieved by using top-down and lateral connections, thus leading to better results in outdoor image analysis.

Another backbone, which is written in C and CUDA, is DarkNet \citep{darknet13}. This rose to fame because it improved the performance of ResNet101-FPN and carried out the detection process 1.5x faster. 

In this work, we shall analyse the behaviour and performance of these architectures as backbones in detection tasks for robotic manipulation in real outdoor environments. We first present the validation results, which were obtained after running our system offline. That is, we used pre-recorded videos of previous navigation missions that had already been carried out (see Section \ref{sec:visual_validation}). Finally, we show additional results in new scenarios. These have not been seen by our NN before and were captured in real-time navigation mode (see Section \ref{sec:outdoor_results}). 

\subsection{Grasping computing and trajectory planning}
\label{sec:grasping}

This section describes how our system estimates the grasping points from the 3D point cloud to collect the item of litter.

Once the object has been recognised and is in the robot's reachable workspace, our 
manipulator robot has to pick it up. It is, therefore, necessary to estimate the grasp. In our case, we obtain grasping points by using a method called GeoGrasp. It is based on geometry and needs a raw scene point cloud as input. But in this work, the input has been changed, as inspired by \citep{de2021domestic}. A filtered point cloud is, therefore, employed. 

An image is initially captured using a {RealSense\texttrademark} D435i depth camera. The depth image has a resolution of 640x480 pixels, which matches that of the RGB image. The RGB image is then processed by Mask R-CNN, YOLACT or YOLACT++. The detection task results in a cluster of pixels containing the object, a BB, and a category. 
This result allows the creation of a new point cloud that includes only the RGBD points considered by the NN as object points. Then, GeoGrasp calculates the grasping points in the new point cloud and locates these points in the 3D space referenced at the base of the robot. Two values are required for this task: the transformation from the proposed grasping points to the camera and the transformation from the camera to the robot base. 

The first transformation is calculated using the proposed grasping points and the camera intrinsic parameters as in Eq. (\ref{eqn:pointcloud_coord}). 

\begin{equation}
\begin{bmatrix}
x_{\mathrm{C}} \\ y_{\mathrm{C}} \\ z_{\mathrm{C}}
\end{bmatrix}
=
\begin{bmatrix}
R^{\mathrm{t}}
\end{bmatrix}
\begin{bmatrix}
M_{\mathrm{c}}
\end{bmatrix}
^{\mathrm{-1}}
\begin{bmatrix}
x_{\mathrm{G}} \\ y_{\mathrm{G}} \\ 1
\end{bmatrix}
d
=
\underset{\mathbf{R^{\mathrm{t}}}}{\underbrace{\begin{bmatrix}
\mathbf{0}&\mathbf{0}&\mathbf{1} \\ \mathbf{1}&\mathbf{0}&\mathbf{0} \\ \mathbf{0}&\mathbf{1}&\mathbf{0}
\end{bmatrix}}}
\underset{\begin{bmatrix}
M_{\mathrm{c}}
\end{bmatrix}
^{\mathrm{-1}}}{\underbrace{\begin{bmatrix}
\mathbf{1/f_{\mathrm{x}}}&\mathbf{0}&\mathbf{-c_{\mathrm{x}}/f_{\mathrm{x}}} \\ \mathbf{0}&\mathbf{1/f_{\mathrm{y}}}&\mathbf{-c_{\mathrm{y}}/f_{\mathrm{y}}} \\ \mathbf{0}&\mathbf{0}&\mathbf{1}
\end{bmatrix}}}
\begin{bmatrix}
x_{\mathrm{G}} \\ y_{\mathrm{G}} \\ 1
\end{bmatrix}
d
\in \mathbb{N}_{\geq 0}
\label{eqn:pointcloud_coord}
\end{equation}

where [$x_{\mathrm{G}}$, $y_{\mathrm{G}}$] denote the coordinates of the proposed grasping point in a 2D image for the $x$ and $y$ axes respectively, ($f_{\mathrm{x}}$, $f_{\mathrm{y}}$) and ($c_{\mathrm{x}}$, $c_{\mathrm{y}}$) respectively denote the focal length and the camera center (obtained from the intrinsic parameters of the camera) in pixels, $R^{\mathrm{t}}$ denotes a consecutive rotation of -90 degrees in the x and y axes respectively, while $M_{\mathrm{c}}$ denotes the calibration matrix (obtained by using ArUco markers \citep{garrido2014automatic}) and $d$ denotes the depth of the point in mm (obtained from the D channel of the RGBD image captured).

The second transformation consists of transforming [$x_{\mathrm{C}}$, $y_{\mathrm{C}}$, $z_{\mathrm{C}}$] into the coordinates of the base of the robot by following Eq. (\ref{eqn:base_coord}).


\begin{equation}
\begin{bmatrix}
x_{\mathrm{B}} \\ y_{\mathrm{B}} \\ z_{\mathrm{B}} \\ 1
\end{bmatrix}
=
\begin{bmatrix}
\mathbf{R(q)}&\mathbf{p(q)} \\ \mathbf{0^{\mathrm{T}}}&1
\end{bmatrix}
\begin{bmatrix}
^{\mathrm{E}} T_{\mathrm{C}}
\end{bmatrix}
\begin{bmatrix}
x_{\mathrm{C}} \\ y_{\mathrm{C}} \\ z_{\mathrm{C}} \\ 1
\end{bmatrix}
\label{eqn:base_coord}
\end{equation}

where [$x_{\mathrm{C}}$, $y_{\mathrm{C}}$, $z_{\mathrm{C}}$] denote the 3D coordinates relative to the camera, [$x_{\mathrm{B}}$, $y_{\mathrm{B}}$, $z_{\mathrm{B}}$] denote the new coordinates relative to the base in meters, $\mathbf{q} \in \Re^{\mathrm{6}}$ denotes a vector of joint coordinates, $\mathbf{p(q)} \in \Re^{\mathrm{3}}$ denotes the position vector from the robot´s base to the coordinate frame (forward kinematics), $\mathbf{R(q)} \in SO(3)$ is the end-effector orientation and $^{\mathrm{E}} T_{\mathrm{C}}$ is the fixed transformation between the camera and the robot end-effector. These coordinate systems can be seen on the right-hand side of Fig. \ref{fig:blue_poses}.

\begin{figure}[!htb]
     \centering
     \includegraphics[width=0.98\textwidth, height=5.5cm]{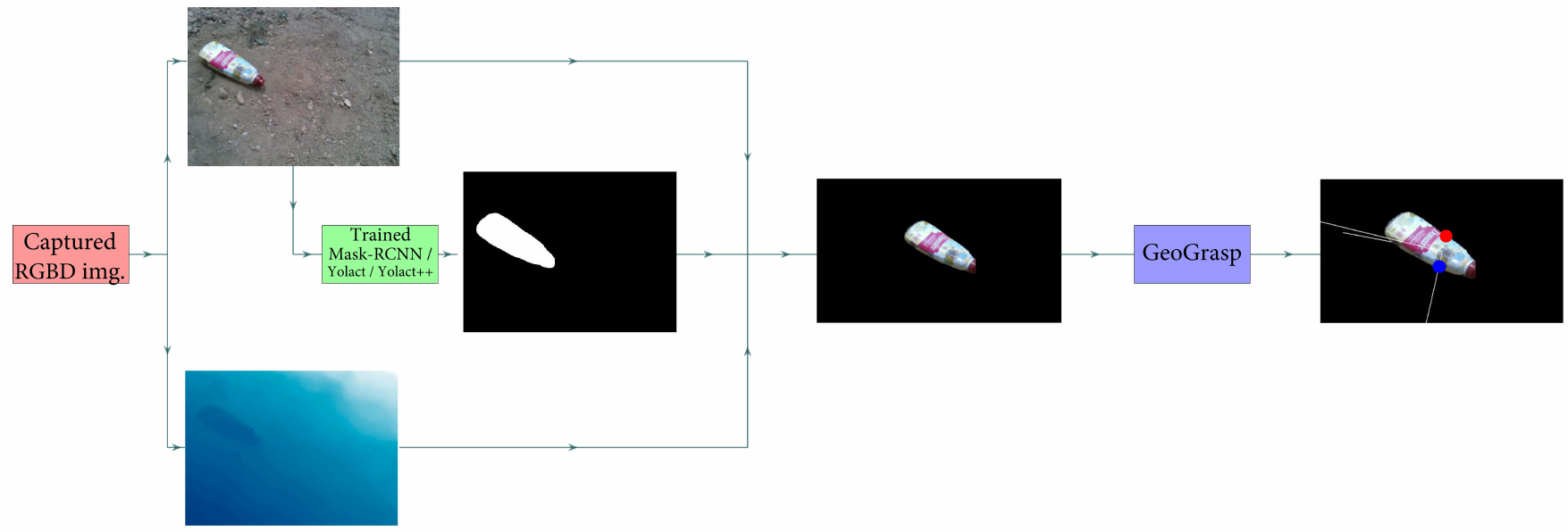}
     \caption{Scheme of our object detection and grasping points calculation process. Our NNs obtain the segmented mask from the RGB image, which is used to extract the object point cloud in order to calculate the grasping points with our new version of the GeoGrasp algorithm}
     \label{fig:new_geograsp}
\end{figure}

One drawback of using only the segmented object points is that we have to trust the NN detection since our calculation is based on its result. The worse the NN detection is, the worse the grasping points will be positioned. The full process is shown in Fig. \ref{fig:new_geograsp}.

Once the objects have been detected and localised, and their grasping estimated, the robotic arm mounted on the BLUE robot has to reach them. This is done using ROS Moveit! \citep{coleman_sucan_chitta_correll_2014}.
Moveit! provides the UR5e arm with a motion planning framework in order to compute and test the trajectories before operating the real physical robot. A wide variety of motion planning controllers are provided, but we use RRT* \citep{lavalle1998rapidly}, \citep{lavalle2001randomized} (an asymptotically optimal version of Rapidly-exploring Random Trees).  

It is worth mentioning that some of the trajectories computed will always be the same. This has been borne in mind, and some of them  have, therefore, been prerecorded. These are from the navigation pose (see Fig. \ref{fig:blue_poses}-bottom-left) to the home pose (see Fig. \ref{fig:blue_poses}-left), the home pose to the detection pose (see Fig. \ref{fig:blue_poses}-right) and the home pose to each of the container poses, and vice versa. 
In necessary, we simply need to play them. That will allow the robot to save computing time and avoid undesired or dangerous movements.
The aforementioned trajectories are shown in Fig. \ref{fig:blue_poses}-left. The remaining movements are planned in situ (blue paths - see Fig. \ref{fig:blue_poses}-right). The green working area has dimensions of 600x500 mm. All objects in this area are easily graspable and easy for the camera to see.

Four containers are attached to the base of BLUE. This is the same as the number of categories used to classify objects in Section \ref{sec:detection}. Each class of the dataset will, therefore, be stored in different containers.

Finally, all positions have pre-positions. These are used to avoid pushing, moving, or colliding with the objects. There is a downward movement of $\approx 100$ mm from the pre-position to the final position. This value was obtained empirically from the experiments and was that which obtained the best results.

\subsection{Tactile manipulation}
\label{sec:tactile}

This module consists of detecting physical contact and slippage between the items of litter and the gripper during the manipulation task. This task involves picking up the item of litter from the ground and placing it in the desired bin, given the previously calculated trajectory and grasping points. In order to ensure grasping safety, the contact is detected before the robot lifts the object. Once the contact has been guaranteed, the robot will be able to adjust the grasping opening if a slip is detected. These operations cannot be carried out without tactile feedback in real time. We, therefore, implement a closed-loop controller for each task as shown in the blue scheme in Fig. \ref{fig:flow_chart}. 

It is known in the literature that force sensors are suitable for this kind of task because they allow the implementation of force-feedback controllers. Nonetheless, in this paper, the objective is to demonstrate that grasping and slipping controllers can be implemented in order to successfully solve the task by using optical tactile sensors that do not provide force values, which are known as DIGITs. These tactile sensors are more economical, smaller and provide more information about the features of the object such as texture or shape.

\begin{figure}[ht]
     \centering
     \begin{subfigure}[b]{0.30\textwidth}
         \includegraphics[width=\linewidth, height=2.5cm]{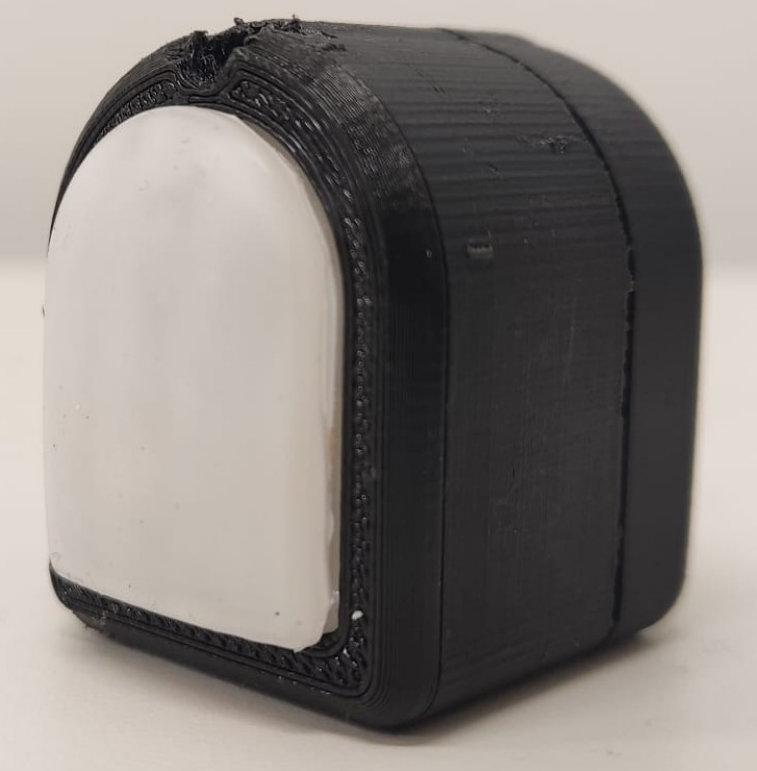}
    \caption{}
     \end{subfigure}
     \begin{subfigure}[b]{0.30\textwidth}
         \centering
         \includegraphics[width=\textwidth, height=2.5cm]{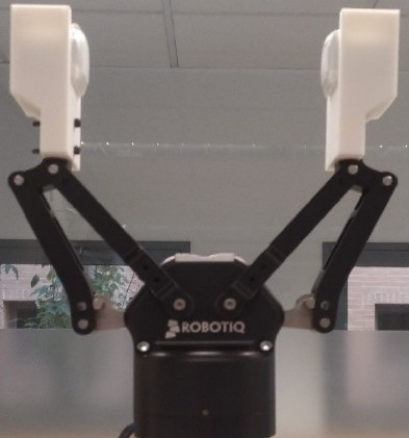}
         \caption{}
     \end{subfigure}
     
       \caption{(a) DIGIT sensor (b) Two DIGIT sensors mounted on a 2F-140 ROBOTIQ gripper}\label{fig:digit_example}
\end{figure}

The DIGIT sensors provide tactile images and are used in order to implement the tactile feedback and closed-loop controllers. These sensors, which were originally designed in \citep{Lambeta2020}, contain a physical structure that is made up of: a reflective elastomer, an acrylic window, a 3D printed housing, a LED Printed Circuit Board (PCB) and a camera PCB. The sensor operates by recording the change in colours as a result of the deformation of the elastomer during the contact state. DIGIT sensors capture up to 30 Frames per Second (FPS) of RGB images with a resolution of 240x320 pixels. We mounted one DIGIT sensor (see Fig. \ref{fig:digit_example}) on each fingertip of the 2F-140 ROBOTIQ gripper. An additional 3D printed piece was designed and built so as to attach each sensor to each fingertip.

As the main contribution of this paper, we use touch images obtained from DIGIT sensors associated to contact properties to perform a tactile control. We design controllers to regulate the contact of a gripper to interact with objects without mapping the force applied to the sensing surface. The force values cannot be reconstructed because this low-cost image-based sensor does not provide a pixel-to-force mapping as other optical sensors \citep{ward2018tactip} or capacitive/resistive sensors \citep{pagoli2022large}. The novelty consists in constructing a version of a tactile controller for unknown environments without using the tactile Jacobian and the features of the forces as in other works as \citep{kappassov2020touch}.

\subsubsection{Contact detection}
\label{sec:contactdetection}
\begin{figure}[h]
     \centering
     \begin{subfigure}[b]{0.17\textwidth}
         \centering
         \includegraphics[width=\textwidth, height=2.5cm]{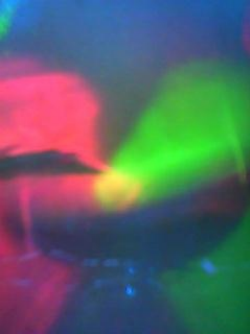}
         \caption{Unit A}
     \end{subfigure}
     \begin{subfigure}[b]{0.17\textwidth}
         \centering
         \includegraphics[width=\textwidth, height=2.5cm]{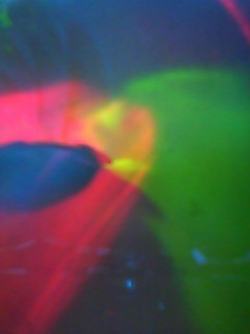}
         \caption{Unit A}
     \end{subfigure}
     \begin{subfigure}[b]{0.17\textwidth}
         \centering
         \includegraphics[width=\textwidth, height=2.5cm]{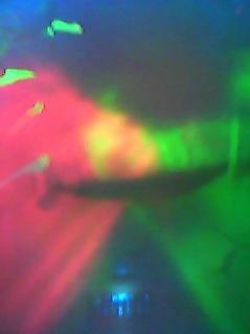}
         \caption{Unit B}
     \end{subfigure}
     \begin{subfigure}[b]{0.17\textwidth}
         \centering
         \includegraphics[width=\textwidth, height=2.5cm]{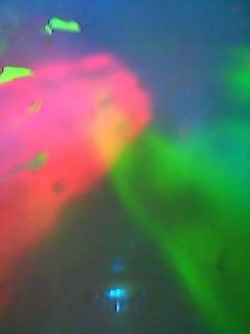}
         \caption{Unit B}
     \end{subfigure}
     \caption{Contact (a,c) and no contact (b,d) $image\_sensor$ from DIGIT units A (a,b) and B (c,d). The images from both sensors are not identical because both sensors were manufactured by hand}\label{fig:digit_images}
\end{figure}

Contact detection is formulated as a binary classification task as described in Algorithm \ref{grasping_contact_algorithm}. Given an input image $\boldsymbol{\varphi} = [R, G, B]$ acquired from the DIGIT sensor, a ground truth label y $\in [0,1]$ is assigned to no contact and contact images (see Fig. \ref{fig:digit_images}), respectively. Bearing this in mind, we use a CNN to solve this task. This method was chosen owing to its ability to learn and extract features from images, such as edges and simple textures, to more complex textures, patterns and parts of objects.

\begin{figure}[!htb]
     \centering
     \includegraphics[width=1\textwidth, height=4.5cm]{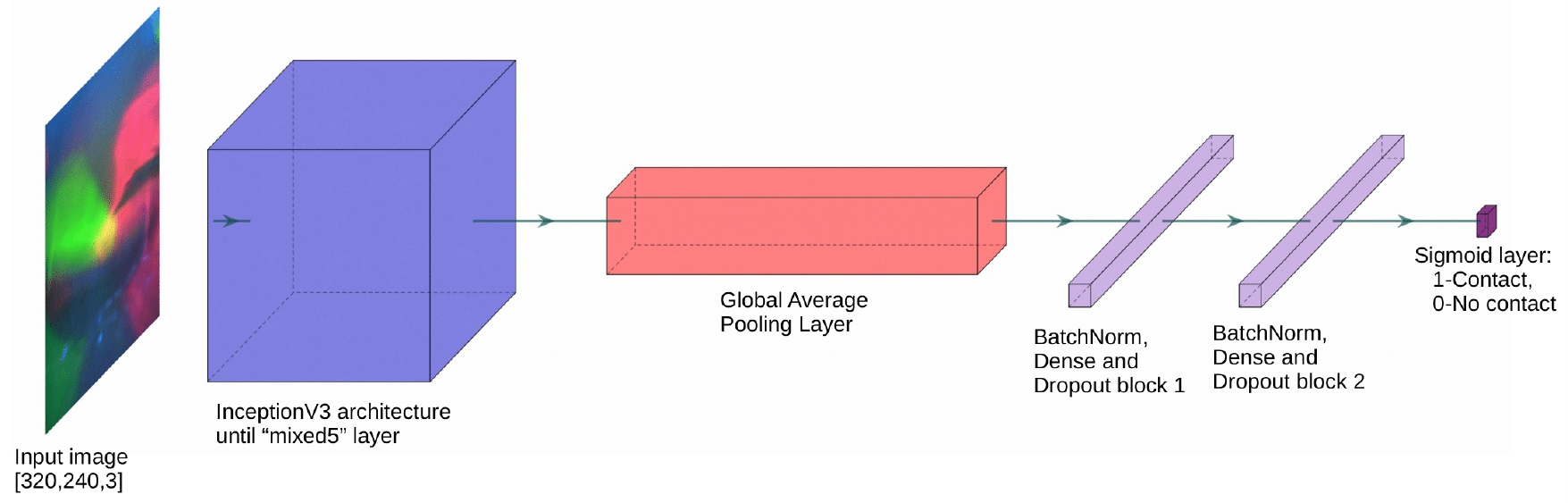}
     \caption{Architecture of our $\theta^{\mathrm{sensor\_unit}}_{\mathrm{contact}}$. It is based on the well-known InceptionV3 architecture, although the final layers have been modified}
     \label{fig:arch_net_contact}
\end{figure}

In our previous work \citep{castano2021touch}, we carried out extensive and rigorous experimentation with CNN architectures in order to discover which would be the most suitable for the purpose of $contact\_prediction$. We trained three different architectures: VGG16 \citep{simonyan2014very}, InceptionV3 \citep{szegedy2016rethinking} and MobileNetV2 \citep{sandler2018mobilenetv2} following a transfer learning strategy. Our dataset contained $\approx 16.000$ images, which were manually annotated by comparing a contact image with a no-contact image reference, from three DIGIT units and nine objects with different shapes and textures.  This work showed that InceptionV3 was the most appropriate architecture in terms of accuracy, robustness, and inference time. However, in 
the present work, it is necessary to reduce the size of its architecture in order to speed up the inference process in our embedded system for the robot. This reduction led to the decision to train one NN  ($\theta^{\mathrm{sensor\_unit}}_{\mathrm{contact}}$) can be unit A or B, thus, improving performance and evaluation values. 
Specifically, we used the InceptionV3 backbone up to the ``mixed5" layer as a feature extractor and modified the final layers to adapt them to our task. The 
final layers are made up of a GlobalAveragePooling2D layer and two blocks of Batch Normalisation, Dense, and Dropout layers. Finally, the output layer was added with a single neuron and a sigmoid activation function with a threshold 
$T^{\mathrm{sensor}}_{\mathrm{contact}}$ (Fig. \ref{fig:arch_net_contact}).

In order to execute the desired manipulation tasks, a closed-loop controller is implemented, as shown in 
Algorithm \ref{grasping_contact_algorithm}. First, 
it checks whether the pose of the robot is that of grasping or releasing the item of litter. It then executes {the contact prediction model to obtain the contact state for each sensor. 
When carrying out the subsequent grasping task, the robot closes the gripper by one step each time that the $contact\_prediction$ is equal to 0, signifying that the item of litter has not yet been grasped. During the releasing task, however, the robot opens the gripper by one step each time that the $contact\_prediction$ is equal to 1, signifying that the item of litter has not yet been released. The execution of Algorithm \ref{grasping_contact_algorithm} ends when the $contact\_prediction$ is equal to 1 (grasping task) or 0 (release task).

    

        

\begin{algorithm}[htpb]
    \centering
    \caption{Grasping-Contact algorithm}\label{grasping_contact_algorithm}
    \begin{algorithmic}[1]
    \Repeat
        \If{$robot\_in\_pose = 1$} \Comment{Grasp or release pose}
            \State norm\_image\_sensorA $\gets normaliseInput(image\_sensorA)$
            \State norm\_image\_sensorB $\gets normaliseInput(image\_sensorB)$
            \State contact\_prediction\_A $\gets \theta^{\mathrm{sensorA}}_{\mathrm{contact}}(norm\_image\_sensorA)$
            \State contact\_prediction\_B $\gets \theta^{\mathrm{sensorB}}_{\mathrm{contact}}(norm\_image\_sensorB)$
            \State contact\_prediction $\gets contact\_prediction\_A \land contact\_prediction\_B$
            \If{$task = grasp$} \Comment{Grasp litter object}
                \If{$contact\_prediction = 0$} \Comment{Detected no contact}
                    \State closeGripper(1) \Comment{Closing one step of the motor}
                
                \Else \Comment{Detected contact}
                    \State done $\gets 1$ \Comment{End}   
                \EndIf
            \EndIf
            \If{$task = release$} \Comment{Release litter object}
                \If{$contact\_prediction = 1$} \Comment{Detected contact}
                    \State openGripper(1) \Comment{Opening one step of the motor}
                
                \Else \Comment{Detected no contact}
                    \State done $\gets 1$    \Comment{End}
                \EndIf
            \EndIf
        
        \EndIf
    \Until {$done = 1$}
    \end{algorithmic}
\end{algorithm}

\subsubsection{Slip detection}
\label{sec:slipdetection}
 Various approaches with which to solve the slip detection task by classifying image sequences have been proposed in the literature (see Section \ref{sec:btactileandgraspingcontrol}). In contrast, this paper presents a grasping method based on slip detection as described in Algorithms \ref{slip_detection_algorithm} and  \ref{grasping_slip_algorithm}. Our proposal formulates the stage of slip detection  as an image classification problem between two classes: slip and stable. The slip class refers to object movement during robot manipulation, while the stable class implies no movement.

The slip detection (Algorithm \ref{slip_detection_algorithm}) takes as input a grayscale image sequence $\Phi = [\boldsymbol{\delta}_{\mathrm{t}}, \boldsymbol{\delta}_{\mathrm{t+1}}, \boldsymbol{\delta}_{\mathrm{t+2}}, \boldsymbol{\delta}_{\mathrm{t+3}}]$, whose length is empirically calculated in order to attain the best results in terms of accuracy and inference time and where $\boldsymbol{\delta} = 0.299 \times R + 0.587 \times G + 0.114 \times B$. A subtraction operation $\psi = \boldsymbol{\delta}_{\mathrm{t+3}} - \boldsymbol{\delta}_{\mathrm{t}}$ is applied to obtain the changes in the deformation of the elastomer. The subtracted image $\psi$ is very noisy because the pixel values are not identical in consecutive images. $\psi$ is, therefore, filtered by applying an opening morphological operation $\Psi = \psi \circ \kappa = (\psi \ominus \kappa) \oplus \kappa$, where $\circ, \ominus, \oplus$, and $\kappa$ denote opening, erosion, dilation and a structuring element, respectively. $\Psi$ is a binary image (black and white) that represents two possible states or labels $y \in [0,1]$. Label $y = 0$ is assigned to the stable class and label $y = 1$ is assigned to the slip class. Figure \ref{fig:slip_stable_example} shows examples of this pre-processing in which slip images produce white patterns (pixel value of 255), and stable images are almost black (pixel value of 0). The subtraction $\psi$ and filtering $\Psi$ operations correspond to the $filterImage$ function in Algorithm \ref{slip_detection_algorithm}.

\begin{figure}[!htb]
  \centering
  \includegraphics[width=\textwidth]{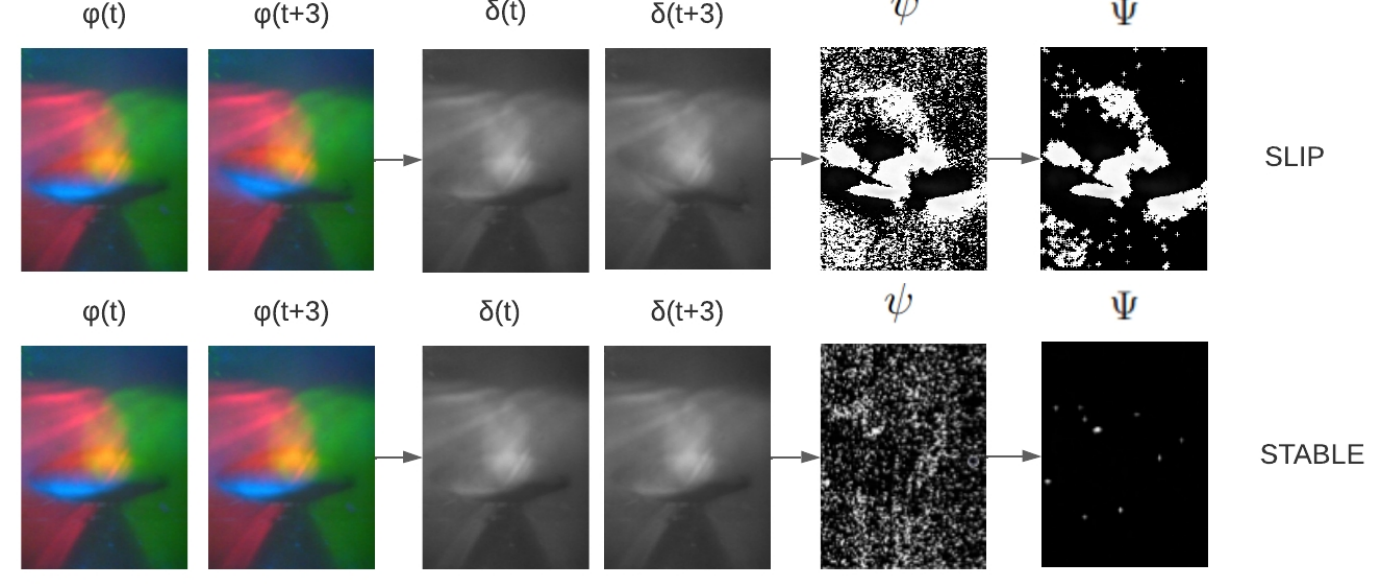}
  \caption{Example of pre-processing steps employed to obtain the final image before the classifying process.}\label{fig:slip_stable_example}
\end{figure}

Once $\Psi$ has been obtained, it can be classified as appertaining to the slip or stable class by using two different approaches and thresholds, which are compared and justified in Section \ref{sec:training_slip_detection}. The first consists of calculating the brightness of the $\Psi$ image, in which a threshold value of $T_{\mathrm{slip}}^{\mathrm{brightness}}$ is established as the final classifier. The second method is a CNN ($CNNPrediction$ in Algorithm \ref{slip_detection_algorithm}) whose architecture is described in Fig. \ref{fig:arch_net_stability}, followed by a threshold value $T_{\mathrm{slip}}^{\mathrm{cnn}}$ as the final classifier. Detailed descriptions of both methods are provided in Algorithms \ref{slip_detection_algorithm} and \ref{grasping_slip_algorithm}. The robot later closes the gripper by one step each time that the slippage event occurs. The execution ends when the robot is in the pose required to release the object.
 
 \begin{figure}[!htb]
     \centering
     \includegraphics[width=1\textwidth]{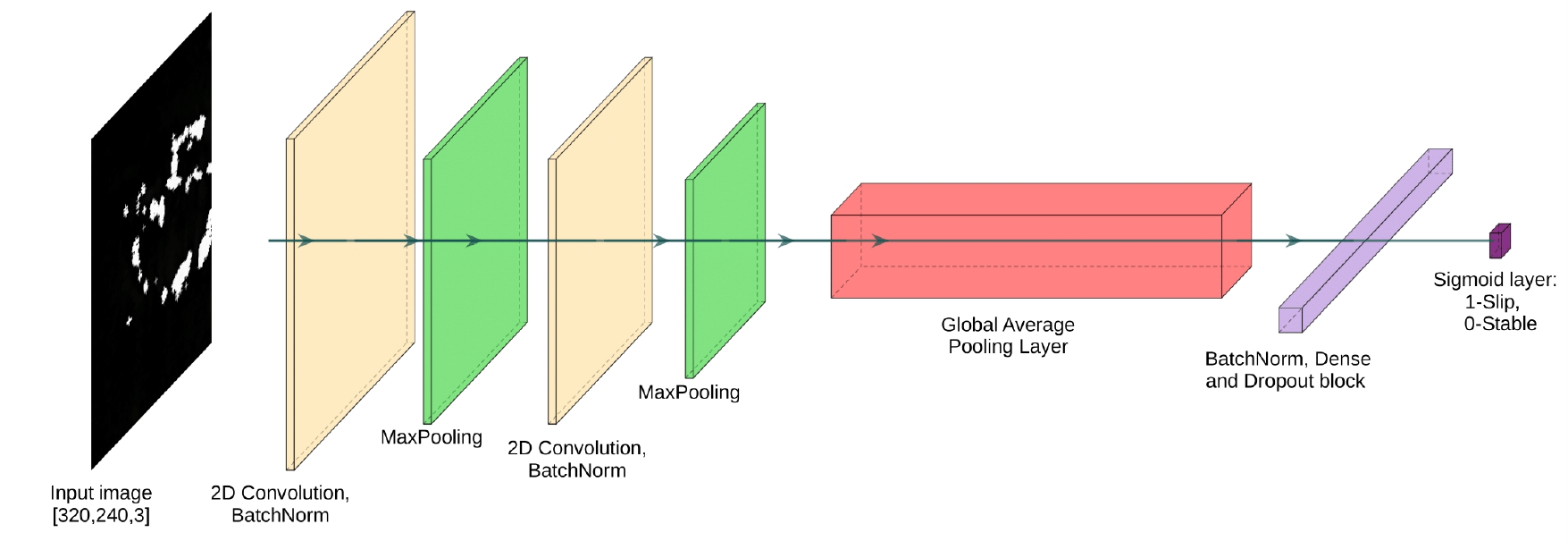}
     \caption{Architecture of $CNNPrediction$, which is based on a simple CNN architecture for image classification}
     \label{fig:arch_net_stability}
 \end{figure}

\begin{algorithm}[htpb]
    \centering
    \caption{Slip-Detection algorithm}\label{slip_detection_algorithm}
    \begin{algorithmic}[1]
    
    \Function{$slipDetection$}{$image\_sensor\_sequence, method$}

        \State $\Psi$ $\gets filterImage(image\_sensor\_sequence)$

        \If{$method = brightness$}
            \State brightness\_value $\gets \frac{\sum_{\mathrm{i=0}}^{\mathrm{h-1}} \sum_{\mathrm{j=0}}^{\mathrm{w-1}} \Psi (i,j)}{h \times w}$ \Comment{(h, w): $\Psi$ height and width}

            \If{$brightness\_value \geq T^{\mathrm{method}}_{\mathrm{slip}}$}
                \State grasping\_state $\gets 1$ 
            \Else
                \State grasping\_state $\gets 0$ 
            \EndIf
        \Else
            \State grasping\_state $\gets CNNPrediction(\Psi)$

            \If{$grasping\_state \geq T^{\mathrm{method}}_{\mathrm{slip}}$}
                \State grasping\_state $\gets 1$
            \Else
                \State grasping\_state $\gets 0$
            \EndIf
        \EndIf           
            \State \Return grasping\_state
    \EndFunction
    \end{algorithmic}
\end{algorithm}

\begin{algorithm}[!htb]
    \centering
        \caption{Grasping-Slip algorithm}\label{grasping_slip_algorithm}
    \begin{algorithmic}[1]
    \Repeat
        \If{$object\_is\_grasped = 1$} 
            \State grasping\_state\_A $\gets slipDetection(image\_sensorA\_sequence, method)$
            \State grasping\_state\_B $\gets slipDetection(image\_sensorB\_sequence, method)$
            \State final\_grasping\_state $\gets grasping\_state\_A \lor grasping\_state\_B$
            \If{$final\_grasping\_state = 1$} \Comment{Detected slip}
                \State closeGripper(1) \Comment{Close to avoid object falling}
            \EndIf

            \State robot\_in\_release\_pose $\gets checkRobotInReleasePose()$
            
        \EndIf
        \Until{$robot\_in\_release\_pose = 1$}
        
    \end{algorithmic}
\end{algorithm}

\section{Experimental results}
\label{sec:expe}

\subsection{Setup, performance metrics, and data}
This section provides descriptions of first the hardware devices used to train and test the NN, second the performance metrics used to express our results, and finally, general information regarding our datasets.

The first device is an NVIDIA A100 Tensor Core GPU with 40 GB of RAM memory. This device was used to train the visual and tactile perception modules.
The other is an NVIDIA Jetson AGX Xavier board. This device was used to test both modules and for real-time execution.

With regard to the evaluation metrics used, we have tested our visual system with an AP metric \citep{everingham2010pascal}, \citep{salton1983introduction} and our tactile system with an accuracy metric \citep{hossin2015review}. These metrics are well-established in literature and expressed our results in a complete and reliable manner. 

The AP metric, which is described in Eq. (\ref{eqn:ap}), is calculated for different $IoU$ thresholds, described in Eq. (\ref{eqn:iou}). 

\begin{equation}
AP_{\mathrm{IoU}} = \sum_{\mathrm{i \in \tilde{r}}}(r_{\mathrm{i+1}} - r_{\mathrm{i}}) \cdot \max_{\mathrm{\tilde{r}:\tilde{r} \geq r_{i+1}}} \rho(\tilde{r})
\label{eqn:ap}
\end{equation}

\begin{equation}
IoU = \frac{area(gt \cap pd)}{area(gt \cup pd)}
\label{eqn:iou}
\end{equation}

where $r$, $\rho$, $\tilde{r}$, $gt$, and $pd$ denote levels of recall, precision, and recall values, the ground truth, and the prediction bounding boxes. All these values are calculated to measure the performance of the NN models of our pipelines.

The accuracy metric described in Eq. (\ref{eqn:acc}) can be used for tactile testing because the tactile datasets are well-balanced.

\begin{equation}
    accuracy = \frac{TP + TN}{TP + TN + FP + FN}
\label{eqn:acc}
\end{equation}    

where $TP$, $TN$, $FP$, and $FN$$\in \mathbb{N}_{\geq 0}$ and denote True Positives, True Negatives, False Positives, and False Negatives. A $TP$ detection means that the system detects contact or slip and this is correct, while a $TN$ detection means that the system detects no contact or no slip and this is also correct. $FP$ and $FN$ detections occur when the system detects that there is contact or slip but this is not correct or when the system does not detect a contact or slip state but it exists, respectively.

We created three datasets for each specific task that required a training phase:  vision-based waste detection (D1), tactile-based contact detection (D2), and tactile-based slip detection (D3). For the visual module, we used 52 different household objects for the four classes (plastic, cardboard, glass, and metal), while for the tactile tasks we used only eight and six objects, for contact and slip detection, respectively. The visual module requires more objects for the training phase in order to learn a large variety of shapes, colours, etc., in different outdoor scenarios. Moreover, with regard to the tactile manipulation modules, different objects may produce similar tactile images because they share similar shapes and geometries, signifying that the tactile datasets do not require such a high number of objects. 

Table \ref{tab:general_info_datasets} shows that the D2 dataset contains more samples because it is easier to collect and label tactile images than household waste images in different environments. The D3 samples are also tactile images, but as explained earlier, a sequence of four gray-scale images is transformed into a single binary image in order to detect slip, and the final number of images is, therefore,  smaller than in the rest.

\begin{table}[hpbt]
\begin{center}
\begin{minipage}{\textwidth}
\caption{Number of total samples of each dataset for all classes, sets, and sensors}
\label{tab:general_info_datasets}
\begin{tabular*}{\textwidth}{@{\extracolsep{\fill}}cccc@{\extracolsep{\fill}}}
\toprule%
 & \textbf{D1}   & \textbf{D2}   & \textbf{D3}    \\
\midrule
\textbf{Number of objects} & 52 & 8 & 6    \\
\textbf{Number of total samples}  & 6943 & 24607 & 3540    \\
\botrule
\end{tabular*}
\end{minipage}
\end{center}
\end{table}

\subsection{Data collection and training for visual perception}
\label{sec:visual_validation}

\label{sec:training_visual_perception}

This section describes the training dataset, the training phase, and the results both with validation and test sets.

As there are not many household waste datasets containing objects that are dirty or partially destroyed, we had to create one ourselves (dataset D1). Images taken in this kind of environment were used for the training task. These environments are all at the Technological Scientific Park, in the area around our university, and include asphalt, pebbles, and green backgrounds. The objects in them are closer and further away, in addition to being partially occluded or shadowed by other elements in the scene and having different lighting conditions (see Fig. \ref{fig:dataset_places}). 

\begin{figure}[!htb]     
     \centering
     \begin{subfigure}[b]{0.24\textwidth}
         \centering
         \includegraphics[width=\textwidth]{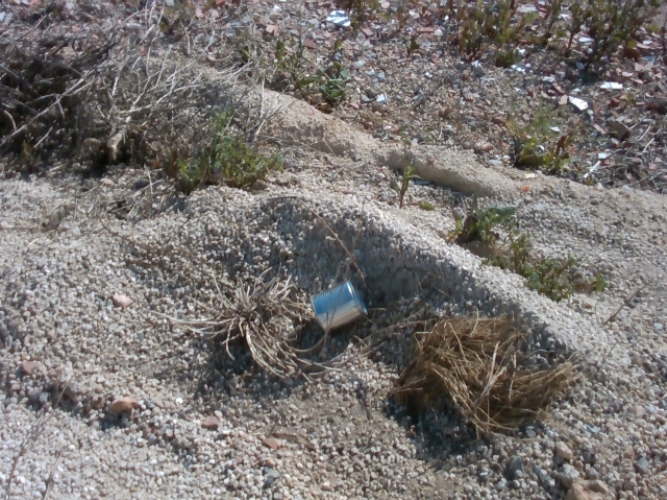}
     \end{subfigure}
     \begin{subfigure}[b]{0.24\textwidth}
         \centering
         \includegraphics[width=\textwidth]{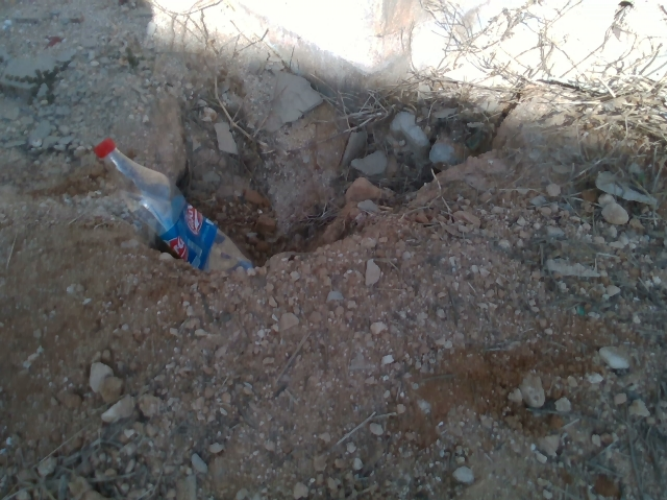}
     \end{subfigure}
     \begin{subfigure}[b]{0.24\textwidth}
         \centering
         \includegraphics[width=\textwidth]{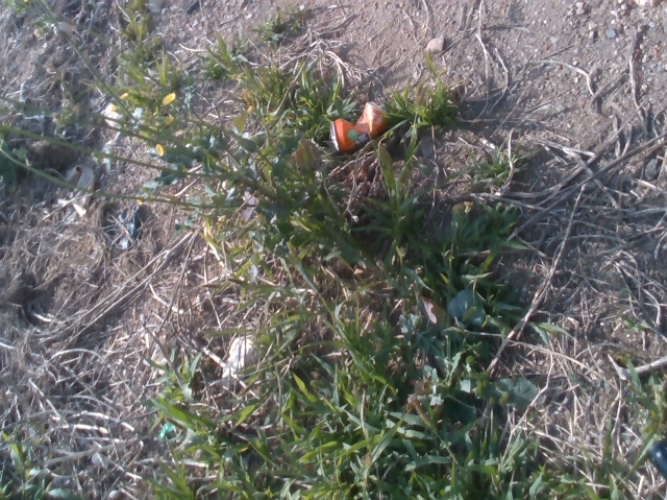}
     \end{subfigure}
     \begin{subfigure}[b]{0.24\textwidth}
         \centering
         \includegraphics[width=\textwidth]{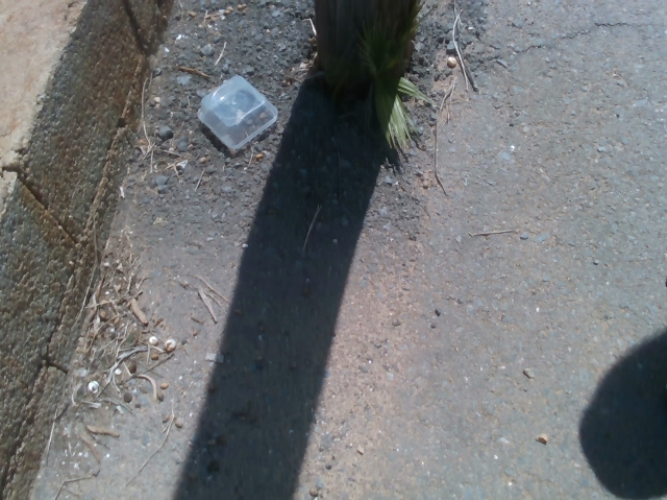}
     \end{subfigure}
    
     \caption{Variability environments in our dataset. They show different visual and physical features }\label{fig:dataset_places}
\end{figure}

Our system learns to classify and instantiate household waste into four different categories: plastic, cardboard, glass, and metal. 

In fact, each class represents a broad range of elements made of each material (see Table \ref{tab:object_distribution}). There are differently posed and sized water bottles, drinking bottles used by practitioners of sports, Tupperware containers, cans, glass beer bottles, or juice cardboard, among others. This way of naming the classes is common to other datasets and will facilitate the addition of new images from other sources if necessary. These images were extracted from video files. The video was recorded at a resolution of 640x480 pixels, using a {RealSense\texttrademark} D435i depth camera. As there is only one object per image, the number of images per class and per instance coincide. 
After processing the videos, 6,943 images were obtained to compose the dataset D1. These images were labeled with LabelMe \citep{5483185}, an image annotation tool. As will be noted,  all the classes are well balanced (see Table \ref{tab:object_distribution}).

\begin{table}[!htb]
\begin{center}
\begin{minipage}{\textwidth}
\caption{Distribution of dataset D1}
\label{tab:object_distribution}
\begin{tabular*}{\textwidth}{@{\extracolsep{\fill}}lccccc@{\extracolsep{\fill}}}
\toprule%
&\textbf{Plastic}    & \textbf{Cardboard}     & \textbf{Glass}      & \textbf{Metal}\\
\midrule
\textbf{Images}            & 1776       & 1626       & 1769       & 1772\\
\textbf{Different objects} &   21       &    9       &   10       &   12 \\
\textbf{Percentage}        & 25.58 \%   & 23.42 \%   & 25.48 \%   & 25.52 \%\\
\botrule
\end{tabular*}
\end{minipage}
\end{center}
\end{table}

For our experimentation, we split the dataset D1 into training, validation, and test sets. This was done by following the 70/20/10 proportion. The division is made randomly to guarantee that results are not dependent on how data is distributed and picked. Indeed, objects in the test set have not been used during training or validation phases before.

The learning process of the NNs has the following methodology. For Mask R-CNN and all its derivates, the training was split into three sub-training tasks. The first consisted of training the network heads for 40 epochs. The second lasted 80 epochs and trained ResNet backbone stage four and upwards. The last involved training the full NN for 40 more epochs.
In the first two sub-stages, the learning rate did not change, while in the last one, it was reduced to 10 times its original value. We used 2 images per GPU and 1,000 steps per epoch, signifying that a random subset of 2,000 images was used for training in each epoch. The learning rate was set to 0.001 using the Stochastic Gradient Descent (SGD) optimizer.

Only one training task was applied to both the second and the third set. This consisted of training the complete NN for 160 epochs.
The learning rate was reduced by \num{2e-7} during the first two and a half epochs,  from 0.0011 to 0.0010. There were also 26 images per GPU and 216 steps per epoch, signifying that each image in the training dataset was used per epoch. With regard to the optimizer, SGD was again used. The following results were obtained after the training process had been carried out (see Tables \ref{tab:results_val_dataset} and \ref{tab:results_test_dataset}). These results are expressed using $AP_{\mathrm{50}}$, $AP_{\mathrm{75}}$, and $AP_{\mathrm{90}}$ as metrics (AP values between 0 and 1).

\begin{table}[!htb]
\begin{center}
\begin{minipage}{\textwidth}
\caption{Results of validation dataset using AP}
\label{tab:results_val_dataset}
\begin{tabular*}{\textwidth}{@{\extracolsep{\fill}}ccccc@{\extracolsep{\fill}}}
\toprule%
\textbf{Algorithm} &\textbf{Backbone} & $\mathbf{AP_{\mathrm{50}}}$ &$\mathbf{AP_{\mathrm{75}}}$ & $\mathbf{AP_{\mathrm{90}}}$ \\ 
\midrule
Mask R-CNN  & ResNet50  & 0.966          & 0.934          & 0.595      \\
Mask R-CNN  & ResNet101 & 0.971          & 0.945          & 0.614      \\
YOLACT    & ResNet50  & 0.994          & 0.982          & 0.690      \\
YOLACT    & ResNet101 & 0.995          & 0.979          & 0.637      \\
YOLACT    & DarkNet53 & 0.997          & 0.983          & 0.720      \\
YOLACT++ & ResNet50  & 0.998          & 0.985          & 0.765      \\
YOLACT++ & ResNet101 & 0.998          & 0.980          & 0.754      \\
\botrule
\end{tabular*}
\end{minipage}
\end{center}
\end{table}

\begin{table}[!htb]
\begin{center}
\begin{minipage}{\textwidth}
\caption{Results of test dataset using AP and inference time. The inference time is calculated as the mean average of every image detected in the test set}
\label{tab:results_test_dataset}
\begin{tabular*}{\textwidth}{@{\extracolsep{\fill}}cccccc@{\extracolsep{\fill}}}
\toprule%
\textbf{Algorithm} &\textbf{Backbone} & $\mathbf{AP_{\mathrm{50}}}$ &$\mathbf{AP_{\mathrm{75}}}$ & $\mathbf{AP_{\mathrm{90}}}$ & \textbf{Inference time (ms)} \\ 
\midrule
Mask R-CNN  & ResNet50   & 0.958          & 0.935          & 0.585           & 74.835\\
Mask R-CNN  & ResNet101  & 0.967          & 0.945          & 0.612           & 86.032\\
YOLACT      & ResNet50   & 0.998          & 0.975          & 0.701           & 18.187 \\
YOLACT      & ResNet101  & 0.996          & 0.968          & 0.661           & 27.644 \\
YOLACT      & DarkNet53  & \textbf{0.998} & \textbf{0.980} & 0.696           & \textbf{17.213}\\
YOLACT++    & ResNet50   & 0.994          & 0.975          & 0.731           & 21.085 \\
YOLACT++    & ResNet101  & \textbf{0.999} & 0.974          & \textbf{0.747}  & 29.831 \\
\botrule
\end{tabular*}
\end{minipage}
\end{center}
\end{table}

During the training process, YOLACT++ with ResNet50 obtained the best results with the validation dataset. It achieved 99.8\%, 98.5\% and 76.5\% in $AP_{\mathrm{50}}$, $AP_{\mathrm{75}}$ and $AP_{\mathrm{90}}$ respectively. The first two metrics did not help much as regards choosing an algorithm, but the last allowed us to choose YOLACT as the best method. After using the test dataset, the final results were obtained and they had a similar trend. The best model was again YOLACT++ with ResNet101 in both $AP_{\mathrm{50}}$ and $AP_{\mathrm{90}}$, achieving a score of 99.9\% and 74.7\% respectively, while in $AP_{\mathrm{75}}$ the best model was YOLACT with DarkNet53, achieving an average precision of 98\%. Since we have two different NNs with bigger and smaller backbones, inference time is also a useful tool for the selection of the best NN model. This will help to determine which combination is the fastest in the inference task. The results are shown in Table \ref{tab:results_test_dataset}. As will be observed, the fastest model is the combination of YOLACT with DarkNet53, achieving 17.2 ms. There is a huge step between the YOLACT and Mask R-CNN models, with YOLACT being an average of more than 50 ms faster.

All of these results led to the decision that the NN chosen for the process would be YOLACT with DarkNet53 as a backbone. We chose this NN since we prioritise speed (it is 57\% faster when compared to that with the best $AP_{\mathrm{90}}$) over accuracy and also because $AP_{\mathrm{75}}$ is sufficiently good for our process.  The accuracy of the NN per class in our dataset is shown in Table \ref{tab:results_class}. 

\begin{table}[hpbt]
\begin{center}
\begin{minipage}{\textwidth}
\caption{Results of test dataset using AP as metrics. These results are calculated per object class and using YOLACT NN with DarkNet53 as a backbone}
\label{tab:results_class}
\begin{tabular*}{\textwidth}{@{\extracolsep{\fill}}cccccc@{\extracolsep{\fill}}}
\toprule%
& \textbf{Plastic} & \textbf{Cardboard} & \textbf{Glass} & \textbf{Metal} & \textbf{Total}  \\
\midrule
$\mathbf{AP_{\mathrm{50}}}$ & 1.000   & 1.000  & 0.996 & 0.996 & 0.998  \\ 
$\mathbf{AP_{\mathrm{75}}}$ & 0.989   & 0.988  & 0.972 & 0.971 & 0.980  \\ 
$\mathbf{AP_{\mathrm{95}}}$ & 0.800   & 0.771  & 0.582 & 0.631 & 0.696  \\
\botrule
\end{tabular*}
\end{minipage}
\end{center}
\end{table}

We also show some outdoor recognition examples when our visual model is used with unknown samples from the four classes. These are provided in Fig. \ref{fig:yolact_results}. The following confusion matrix was obtained when using the NN chosen (see Fig. \ref{fig:confusion_matrix}), in which the most confusion occurs between the plastic and glass classes.

\begin{figure}[htbp]
     \centerline{
         \includegraphics[width=.24\textwidth]{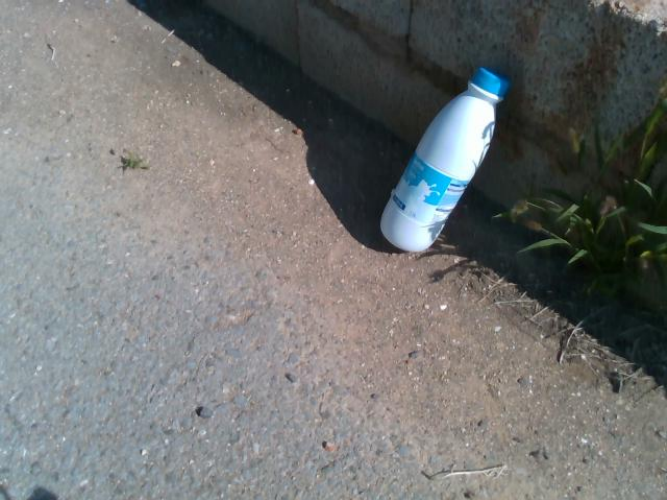}
         \includegraphics[width=.24\textwidth]{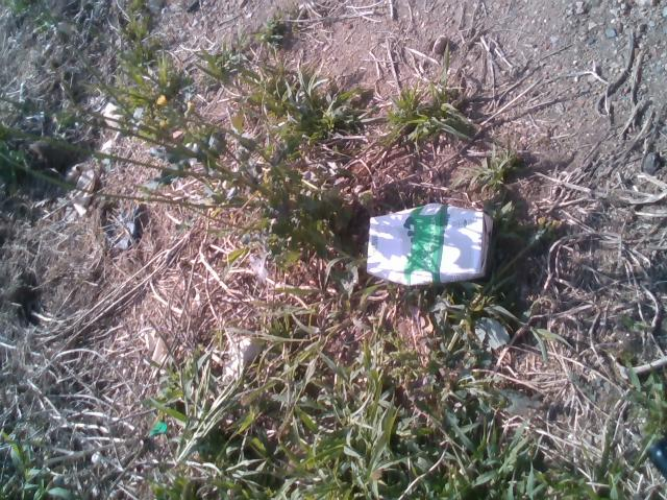}
         \includegraphics[width=.24\textwidth]{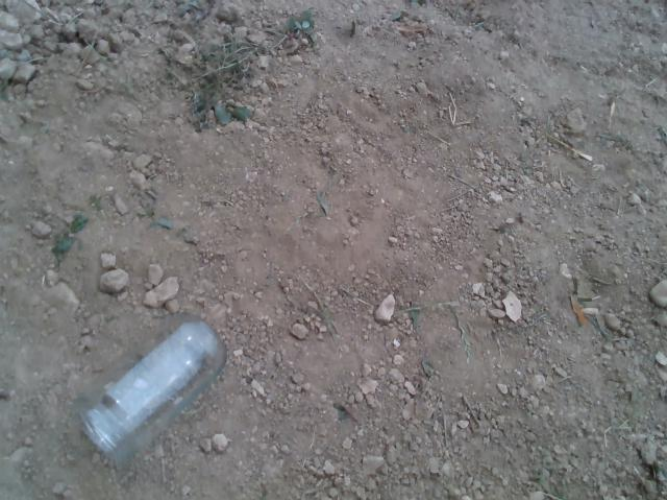}
         \includegraphics[width=.24\textwidth]{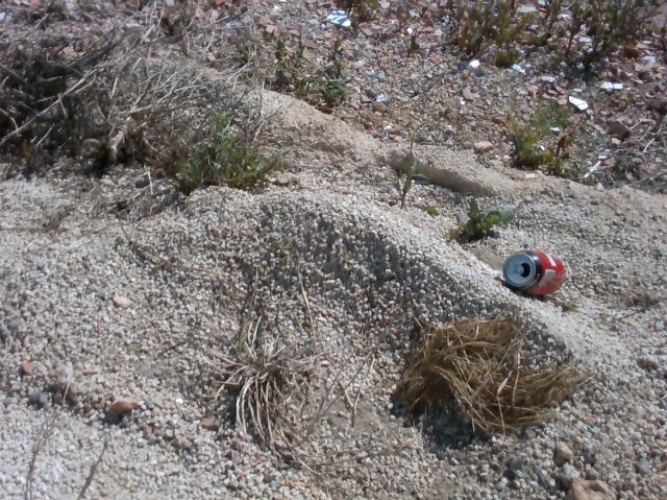}}
     \centerline{
         \includegraphics[width=.24\textwidth]{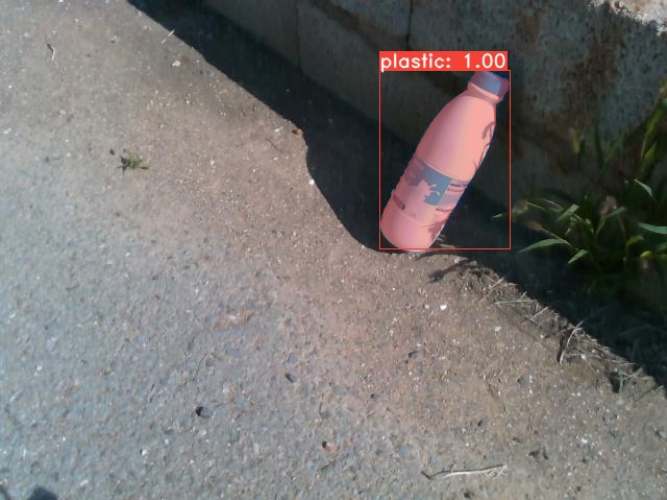}
         \includegraphics[width=.24\textwidth]{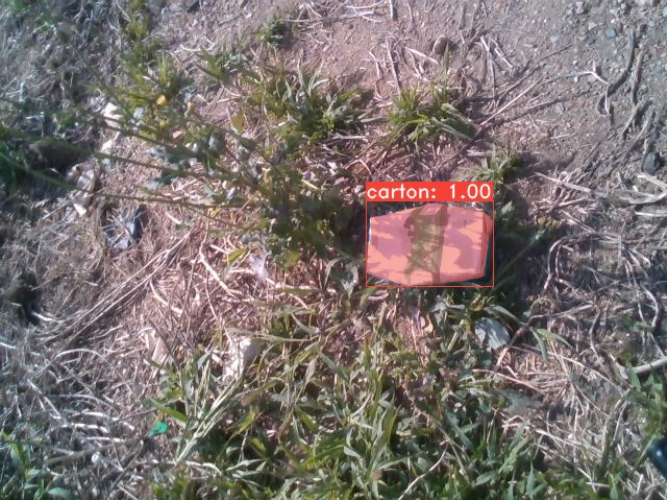}
         \includegraphics[width=.24\textwidth]{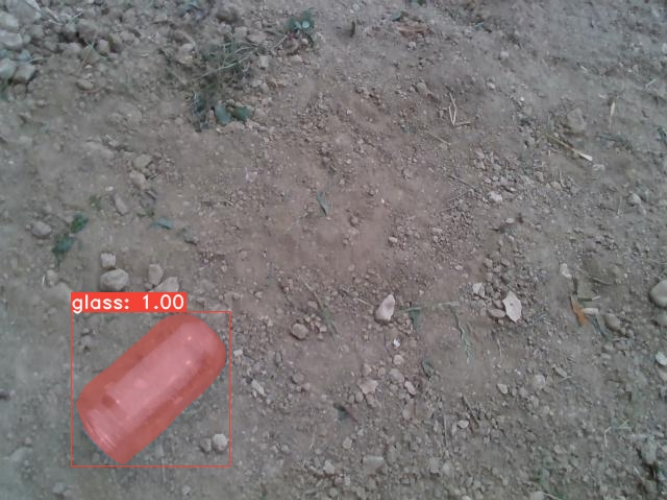}
         \includegraphics[width=.24\textwidth]{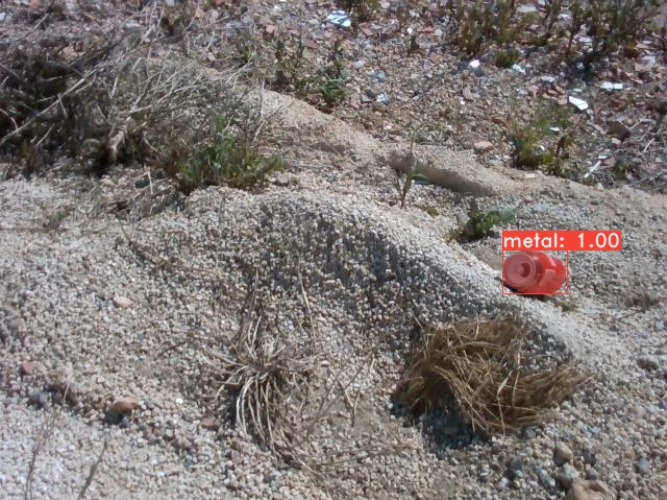}}
     \caption{Examples of outdoor litter detection. These examples contain one object from each class (cardboard, plastic, metal, and glass) in different environments } 
     \label{fig:yolact_results}
\end{figure}

\begin{figure}[htpb]
     \centering

         \includegraphics[width=0.65\textwidth, height=4.5cm]{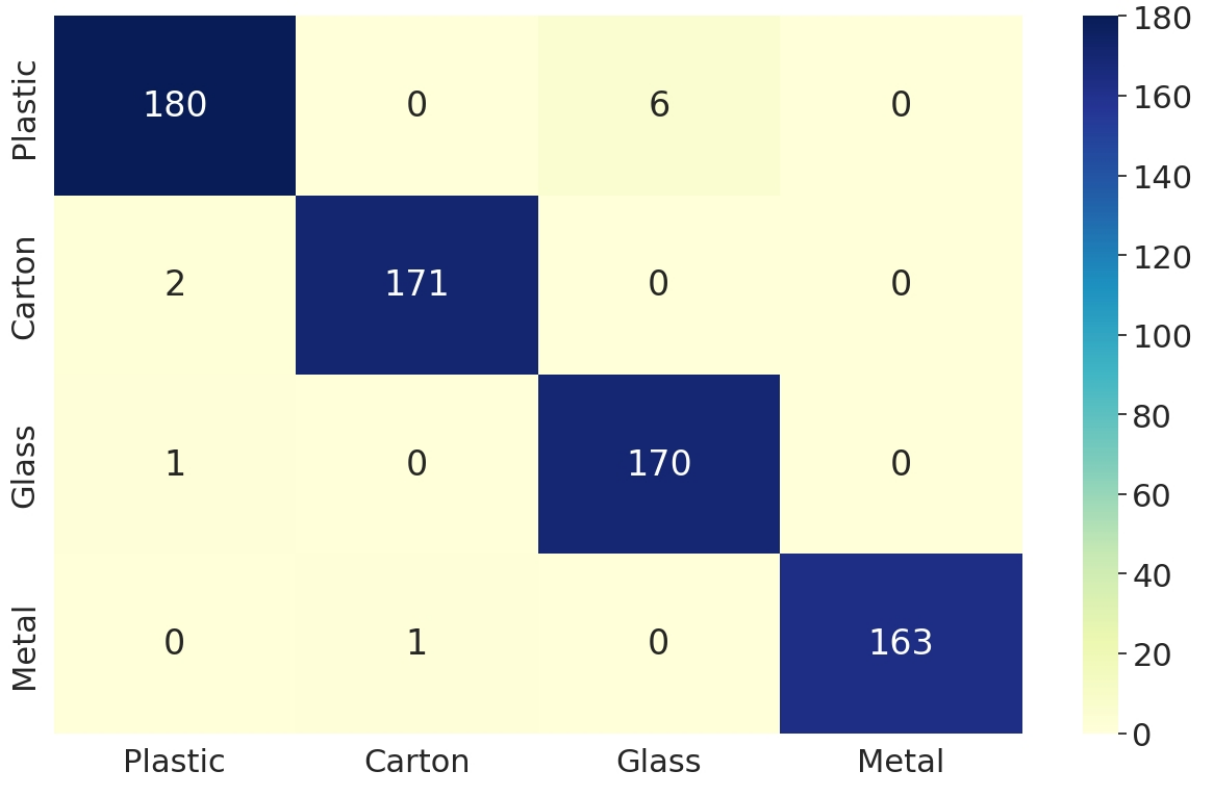}

     \caption{Test confusion matrix with the NN selected. Rows indicate true labels and columns indicate predicted labels}\label{fig:confusion_matrix}
\end{figure}

\subsection{Tactile data collection and training for contact detection}
\label{sec:training_contact_detection}

This section describes the tactile dataset D2 for the contact detection task, the training phase, and the results with the test set.

The tactile data was collected by performing consecutive maneuvers and recording the images from the DIGIT sensors mounted on the gripper. A human operator changed the pose of the objects for each robotic grip.

Eight objects were used to create the dataset D2 = $[\boldsymbol{\varphi}_{\mathrm{1}}, \boldsymbol{\varphi}_{\mathrm{2}}, \boldsymbol{\varphi}_{\mathrm{3}}, ..., \boldsymbol{\varphi}_{\mathrm{n}}]$ where ${\mathrm{n}}$ is the total number of images. Approximately the same number of images was obtained for each object. Eight objects were sufficient to form the dataset because they differed in terms of size, shape, deformation, weight, texture, and material (see Fig. \ref{fig:dataset_objects}). We discarded objects that are completely deformable or very narrow such as plastic bags or cardboard sheets due to the limitations of the fingertips and tactile sensing area in size.

\begin{figure}[htpb]
     \centering
     \begin{subfigure}[b]{0.24\textwidth}
         \centering
         \includegraphics[width=\textwidth, height=2cm]{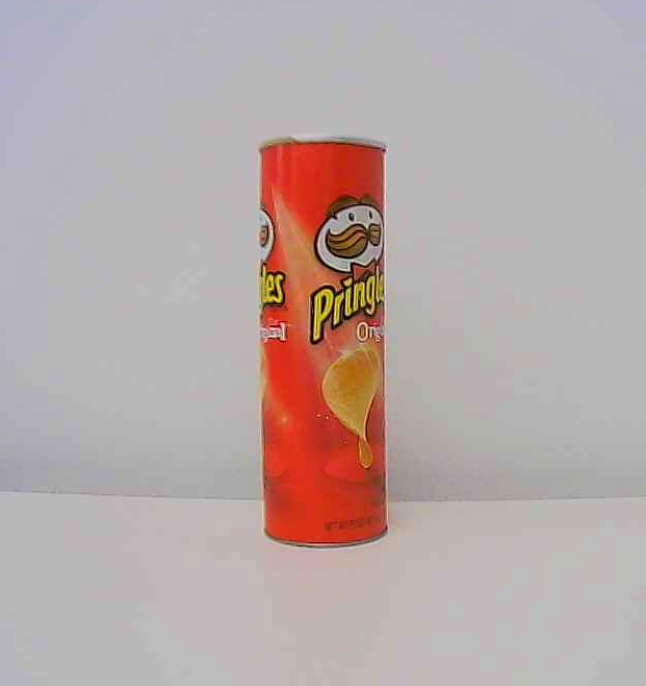}
         \caption{Chips can}
     \end{subfigure}
     \begin{subfigure}[b]{0.24\textwidth}
         \centering
         \includegraphics[width=\textwidth, height=2cm]{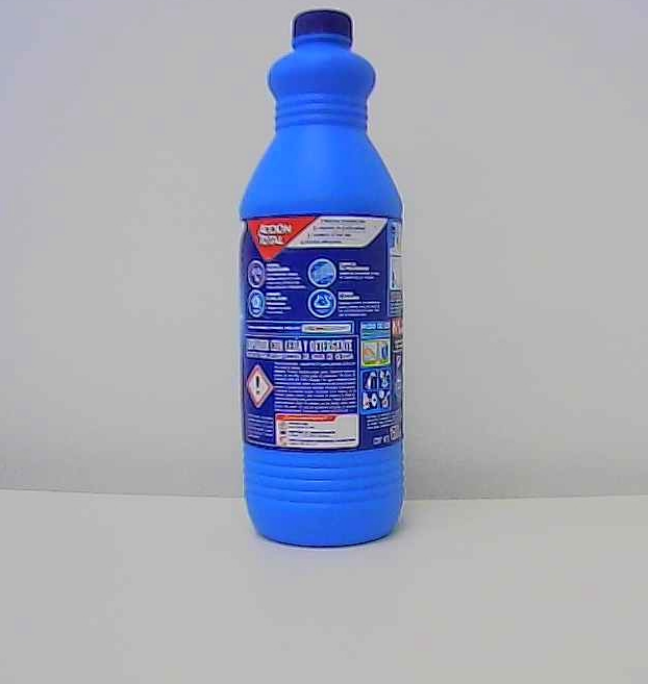}
         \caption{Bleach bottle}
     \end{subfigure}
     \begin{subfigure}[b]{0.24\textwidth}
         \centering
         \includegraphics[width=\textwidth, height=2cm]{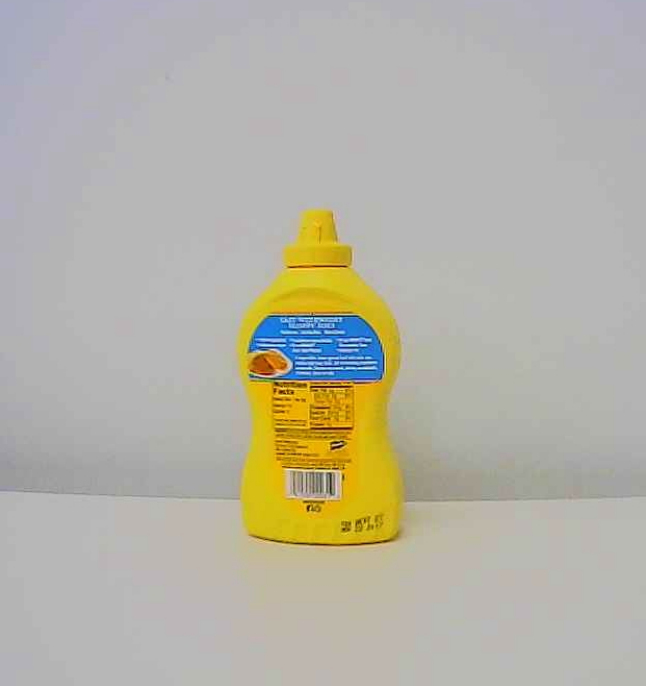}
         \caption{Mustard bottle}
     \end{subfigure}
     \begin{subfigure}[b]{0.24\textwidth}
         \centering
         \includegraphics[width=\textwidth, height=2cm]{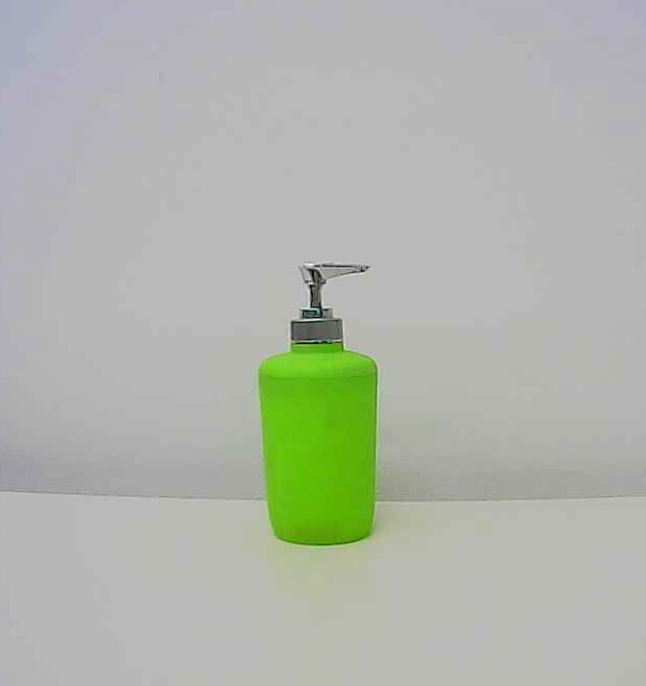}
         \caption{Shampoo bottle}
     \end{subfigure}

     \centering
     \begin{subfigure}[b]{0.24\textwidth}
         \centering
         \includegraphics[width=\textwidth, height=2cm]{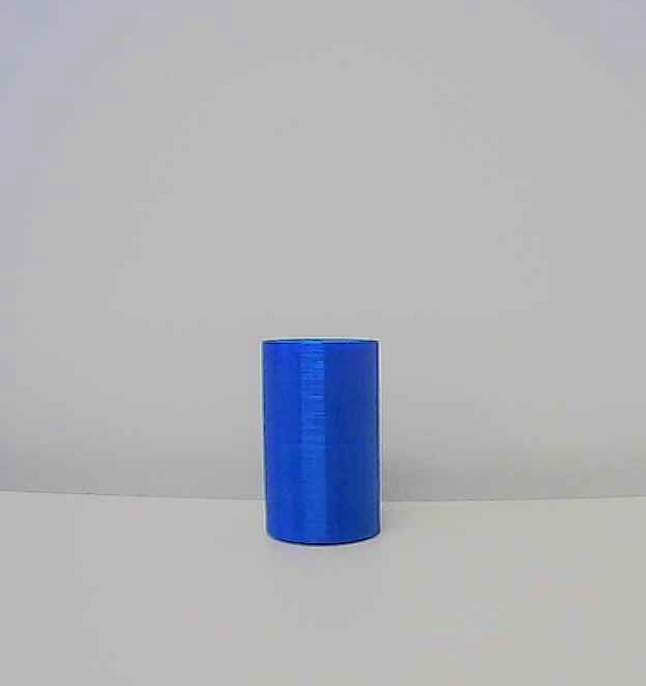}
         \caption{Plastic cylinder}
     \end{subfigure}
     \begin{subfigure}[b]{0.24\textwidth}
         \centering
         \includegraphics[width=\textwidth, height=2cm]{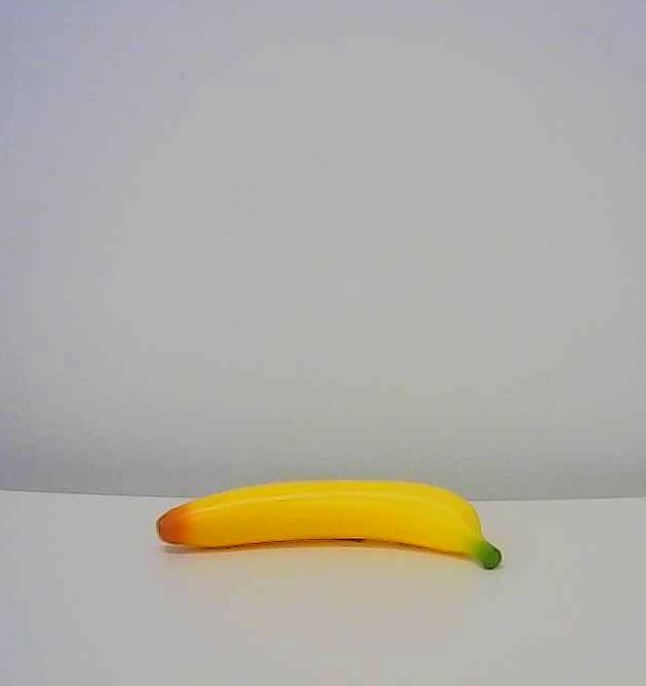}
         \caption{Plastic banana}
     \end{subfigure}
     \begin{subfigure}[b]{0.24\textwidth}
         \centering
         \includegraphics[width=\textwidth, height=2cm]{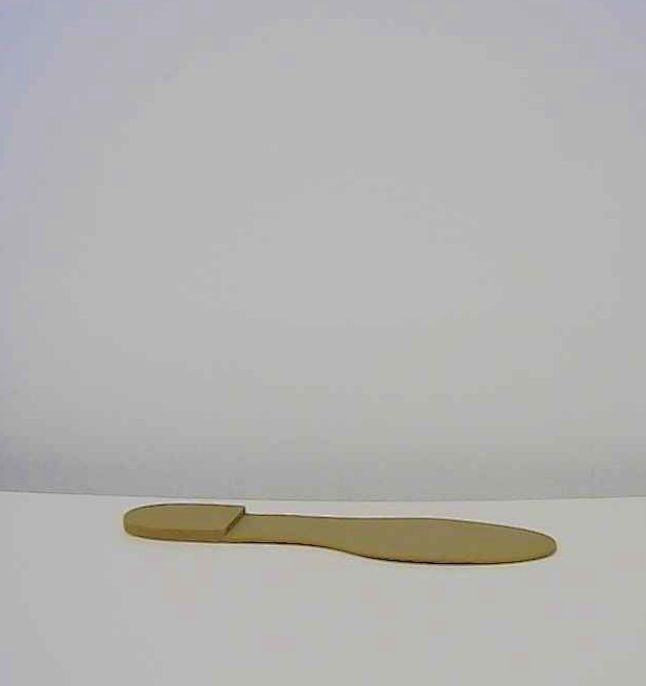}
         \caption{Shoe sole}
     \end{subfigure}
     \begin{subfigure}[b]{0.24\textwidth}
         \centering
         \includegraphics[width=\textwidth, height=2cm]{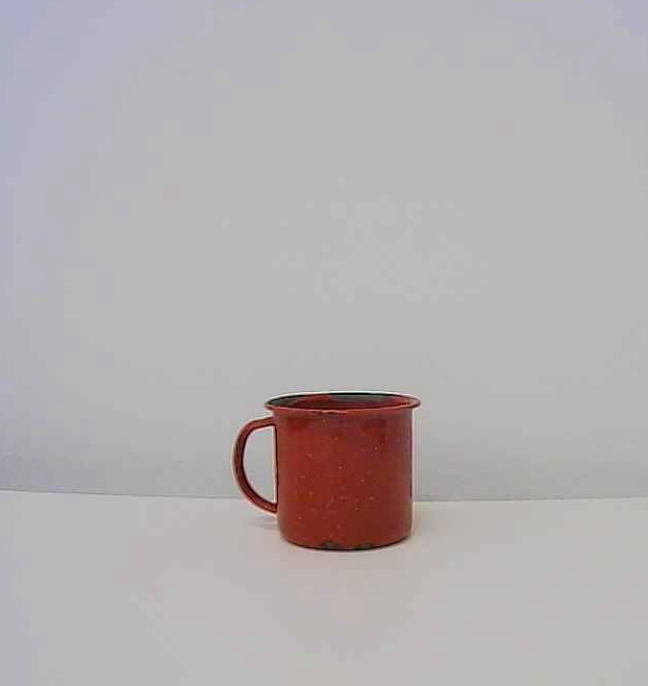}
         \caption{Ceramic cup}
     \end{subfigure}
     \caption{Objects used to create the dataset D2, which was used to train, validate and test the contact detection method and also to train the slip detection CNN}\label{fig:dataset_objects}
\end{figure}

We train our CNN (see Section \ref{sec:contactdetection}) with each sensor, units A and B, because the images extracted from DIGIT sensors are not identical (see Fig. \ref{fig:digit_images}). We, therefore, achieved better results with individual models rather than a single model for all the sensors. Two datasets $[D2_{\mathrm{A}}, D2_{\mathrm{B}}]$ were, therefore, designed (see Table \ref{tab:datasets_info}), and each dataset was split into three subsets: training (70\%), validation (20\%) and test (10\%). The dataset is roughly balanced in terms of sensor and class samples. This procedure enabled our NNs to learn and check their knowledge with 90\% of the data during the training phase. The remaining 10\% of the data was used to perform an evaluation test once the training phase had been completed.

\begin{table}[hpbt]
\begin{center}
\begin{minipage}{\textwidth}
\caption{Distribution of images per sensor unit $[D2_{\mathrm{A}}, D2_{\mathrm{B}}]$ considering} the class (contact, no contact) in training, validation, and test sets
\label{tab:datasets_info}
\begin{tabular*}{\textwidth}{@{\extracolsep{\fill}}ccccccc@{\extracolsep{\fill}}}
\toprule%
\multicolumn{1}{l}{} & \multicolumn{2}{c}{\textbf{Train}}    & \multicolumn{2}{c}{\textbf{Validation}} & \multicolumn{2}{c}{\textbf{Test}}  \\
\midrule
\textbf{Class} & A    & \multicolumn{1}{c}{B} & A    & \multicolumn{1}{c}{B}   & A   & \multicolumn{1}{c}{B}   \\ 
\textbf{Contact} & 4035   & 4683  & 1224 & 1344 & 603 & 663 \\ 
\textbf{No contact} & 3803   & 4683  & 1092 & 1345 & 539 & 663 \\
\botrule
\end{tabular*}
\end{minipage}
\end{center}
\end{table}

A transfer learning strategy was applied in order to carry out the training phase. The layers up to the ``mixed5" layer were, therefore, set as being non-trainable. The remaining layers were set as trainable, including the head. A Root Mean Squared Propagation (RMSProp) optimizer was used with a learning rate of $3  \times 10^{-6}$, a batch size of 24, and a binary cross entropy loss. Once both models had been trained, the evaluation process was executed on the test dataset. The results are expressed in terms of the accuracy metric, previously described in Eq. (\ref{eqn:acc}). Table \ref{tab:contact_results} shows accuracy values for both the sensors and the three different thresholds ($T_{\mathrm{contact}}^{\mathrm{sensor\_unit}}$). Although these values are very similar, $T_{\mathrm{contact}}^{\mathrm{sensor\_unit}} = 0.5$ was chosen in order to prevent wrong detections resulting from the hysteresis of DIGIT sensors. This hysteresis is produced when the elastomer is recovering its initial shape after the contact.

\begin{table}[hpbt]
\begin{center}
\begin{minipage}{\textwidth}
\caption{Accuracy (Acc) values for sensors A and B, and $T_{\mathrm{contact}}^{\mathrm{sensor\_unit}} = 0.4, 0.5, 0.6$. Accuracy values are between 0 and 1}
\label{tab:contact_results}
\begin{tabular*}{\textwidth}{@{\extracolsep{\fill}}cccc@{\extracolsep{\fill}}}
\toprule%
\textbf{Sensor unit} & \textbf{Acc (T=0.4)} & \textbf{Acc (T=0.5)} & \textbf{Acc (T=0.6)}  \\
\midrule
\textbf{A}  & 0.969 & 0.964 & 0.963                     \\
\textbf{B} & 0.996 & 0.996 & 0.994                         \\
\botrule
\end{tabular*}
\end{minipage}
\end{center}
\end{table}

\subsection{Tactile data collection and training for slip detection}
\label{sec:training_slip_detection}

This section describes the tactile dataset D3 for the slip detection task, the training or tuning phase, and the results with the test set.

$D3 = [\boldsymbol{\Psi_{\mathrm{1}}}, \boldsymbol{\Psi}_{\mathrm{2}},$ $\boldsymbol{\Psi}_{\mathrm{3}}, ..., \boldsymbol{\Psi}_{\mathrm{n}}]$ where ${\mathrm{n}}$ is the total number of images. As slip detection is a different task to contact detection, it was necessary to create this new dataset in order to capture the corresponding tactile images generated when the objects slip. The slip class images were generated by applying three external instabilities to the object, while the stable class images were regular images with no disturbances. As noted in Fig. \ref{fig:slip_instabilities}, one rotational and two translational movements were applied to each object. The goal was to detect not only slippage, but also any other possible perturbations during the robot manipulation. These perturbations could result in the object falling to the ground, thus, generating a collection failure.

\begin{figure}[!htb]
     \centering
     \begin{subfigure}[b]{0.3\textwidth}
         \centering
         \includegraphics[width=\textwidth]{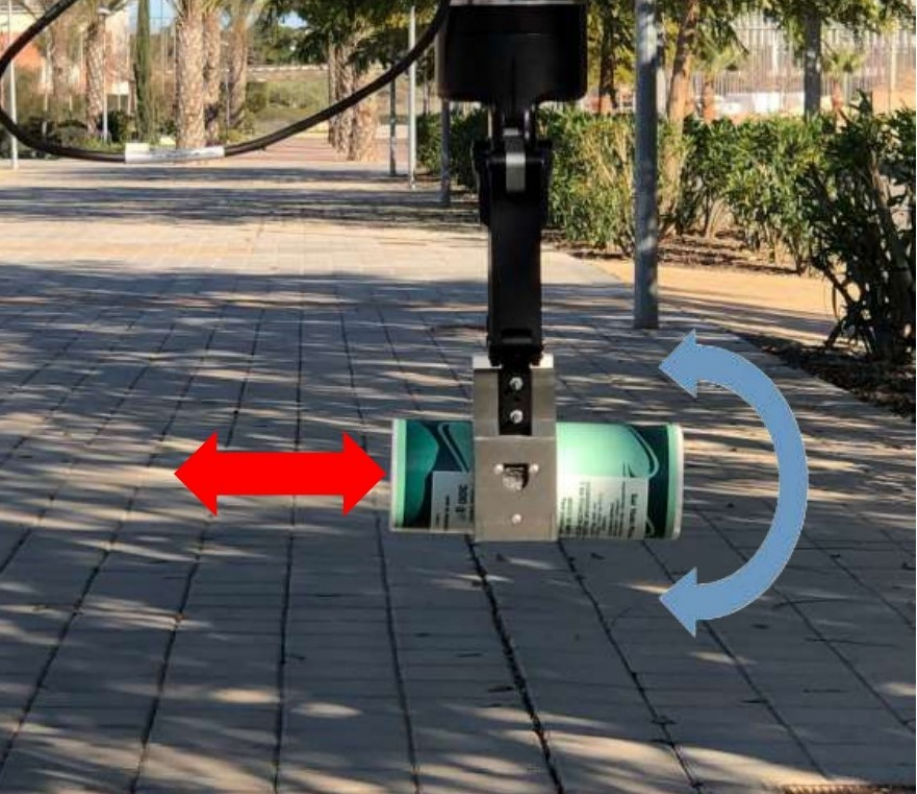}
         
     \end{subfigure}
     \begin{subfigure}[b]{0.3\textwidth}
         \centering
         \includegraphics[width=\textwidth]{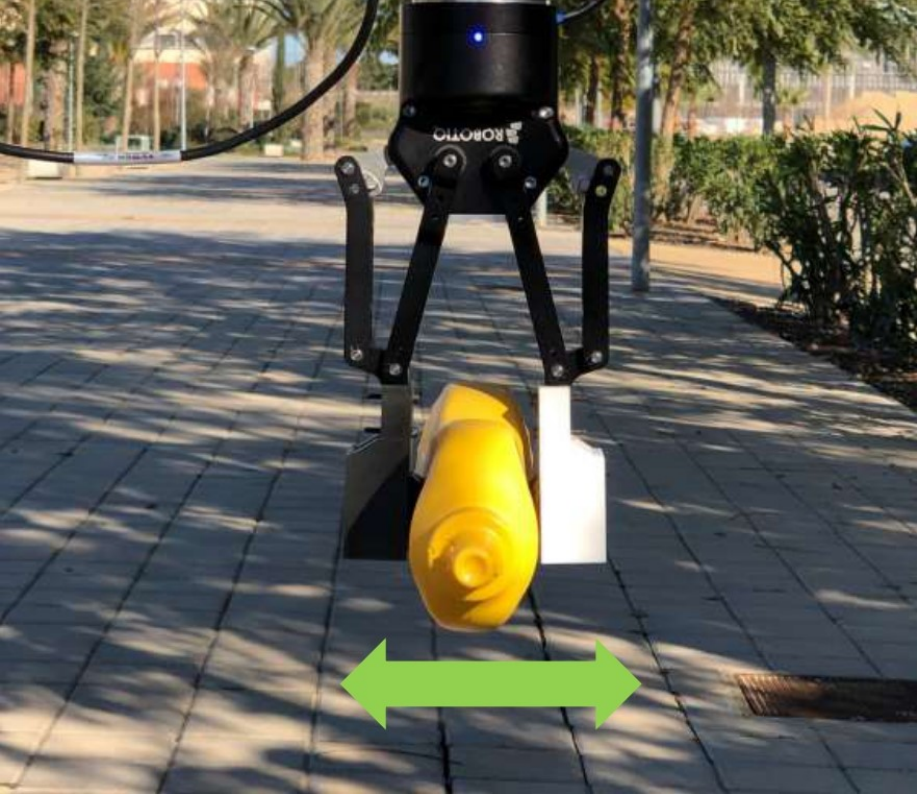}
         
     \end{subfigure}
     \caption{Translational (red and green) and rotational (blue) perturbations produced by a human operator. These movements include all perturbations that the object could undergo during the manipulation task}\label{fig:slip_instabilities}
\end{figure}

Six objects ($b$, $c$, $e$, $f$, $g$, and $h$ in Fig. \ref{fig:dataset_objects}) were selected from the set of eight used for tactile detection. They were sufficiently varied to be able to form the training dataset, which was divided into two subsets: training (70\%) and validation (30\%) (see Table \ref{tab:datasets_info_slip}). Objects $a$ and $d$ were discarded from D3  
in order to avoid repeating cylindrical shapes. The number of images obtained for the six objects was roughly balanced and the images were different from those in the contact detection dataset.

\begin{table}[hpbt]
\begin{center}
\begin{minipage}{\textwidth}
\caption{Distribution of images ($\boldsymbol{\Psi}$) per sensor unit [$D3_{\mathrm{A}}$, $D3_{\mathrm{B}}$]} considering class (stable, slip) in training and validation sets
\label{tab:datasets_info_slip}
\begin{tabular*}{\textwidth}{@{\extracolsep{\fill}}ccccc@{\extracolsep{\fill}}}
\toprule%
\multicolumn{1}{l}{} & \multicolumn{2}{c}{\textbf{Train}}    & \multicolumn{2}{c}{\textbf{Validation}}  \\
\midrule
\textbf{Class} & A    & \multicolumn{1}{c}{B} & A    & \multicolumn{1}{c}{B}      \\ 
\textbf{Stable}                & 620 & 620                   & 265 & 265                       \\
\textbf{Slip}                  & 627 & 613                   & 268 & 262                       \\
\botrule
\end{tabular*}
\end{minipage}
\end{center}
\end{table}

 Three novel objects were also added to D3 for testing (see Fig. \ref{fig:dataset_objects_slip_test}). In this way, we can evaluate the generalisation capabilities of the proposed algorithms with previously unseen objects.

\begin{figure}[!htb]
     \centering
     \begin{subfigure}[b]{0.25\textwidth}
         \centering
         \includegraphics[width=\textwidth, height=2cm]{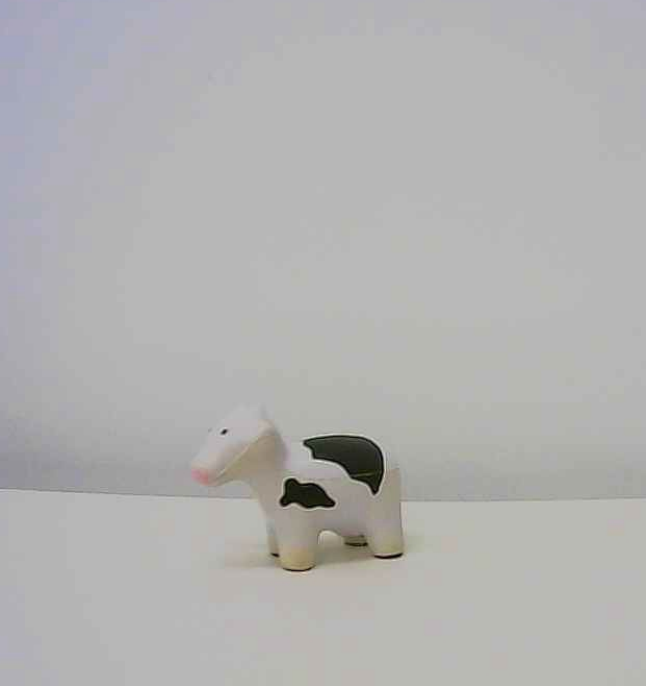}
         \caption{Deformable toy}
     \end{subfigure}
     \begin{subfigure}[b]{0.25\textwidth}
         \centering
         \includegraphics[width=\textwidth, height=2cm]{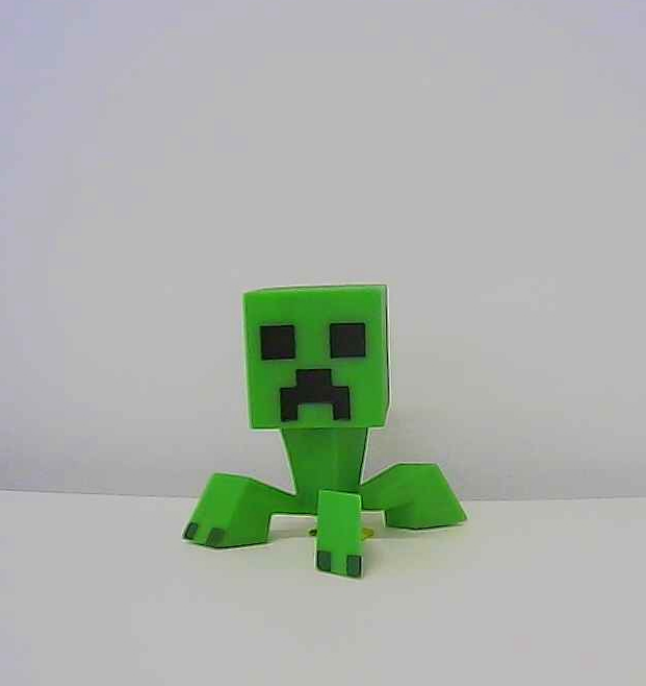}
         \caption{Plastic toy}
     \end{subfigure}
     \begin{subfigure}[b]{0.25\textwidth}
         \centering
         \includegraphics[width=\textwidth, height=2cm]{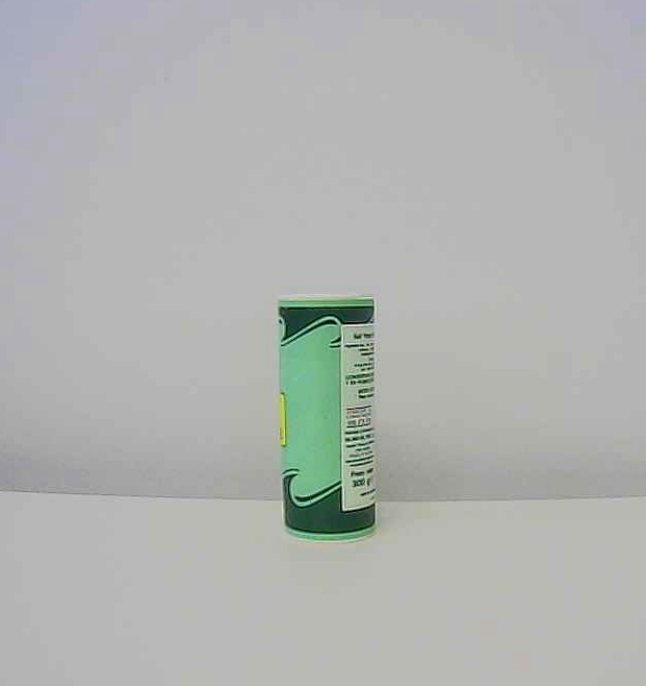}
         \caption{Salt bottle}
     \end{subfigure}
     \caption{Objects in our test set. These objects differ from those in the training set in terms of size, textures, deformability, etc}\label{fig:dataset_objects_slip_test}
\end{figure}

No training process was required in order to calculate the brightness of an image. The CNN did, however, require this training process. The low number of parameters allowed us to train the full network from scratch. An Adam optimizer \citep{kingma2014adam} was used with a learning rate of $1 \times 10^{-4}$, a batch size of 32, and a binary cross entropy loss.

After the training process had been completed, the performance of the model was calculated with the test set. This set was different from the training and validation sets because it was not made up of shuffled images with their respective labels. The test set contained three continuous videos, one per object, to which we applied these three external instabilities, five times each. This signifies that there are 15 instabilities per object, and 45 in total per sensor. 

The aim of this experiment was for the models to detect every case of instability as a slip class without detecting false positives. The results are expressed in terms of instability-detection accuracy and inference time in our embedded system. The accuracy values are between 0 and 1 and the total time in each timestamp is calculated by adding the inference time of each sensor. 

Tables \ref{tab:slip_cnn_results} and \ref{tab:slip_brightness_results} show that detecting slippage by calculating the brightness of $\Psi$ images is more accurate and faster than using a CNN to classify these type of images. This is caused because the $filterImage$ function (from Algorithm \ref{slip_detection_algorithm}) converts the RGB tactile images into binary images, from which the CNN has more difficulties to extract features. 

The $T_{\mathrm{slip}}^{\mathrm{brightness}}$ threshold values of five and ten achieve the same accuracy value, but with $T_{\mathrm{slip}}^{\mathrm{brightness}}=5$, the model frequently detects false positives. We consequently set the brightness method with $T_{\mathrm{slip}}^{\mathrm{brightness}}=10$ as the slip detector.

\begin{table}[hpbt]
\begin{center}
\begin{minipage}{\textwidth}
\caption{Results obtained using CNN method for sensors A and B, and $T_{\mathrm{slip}}^{\mathrm{cnn}}$ = 0.4, 0.5, 0.6}
\label{tab:slip_cnn_results}
\begin{tabular*}{\textwidth}{@{\extracolsep{\fill}}ccccc@{\extracolsep{\fill}}}
\toprule%
\textbf{Sensor unit}     & \textbf{Acc (T=0.4)}  & \textbf{Acc (T=0.5)}   & \textbf{Acc (T=0.6)} & \textbf{Time(ms)}   \\
\midrule
\textbf{A} & 0.733  & 0.733   & 0.689  & 157.5  \\
\textbf{B} & 0.689  & 0.689   & 0.644    & 157.5   \\
\botrule
\end{tabular*}
\end{minipage}
\end{center}
\end{table}

\begin{table}[hpbt]
\begin{center}
\begin{minipage}{\textwidth}
\caption{Results obtained using brightness method for sensors A and B, and $T_{\mathrm{slip}}^{\mathrm{brightness}}$ = 5, 10, 15}
\label{tab:slip_brightness_results}
\begin{tabular*}{\textwidth}{@{\extracolsep{\fill}}ccccc@{\extracolsep{\fill}}}
\toprule%
\textbf{Sensor unit}      & \textbf{Acc (T=5)}  & \textbf{Acc (T=10)}    & \textbf{Acc (T=15)} & \textbf{Time(ms)}       \\
\midrule
\textbf{A} & 1 & 1 & 0.911 & 7.5  \\
\textbf{B} & 1 & 1 & 0.911 & 7.5      \\
\botrule
\end{tabular*}
\end{minipage}
\end{center}
\end{table}

\subsection{Detection and manipulation results in outdoor}
\label{sec:outdoor_results}

In this section, we show the experimentation that we carried out in order to test our system in three different outdoor environments with different objects with respect to the previous sections. Finally, we show the promising and reliable results obtained from this experimentation which prove that our system is able to perform the litter collection task. However, prior to running certain tasks in real mode, it is recommendable to simulate them in similar conditions. We specifically simulated the trajectory planning so as to visualise and check the robot arm and the gripper movement that is necessary in order to accomplish the robotic grasping task. We did this by building our mobile manipulation robot using ROS and RVIZ. Our simulation comprised the BLUE robot, the UR5e robot arm with the gripper and the RGBD camera mounted on the end effector, along with DIGIT sensors mounted on the fingertips of the gripper, as shown in Fig. \ref{fig:robot_simulation}.

\begin{figure}[!htb]
     \centering

         \includegraphics[width=0.7\textwidth, height=4.5cm]{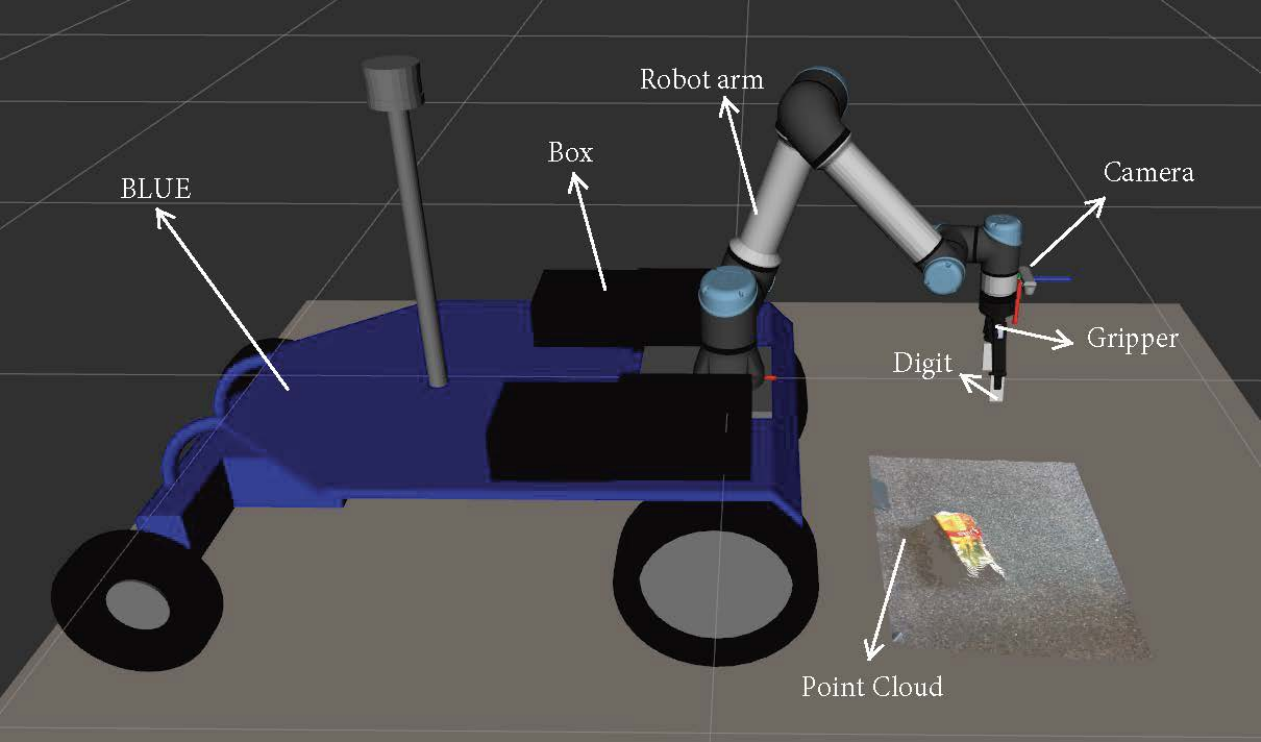}

     \caption{Robotic system that allows tasks to be performed in simulated environments}\label{fig:robot_simulation}
\end{figure}

To complete the outdoor experiment, we decided to use four new objects, one from each of the classes: cardboard, plastic, metal, or glass. These objects had not been used to form any of the previous visual or tactile datasets and were employed with the objective of testing the generalisation capabilities of our system with unknown objects for the waste collection task. Testing our system in different environments brought this experiment closer to reality. We, therefore, chose three new environments that were not been included in our previous datasets in order to test the generalisation of our proposal in unforeseen situations (see Fig. \ref{fig:outdoor_objects_environments}). 
\begin{figure}[!htb]
     \centering
     \begin{subfigure}[b]{0.24\textwidth}
         \centering
         \includegraphics[width=\textwidth, height=2cm]{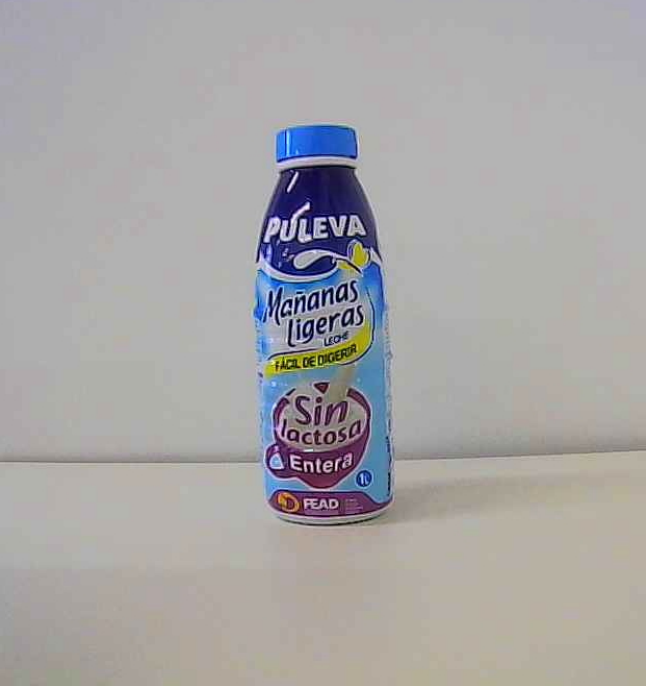}
         \caption{Plastic obj.}
     \end{subfigure}
     \begin{subfigure}[b]{0.24\textwidth}
         \centering
         \includegraphics[width=\textwidth, height=2cm]{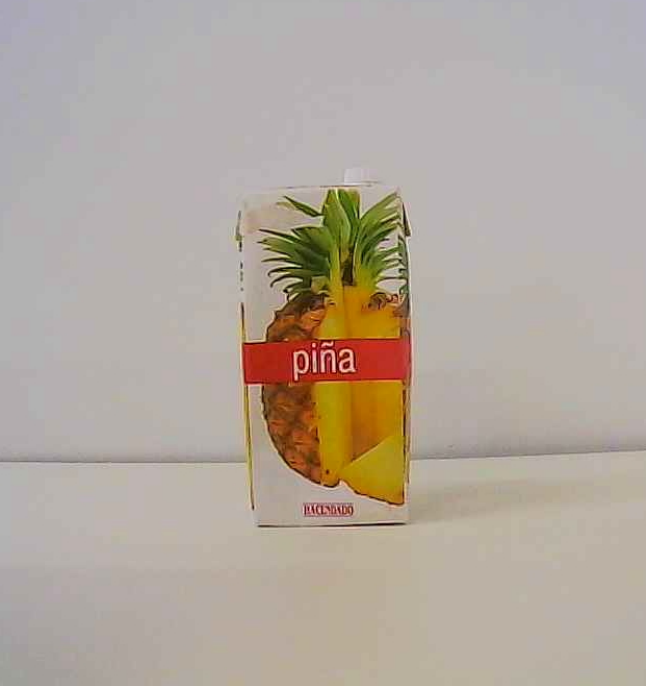}
         \caption{Cardboard obj.}
     \end{subfigure}
     \begin{subfigure}[b]{0.24\textwidth}
         \centering
         \includegraphics[width=\textwidth, height=2cm]{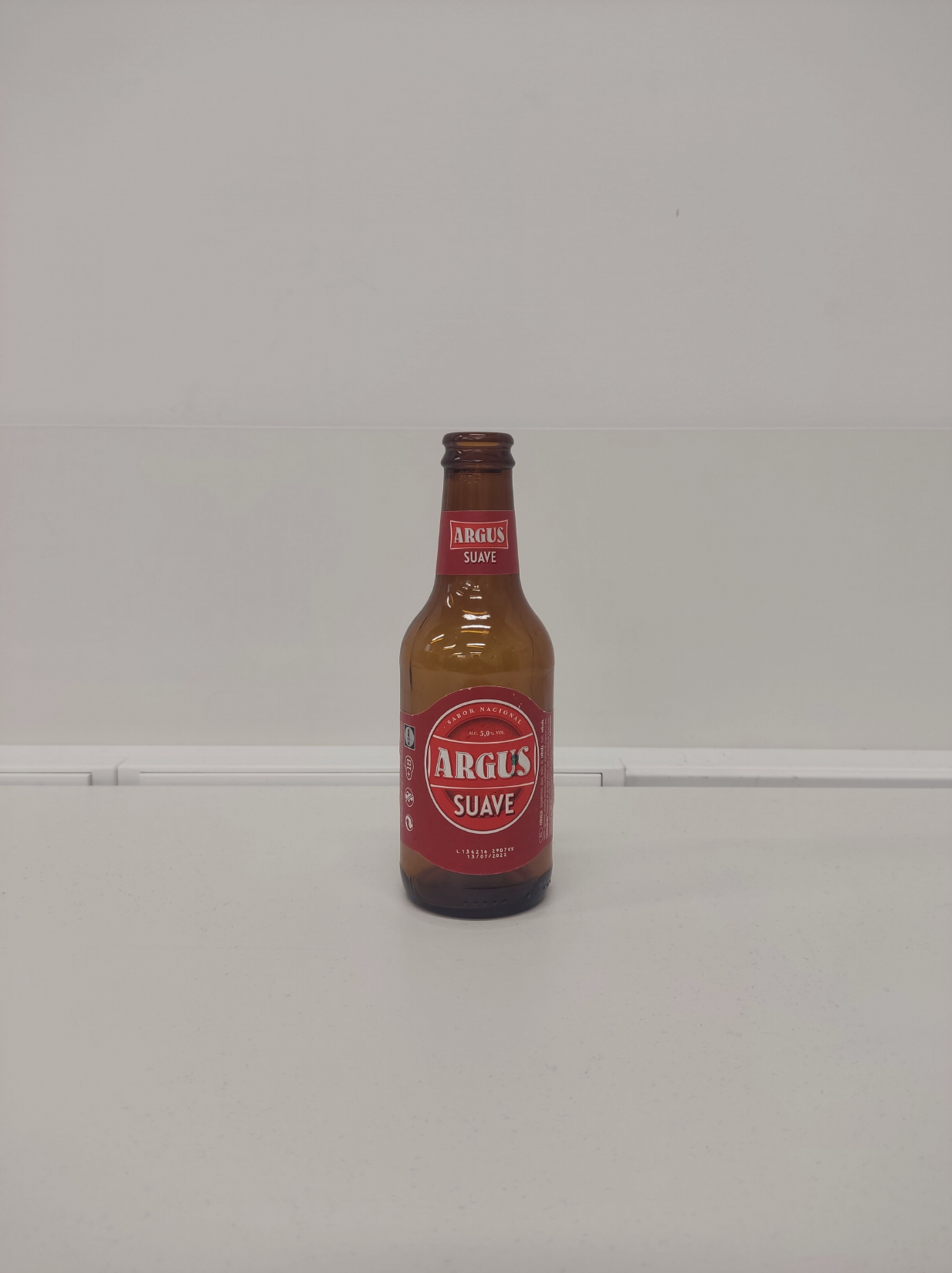}
         \caption{Glass obj.}
     \end{subfigure}
     \begin{subfigure}[b]{0.24\textwidth}
         \centering
         \includegraphics[width=\textwidth, height=2cm]{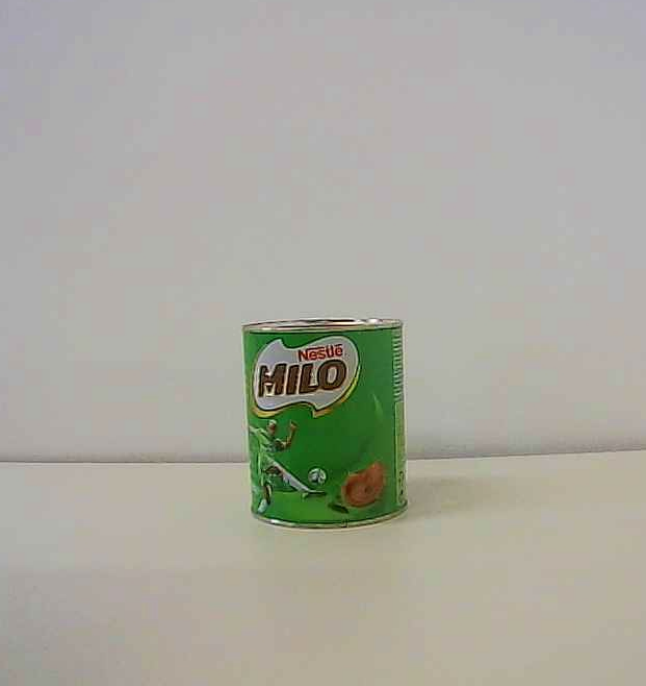}
         \caption{Metal obj.}
     \end{subfigure}
     
     \hspace{0.1cm}
     
     \begin{subfigure}[b]{0.24\textwidth}
         \centering
         \includegraphics[width=\textwidth]{FIG18-E.pdf}
         \caption{Tiled pavement}
     \end{subfigure}
     \begin{subfigure}[b]{0.24\textwidth}
         \centering
         \includegraphics[width=\textwidth]{FIG18-F.pdf}
         \caption{Grass}
     \end{subfigure}
     \begin{subfigure}[b]{0.24\textwidth}
         \centering
         \includegraphics[width=\textwidth]{FIG18-G.pdf}
         \caption{Tiled pavement}
     \end{subfigure}
     \begin{subfigure}[b]{0.24\textwidth}
         \centering
         \includegraphics[width=\textwidth]{FIG18-H.pdf}
         \caption{Stone/soil}
     \end{subfigure}
     
     \caption{Objects (one from each class) and environments (3 different environments) used to test  our robot in real outdoor experiments }\label{fig:outdoor_objects_environments}
\end{figure}

The following images show the results obtained for the  YOLACT detection and grasping points (Section \ref{sec:grasping}). Once the coordinates of these points have been referenced to the camera, it is necessary to obtain their global position. This position will be obtained from the robot base. Some of these coordinates are shown in Fig. \ref{fig:geograsp_points}.

\begin{figure}[htbp]
     \centerline{
         \includegraphics[width=.24\textwidth, height=.15\textwidth]{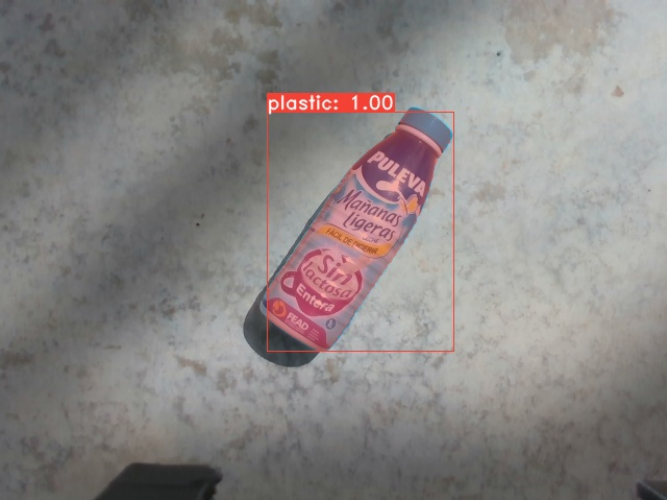}
         \includegraphics[width=.24\textwidth, height=.15\textwidth]{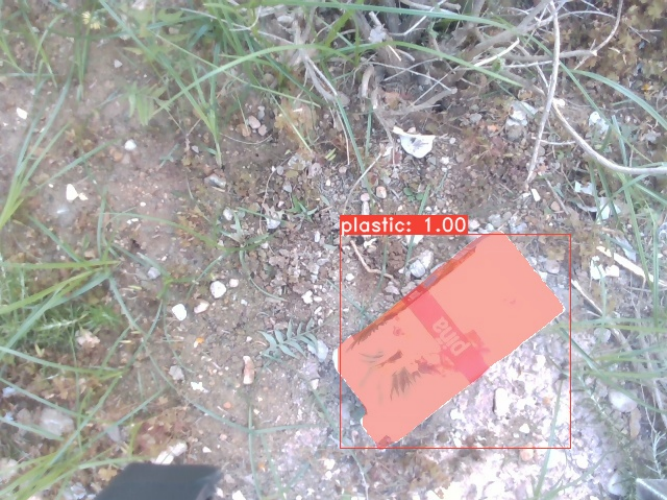}
         \includegraphics[width=.24\textwidth, height=.15\textwidth]{FIG19-3.pdf}
         \includegraphics[width=.24\textwidth, height=.15\textwidth]{FIG19-4.pdf}}
     \centerline{
         \includegraphics[width=.24\textwidth, height=.15\textwidth]{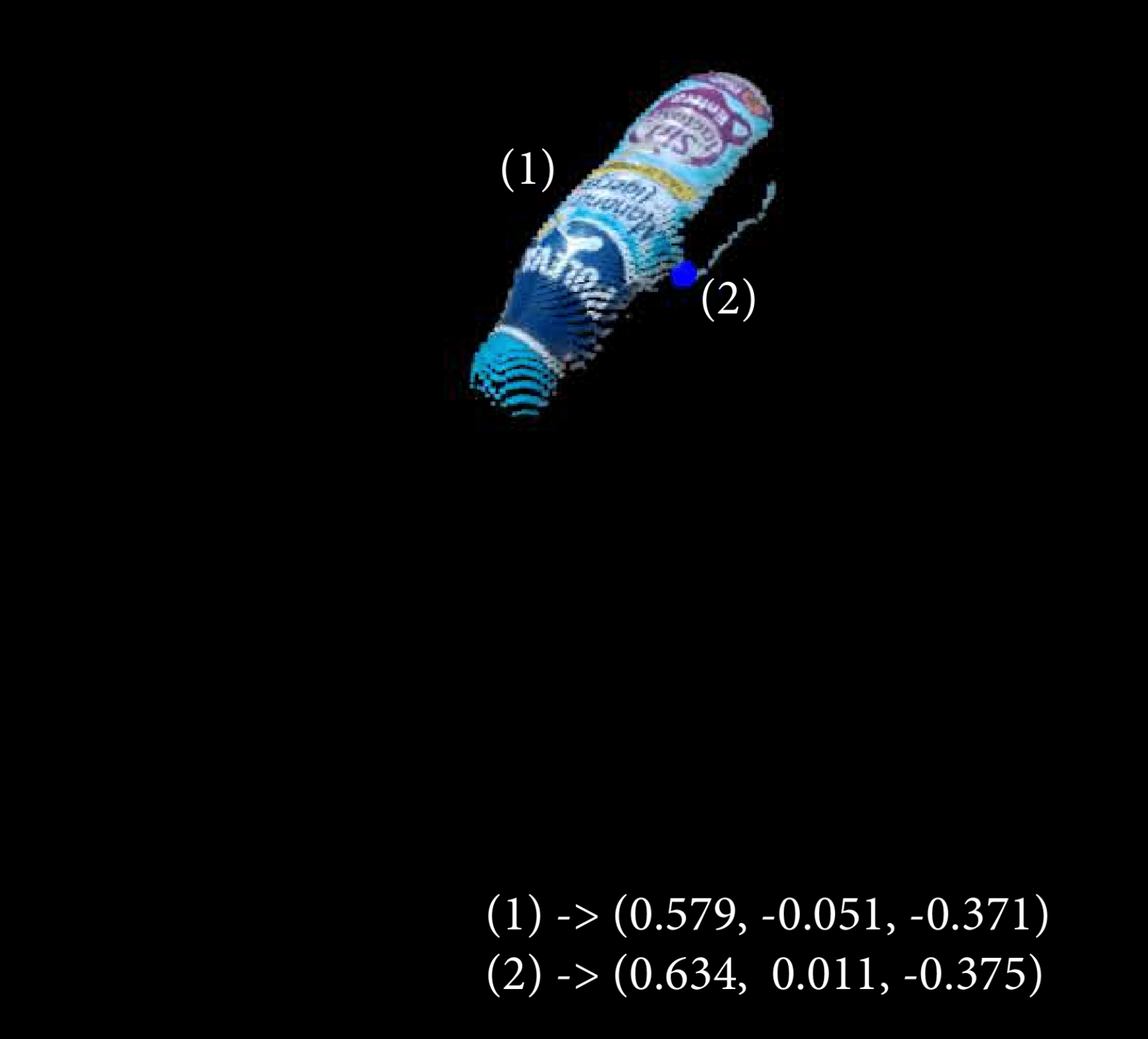}
         \includegraphics[width=.24\textwidth, height=.15\textwidth]{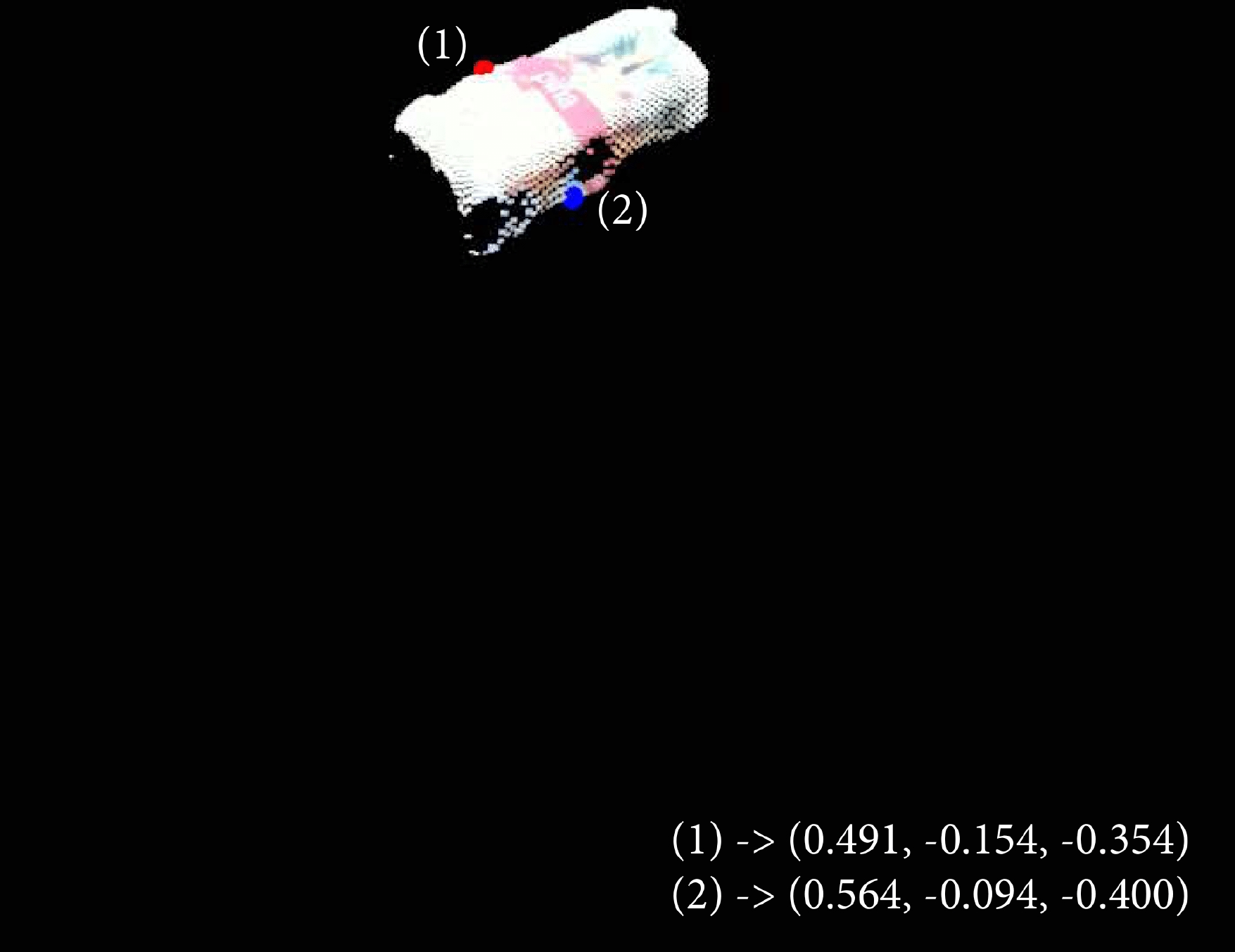}
         \includegraphics[width=.24\textwidth, height=.15\textwidth]{FIG19-7.pdf}
         \includegraphics[width=.24\textwidth, height=.15\textwidth]{FIG19-8.pdf}}
     \caption{Examples of YOLACT detection (top row) and grasping points found by GeoGrasp using robot (global) frame (bottom row) } 
     \label{fig:geograsp_points}
\end{figure}

Before carrying out the main outdoor experiment, two sub-experiments were carried out in order to establish the variables of the tactile manipulation module. The goal of the first sub-experiment was to demonstrate that sending a closing command through ROS to the gripper when slippage was detected would make the grasping more stable. In the second, we established a variable to denote the number of contact detections predicted by the model, in order to consider that the object had been grasped. Finally, in the main outdoor experiment, we evaluated our system by running our pipeline five times per object in each environment, namely, 60 object pick ups.

In order to demonstrate that grasping required compensation, we filled one object with a small quantity of water. The idea was to perform the grasping task, lift the object and evaluate the slip detections with and without grasping compensation. Figures \ref{fig:slip_detections_grasping_task} and \ref{fig:object_evolution_grasping_compensation} show that when grasping compensation is applied, only one slip event is detected and the object does not fall.  We did not use any compensation algorithm that the software of the gripper has, but instead, we created our own compensation algorithm. 

\begin{figure}[htbb]
     \centering
     \begin{subfigure}[b]{0.35\textwidth}
         \centering
         \includegraphics[width=\textwidth, height=3cm]{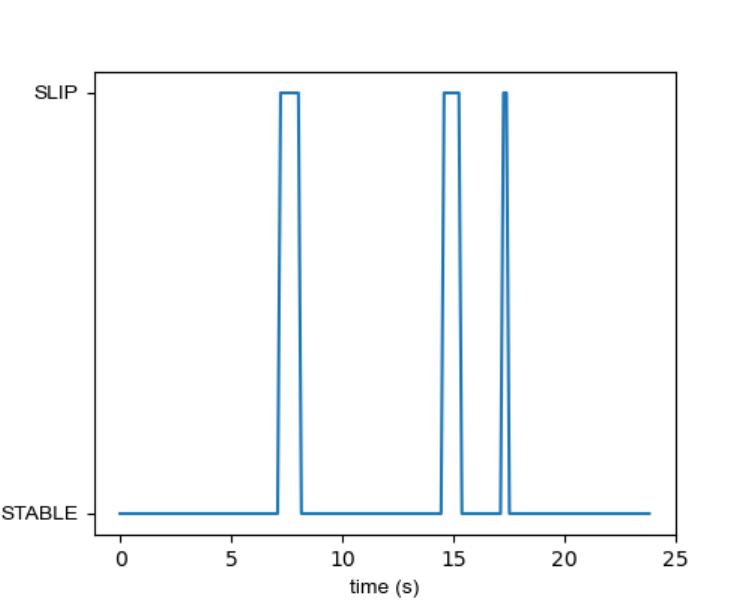}
        \caption{Without compensation}         
     \end{subfigure}
     \begin{subfigure}[b]{0.35\textwidth}
         \centering
         \includegraphics[width=\textwidth, height=3cm]{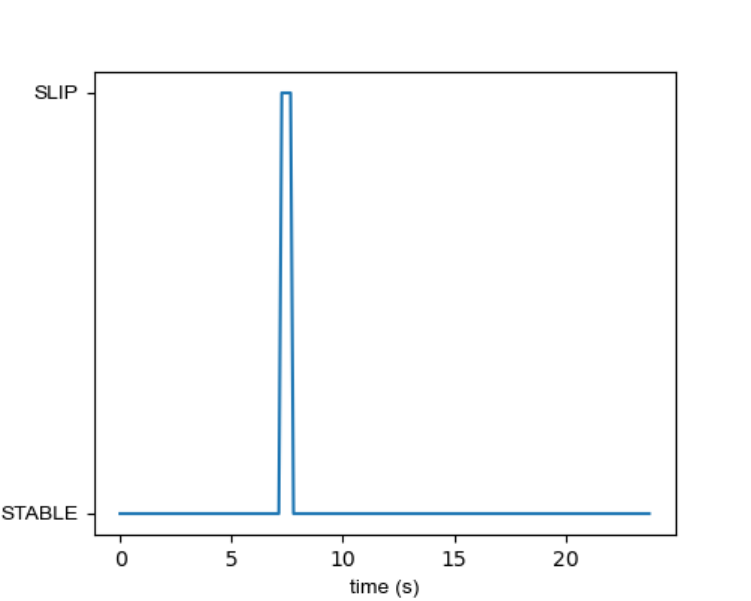}
         \caption{With compensation}
     \end{subfigure}
     \caption{Number of slip detections while the robot arm is lifting the object. This number is smaller when applying grasping compensation}\label{fig:slip_detections_grasping_task}
\end{figure}

\begin{figure}[!htb]
     \centering
     \begin{subfigure}[b]{0.24\textwidth}
         \centering
         \includegraphics[width=\textwidth]{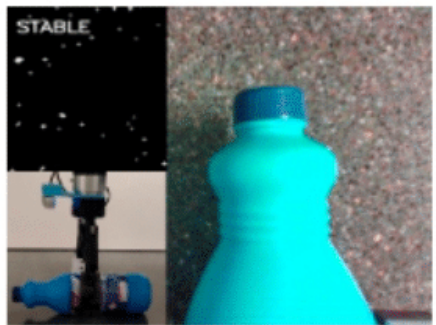}
         \caption{Time=0s}
     \end{subfigure}
     \begin{subfigure}[b]{0.24\textwidth}
         \centering
         \includegraphics[width=\textwidth]{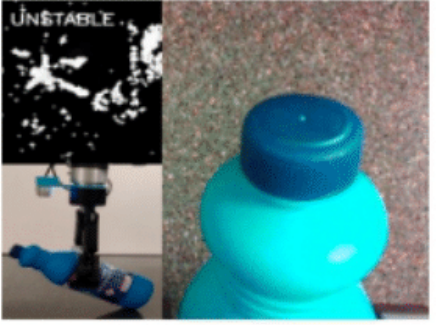}
         \caption{Time=7s}
     \end{subfigure}
     \begin{subfigure}[b]{0.24\textwidth}
         \centering
         \includegraphics[width=\textwidth]{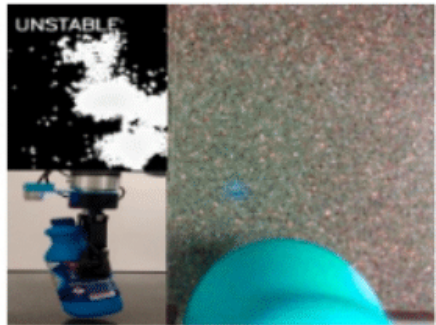}
         \caption{Time=15s}
     \end{subfigure}
     
     
     \begin{subfigure}[b]{0.24\textwidth}
         \centering
         \includegraphics[width=\textwidth]{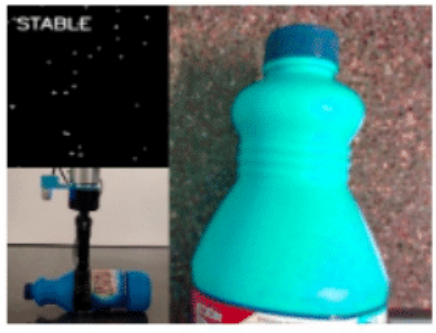}
         \caption{Time=0s}
     \end{subfigure}
     \begin{subfigure}[b]{0.24\textwidth}
         \centering
         \includegraphics[width=\textwidth]{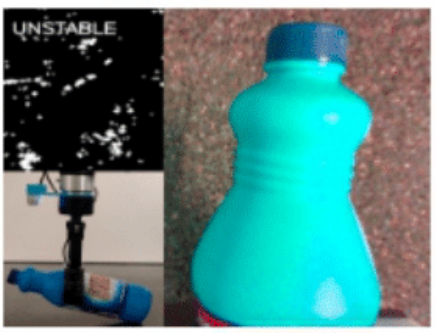}
         \caption{Time=7s}
     \end{subfigure}
     \begin{subfigure}[b]{0.24\textwidth}
         \centering
         \includegraphics[width=\textwidth]{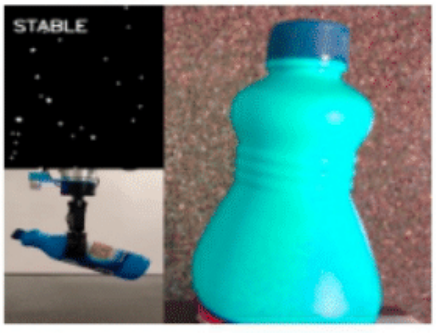}
         \caption{Time=15s}
     \end{subfigure}

      \caption{(a, b, c) Object falling without grasping compensation. Slip state is produced and detected (unstable). (d, e, f) Object does not fall with grasping compensation. Slip state is detected and compensated by closing the gripper}\label{fig:object_evolution_grasping_compensation}
\end{figure}

In order to lift the object, our system first needs to know whether or not the object is being grasped. An object is considered to be grasped when a certain number of contact detections are completed. We, therefore, ran the entire perception and tactile system with each one of the four objects, using different numbers of contact detections. The decision as to which number was the most suitable for each object was made by calculating the same graphics as those shown in Fig. \ref{fig:slip_detections_grasping_task}. Figure \ref{fig:second_experimentation_outdoor} shows that a stable grasp can be achieved for the four objects with a threshold of 3 contact detections for the objects of cardboard, plastic and metal, and 4 contact detections for the glass object because they usually weigh more.

\begin{figure}[!htb]
     \centering
     \begin{subfigure}[b]{0.35\textwidth}
         \centering
         \includegraphics[width=\textwidth, height=3cm]{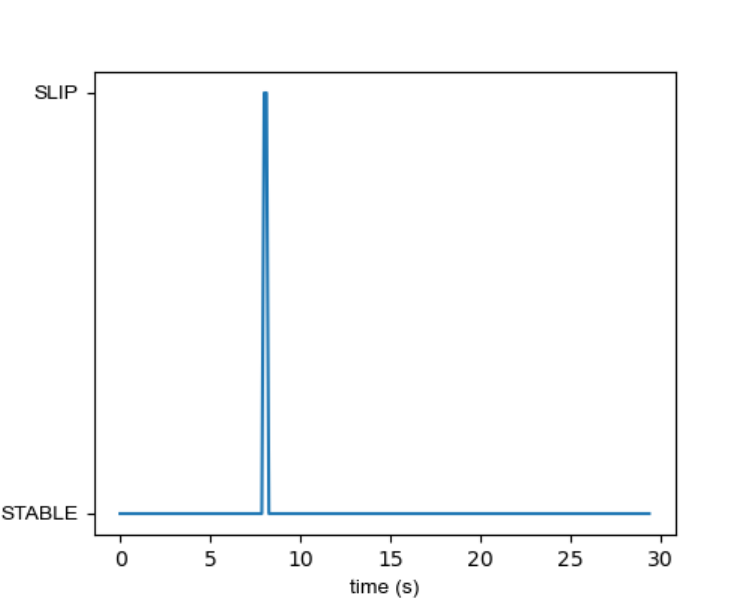}
         \caption{Cardboard}
     \end{subfigure}
     \begin{subfigure}[b]{0.35\textwidth}
         \centering
         \includegraphics[width=\textwidth, height=3cm]{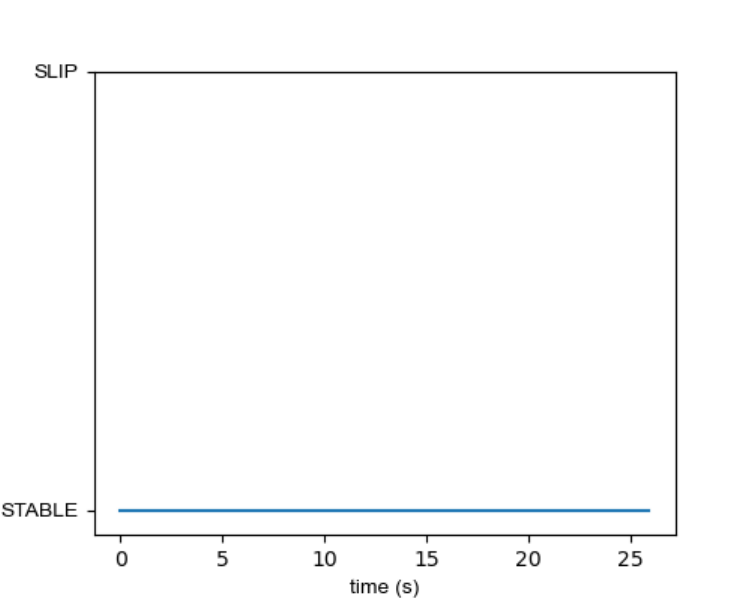}
        \caption{Plastic}
     \end{subfigure}
     
     \begin{subfigure}[b]{0.35\textwidth}
         \centering
         \includegraphics[width=\textwidth, height=3cm]{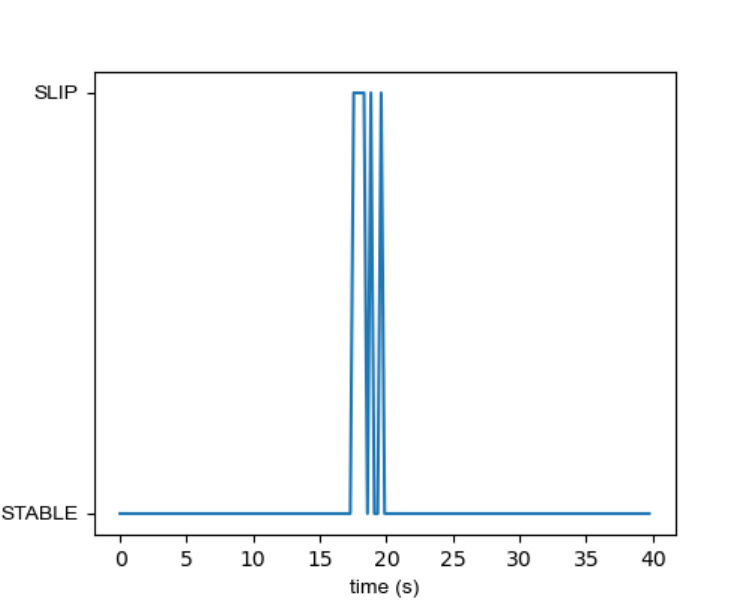}
         \caption{Metal}
     \end{subfigure}
     \begin{subfigure}[b]{0.35\textwidth}
         \centering
         \includegraphics[width=\textwidth, height=3cm]{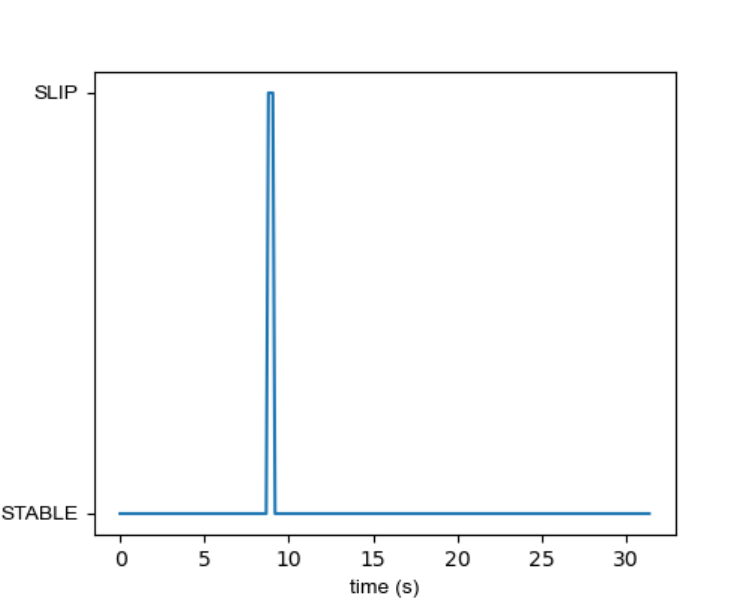}
         \caption{Glass}
     \end{subfigure}
     \caption{Slip detection graphic for each object. These plots show that slip is produced only once after compensating the grasping opening }\label{fig:second_experimentation_outdoor}
\end{figure}

Once the parameters had been established, the final experiment could be carried out. The results are expressed as the accuracy value that denotes the percentage of successful graspings, which we denominate as the Collection Success Rate (CSR). 

A grasping is considered successful when it achieves all of the following conditions: the object is segmented correctly, the grasping points are located on the object, the contact is detected and the robot is able to lift and move the object to the corresponding box, without dropping it. The success of collecting an object at the first attempt is 80\%. However, a failed first attempt can be successful in the second one since grasps are independent probabilistic events. That is, the pipeline is launched again if needed and the object could have changed its position when falling to the floor on the previous attempt.

As can be seen in Table \ref{tab:results_environments}, the overall collecting success rate varied from 75\% in the worst scenario to 85\% in the best one. Our system works better on flat surfaces (tiled pavements) or on surfaces that the gripper can cross in order to grasp the object (grass). Nonetheless, the system also attains promising results on irregular surfaces on which the gripper could collide with the ground. Good results are also obtained when using the same coloured object and environment (metal object and grass environment).

\begin{table}[hpbt]
\begin{center}
\begin{minipage}{\textwidth}
\caption{CSR values obtained splitting the results in terms of the environment (CSR-Env). These values are between 0 and 1 }
\label{tab:results_environments}
\begin{tabular*}{\textwidth}{@{\extracolsep{\fill}}cccc@{\extracolsep{\fill}}}
\toprule%
   & \textbf{Tiled pavement}  & \textbf{Surface of stone/soil}    & \textbf{Grass}                                    \\ 
\midrule
\textbf{CSR-Env} & 0.80 & 0.75 & 0.85  \\
\botrule
\end{tabular*}
\end{minipage}
\end{center}
\end{table}

With regard to the results in terms of object class (see Table \ref{tab:results_objects}), our system attained a CSR of 60\% in the worst case and 93\% in the best one, when picking up the objects at the first attempt. The drop in the CSR that occurred with glass objects was owing to their transparency, which makes object segmentation and grasping point calculation more difficult. The grasping pose is not, therefore, precise and the gripper fails when picking up the object. 

\begin{table}[hpbt]
\begin{center}
\begin{minipage}{\textwidth}
\caption{CSR values obtained after splitting the results in terms of the object class (CSR-Obj). These values are between 0 and 1 }
\label{tab:results_objects}
\begin{tabular*}{\textwidth}{@{\extracolsep{\fill}}ccccc@{\extracolsep{\fill}}}
\toprule%
   & \textbf{Cardboard}  & \textbf{Plastic}    & \textbf{Metal} & \textbf{Glass}                                    \\ 
\midrule
\textbf{CSR-Obj} & 0.93 & 0.8 & 0.87 & 0.6 \\
\botrule
\end{tabular*}
\end{minipage}
\end{center}
\end{table}

Figure \ref{fig:grasping_example} shows each step of an example of one litter collection attempt with a cardboard object in a tiled pavement environment. In this example our system detects and locates the item of litter as cardboard class, calculates the grasping pose correctly, and performs the manipulation tasks in order to collect the litter. 

The most common errors are produced as a result of the wrong locations of the grasping points, failed contact detections, low quality of the point clouds produced by the RGBD camera, or wrong object detection and segmentation. Our system failed at the first attempt on 18\% of occasions. Figure \ref{fig:distribution_error_types}-a shows the distribution of error types that led to the collection failure. Figure \ref{fig:distribution_error_types}-b shows the CSR of each module independently.

\begin{figure}[!htb]
     \centering
     \begin{subfigure}[b]{0.4\textwidth}
         \centering
         \includegraphics[width=\textwidth, height=3cm]{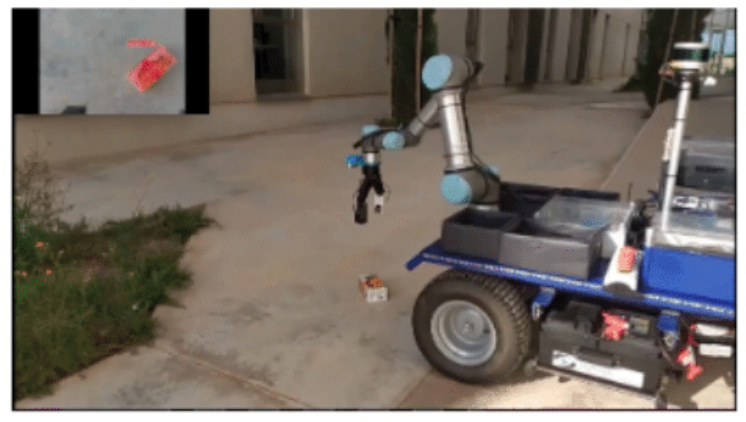}
         \caption{Litter detection}
     \end{subfigure}
     \begin{subfigure}[b]{0.4\textwidth}
         \centering
         \includegraphics[width=\textwidth, height=3cm]{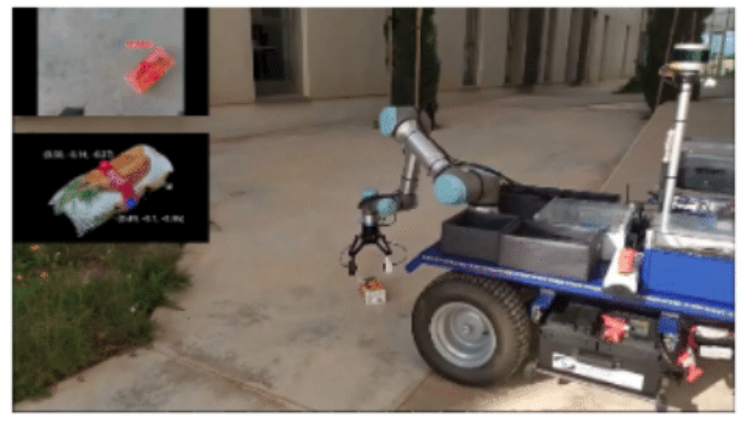}
        \caption{Grasping points calculation}
     \end{subfigure}
     
     \begin{subfigure}[b]{0.4\textwidth}
         \centering
         \includegraphics[width=\textwidth, height=3cm]{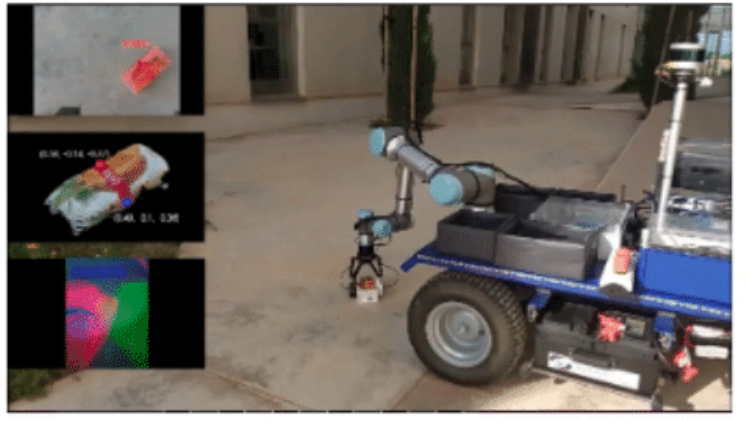}
         \caption{Contact state detection}
     \end{subfigure}
     \begin{subfigure}[b]{0.4\textwidth}
         \centering
         \includegraphics[width=\textwidth, height=3cm]{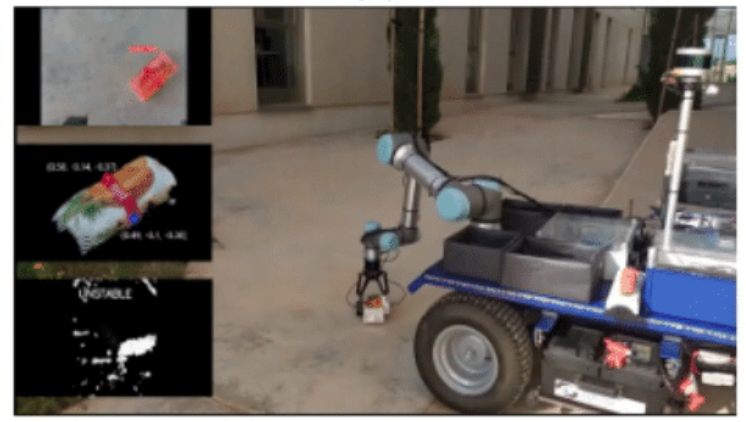}
         \caption{Slip state detection}
     \end{subfigure}
     \caption{Example of picking up a cardboard object in a tiled pavement environment}\label{fig:grasping_example}
\end{figure}

\begin{figure}[htpb]
     \centering
     \begin{subfigure}[b]{0.4\textwidth}
         \centering
         \includegraphics[width=\textwidth]{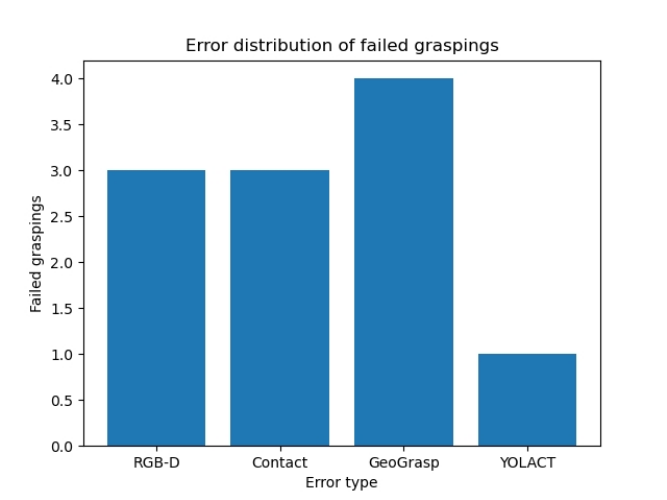}
     \caption{}
     \end{subfigure}
     \begin{subfigure}[b]{0.4\textwidth}
         \centering
         \includegraphics[width=\textwidth]{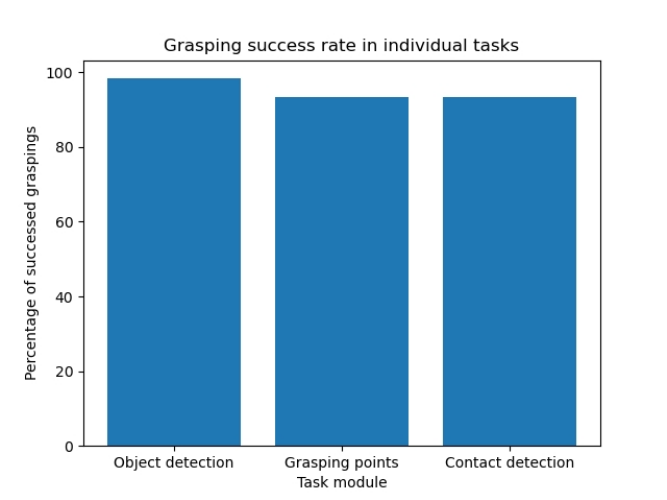}
         \caption{}
     \end{subfigure}
     \caption{a) Error type distribution that led to picking up failures. Error types are 1) bad reconstruction from the RGBD camera, 2) failed contact detection, 3) wrong location of the grasping points, and 4) wrong object detection. b) CSR when considering only the errors produced by each individual module}\label{fig:distribution_error_types}
\end{figure}

\section{Discussion and conclusions}
\label{sec:conclusions}

In this work, we present a robotic system that is able to collect household objects in outdoor and natural environments. This system was created by implementing three main modules: litter detection, grasping point calculation, and tactile manipulation. These three modules together allow our system to obtain promising and reliable results with unknown objects and in outdoor scenarios.


We have implemented two detection and control methods: one for contact detection, which is based on CNN, in order to know when the object is being grasped, and another for the slip detection, which is based on morphological operations, in order to readjust the gripper opening. Moreover, we have used well-known neural networks for the recognition and segmentation of objects. The result is then combined with 3D techniques to extract the surface features of those objects. Also, we have carried out extensive and rigorous experimentation in order to adjust our visual model to different objects in different scenarios. We first carried out offline experimentation (Section \ref{sec:training_visual_perception}, \ref{sec:training_contact_detection} and \ref{sec:training_slip_detection}) with real recorded data, after which we tested our system in online mode with real-time data (Section \ref{sec:outdoor_results}). When all the modules work together, our system is able to perform the complex task of picking up household objects in outdoor scenarios by combining the information concerning each module. This is assuming that the biggest object that the system could handle is 118 $mm$, in which we could manage a rotation error up to 13º with respect to the $x_{\mathrm{C}}$ reference frame system of Fig. \ref{fig:blue_poses} - right and a translation error of 3\% with respect to the size of the object with respect to the $y_{\mathrm{C}}$ and $z_{\mathrm{C}}$ reference frame systems of Fig. \ref{fig:blue_poses} - right (approx 3.5 mm).

Overall, our system performs correctly when collecting unknown objects in new environments, obtaining high accuracy when making more than one attempt. Nonetheless, our system has some limitations that we present as future lines of work. For example, transparent objects cause the detection and grasping point calculation modules several problems resulting from a bad quality point cloud obtained by the RGBD camera. This issue could be solved by a complementary object reconstruction process. Another limitation is the detection of dynamic objects, namely, an object whose position or orientation changes while the detection is taking place, for example, due to the wind. We could solve this by performing a tracking operation in order to update the object pose in real time. We are currently resolving this issue by executing the waste detection in a loop in order to update the object's pose.


\bmhead{Supplementary information}

Authors reporting more examples of robotic grasps for other litter objects in \href{Video 1}{https://www.youtube.com/watch?v=jME3Ozsz6iE&ab_channel=AUROVA\%3AAutomatics\%2CRobotics\%26Vision} and \href{Video 2}{https://www.youtube.com/watch?v=X7m_dAOGP78&ab_channel=AUROVA\%3AAutomatics\%2CRobotics\%26Vision}.

\bmhead{Acknowledgments}

This research was funded by the Valencian Regional Government through the PROMETEO/2021/075 project. The computer facilities were provided by the Valencian Government and FEDER through the IDIFEFER/2020/003 project.


\section*{Declarations}
\begin{itemize}
    \item \textbf{Funding} Research work was funded by the Valencian Regional Government and FEDER through the PROMETEO/2021/075 project. The computer facilities were provided through the IDIFEFER/2020/003 project.
    \item \textbf{Conflict of interest} The authors declare that they have no known competing finantial interests or personal relationships that could have appeared to influence the work reported in this paper.
    \item \textbf{Ethics approval} Not applicable.
    \item \textbf{Consent to participate} Not applicable.
    \item \textbf{Consent for publication} Authors gives the Publisher the permission to publish the Work in this journal.
    \item \textbf{Availability of data and materials} The dataset D1 designed and used is publicly available at \href{HOWA_dataset}{https://www.iuii.ua.es/datasets/howa/index.html}. Datasets D2 and D3 designed and used are not publicly available in this moment but they could be available from Institutional Repository of the University of Alicante or asking authors.
    \item \textbf{Code availability} Not applicable.
    \item \textbf{Author's contributions} I.L.P and S.T.P contributed to the design and implementation of the vision system. J.C.A and P.G contributed to the design and implementation of the tactile system. I.L.P and J.C.A carried out the experiments in real environments. All authors analysed the results and wrote the manuscript.
\end{itemize}





\bigskip\noindent

\bibliography{sn-bibliography}


\begin{thebibliography}{47}
\ifx \bisbn   \undefined \def \bisbn  #1{ISBN #1}\fi
\ifx \binits  \undefined \def \binits#1{#1}\fi
\ifx \bauthor  \undefined \def \bauthor#1{#1}\fi
\ifx \batitle  \undefined \def \batitle#1{#1}\fi
\ifx \bjtitle  \undefined \def \bjtitle#1{#1}\fi
\ifx \bvolume  \undefined \def \bvolume#1{\textbf{#1}}\fi
\ifx \byear  \undefined \def \byear#1{#1}\fi
\ifx \bissue  \undefined \def \bissue#1{#1}\fi
\ifx \bfpage  \undefined \def \bfpage#1{#1}\fi
\ifx \blpage  \undefined \def \blpage #1{#1}\fi
\ifx \burl  \undefined \def \burl#1{\textsf{#1}}\fi
\ifx \doiurl  \undefined \def \doiurl#1{\url{https://doi.org/#1}}\fi
\ifx \betal  \undefined \def \betal{\textit{et al.}}\fi
\ifx \binstitute  \undefined \def \binstitute#1{#1}\fi
\ifx \binstitutionaled  \undefined \def \binstitutionaled#1{#1}\fi
\ifx \bctitle  \undefined \def \bctitle#1{#1}\fi
\ifx \beditor  \undefined \def \beditor#1{#1}\fi
\ifx \bpublisher  \undefined \def \bpublisher#1{#1}\fi
\ifx \bbtitle  \undefined \def \bbtitle#1{#1}\fi
\ifx \bedition  \undefined \def \bedition#1{#1}\fi
\ifx \bseriesno  \undefined \def \bseriesno#1{#1}\fi
\ifx \blocation  \undefined \def \blocation#1{#1}\fi
\ifx \bsertitle  \undefined \def \bsertitle#1{#1}\fi
\ifx \bsnm \undefined \def \bsnm#1{#1}\fi
\ifx \bsuffix \undefined \def \bsuffix#1{#1}\fi
\ifx \bparticle \undefined \def \bparticle#1{#1}\fi
\ifx \barticle \undefined \def \barticle#1{#1}\fi
\bibcommenthead
\ifx \bconfdate \undefined \def \bconfdate #1{#1}\fi
\ifx \botherref \undefined \def \botherref #1{#1}\fi
\ifx \url \undefined \def \url#1{\textsf{#1}}\fi
\ifx \bchapter \undefined \def \bchapter#1{#1}\fi
\ifx \bbook \undefined \def \bbook#1{#1}\fi
\ifx \bcomment \undefined \def \bcomment#1{#1}\fi
\ifx \oauthor \undefined \def \oauthor#1{#1}\fi
\ifx \citeauthoryear \undefined \def \citeauthoryear#1{#1}\fi
\ifx \endbibitem  \undefined \def \endbibitem {}\fi
\ifx \bconflocation  \undefined \def \bconflocation#1{#1}\fi
\ifx \arxivurl  \undefined \def \arxivurl#1{\textsf{#1}}\fi
\csname PreBibitemsHook\endcsname

\bibitem{chiang2015vision}
\begin{bchapter}
\bauthor{\bsnm{Chiang}, \binits{C.-H.}}:
\bctitle{Vision-based coverage navigation for robot trash collection task}.
In: \bbtitle{IEEE Int. Conf. on Advanced Robotics and Intelligent Systems (ARIS)},
pp. \bfpage{1}--\blpage{6}
(\byear{2015}).
\doiurl{10.1109/ARIS.2015.7158229}.
\bcomment{IEEE}
\end{bchapter}
\endbibitem

\bibitem{muthugala2020tradeoff}
\begin{barticle}
\bauthor{\bsnm{Muthugala}, \binits{M.V.J.}},
\bauthor{\bsnm{Samarakoon}, \binits{S.B.P.}},
\bauthor{\bsnm{Elara}, \binits{M.R.}}:
\batitle{Tradeoff between area coverage and energy usage of a self-reconfigurable floor cleaning robot based on user preference}.
\bjtitle{IEEE Access}
\bvolume{8},
\bfpage{76267}--\blpage{76275}
(\byear{2020}).
\doiurl{10.1109/ACCESS.2020.2988977}
\end{barticle}
\endbibitem

\bibitem{zapata2018autotrans}
\begin{barticle}
\bauthor{\bsnm{Zapata-Impata}, \binits{B.S.}},
\bauthor{\bsnm{Shah}, \binits{V.}},
\bauthor{\bsnm{Singh}, \binits{H.}},
\bauthor{\bsnm{Platt}, \binits{R.}}:
\batitle{Autotrans: an autonomous open world transportation system}.
\bjtitle{arXiv preprint arXiv:1810.03400}
(\byear{2018}).
\doiurl{10.48550/arXiv.1810.03400}
\end{barticle}
\endbibitem

\bibitem{pmlr-v164-sun22a}
\begin{bchapter}
\bauthor{\bsnm{Sun}, \binits{C.}},
\bauthor{\bsnm{Orbik}, \binits{J.}},
\bauthor{\bsnm{Devin}, \binits{C.M.}},
\bauthor{\bsnm{Yang}, \binits{B.H.}},
\bauthor{\bsnm{Gupta}, \binits{A.}},
\bauthor{\bsnm{Berseth}, \binits{G.}},
\bauthor{\bsnm{Levine}, \binits{S.}}:
\bctitle{Fully autonomous real-world reinforcement learning with applications to mobile manipulation}.
In: \bbtitle{5th Conf. on Robot Learning (CoRL)}
(\byear{2021}).
\doiurl{10.48550/arXiv.2107.13545}
\end{bchapter}
\endbibitem

\bibitem{sultana2020trash}
\begin{bchapter}
\bauthor{\bsnm{Sultana}, \binits{R.}},
\bauthor{\bsnm{Adams}, \binits{R.D.}},
\bauthor{\bsnm{Yan}, \binits{Y.}},
\bauthor{\bsnm{Yanik}, \binits{P.M.}},
\bauthor{\bsnm{Tanaka}, \binits{M.L.}}:
\bctitle{Trash and recycled material identification using convolutional neural networks (cnn)}.
In: \bbtitle{SoutheastCon},
pp. \bfpage{1}--\blpage{8}
(\byear{2020}).
\doiurl{10.1109/SoutheastCon44009.2020.9249739}
\end{bchapter}
\endbibitem

\bibitem{lin2015robot}
\begin{barticle}
\bauthor{\bsnm{Lin}, \binits{Y.}},
\bauthor{\bsnm{Sun}, \binits{Y.}}:
\batitle{Robot grasp planning based on demonstrated grasp strategies}.
\bjtitle{The International Journal of Robotics Research}
\bvolume{34}(\bissue{1}),
\bfpage{26}--\blpage{42}
(\byear{2015}).
\doiurl{10.1177\%2F0278364914555544}
\end{barticle}
\endbibitem

\bibitem{article}
\begin{botherref}
\oauthor{\bsnm{Zapata-Impata}, \binits{B.}},
\oauthor{\bsnm{Gil}, \binits{P.}},
\oauthor{\bsnm{Pomares}, \binits{J.}},
\oauthor{\bsnm{Medina}, \binits{F.}}:
Fast geometry-based computation of grasping points on three-dimensional point clouds.
International Journal of Advanced Robotic Systems
\textbf{16}
(2019).
\doiurl{10.1177/1729881419831846}
\end{botherref}
\endbibitem

\bibitem{delPino2019}
\begin{barticle}
\bauthor{\bparticle{del} \bsnm{Pino}, \binits{I.}},
\bauthor{\bsnm{Mu{\~{n}}oz-Ba{\~{n}}on}, \binits{M.{\'A}.}},
\bauthor{\bsnm{Cova-Rocamora}, \binits{S.}},
\bauthor{\bsnm{Contreras}, \binits{M.{\'A}.}},
\bauthor{\bsnm{Candelas}, \binits{F.A.}},
\bauthor{\bsnm{Torres}, \binits{F.}}:
\batitle{Deeper in blue}.
\bjtitle{Journal of Intelligent {\&} Robotic Systems}
\bvolume{98},
\bfpage{207}--\blpage{225}
(\byear{2020}).
\doiurl{10.1007/s10846-019-00983-6}
\end{barticle}
\endbibitem

\bibitem{Lambeta2020}
\begin{barticle}
\bauthor{\bsnm{Lambeta}, \binits{M.}},
\bauthor{\bsnm{Chou}, \binits{P.-W.}},
\bauthor{\bsnm{Tian}, \binits{S.}},
\bauthor{\bsnm{Yang}, \binits{B.}},
\bauthor{\bsnm{Maloon}, \binits{B.}},
\bauthor{\bsnm{Most}, \binits{V.R.}},
\bauthor{\bsnm{Stroud}, \binits{D.}},
\bauthor{\bsnm{Santos}, \binits{R.}},
\bauthor{\bsnm{Byagowi}, \binits{A.}},
\bauthor{\bsnm{Kammerer}, \binits{G.}},
\bauthor{\bsnm{Jayaraman}, \binits{D.}},
\bauthor{\bsnm{Calandra}, \binits{R.}}:
\batitle{Digit: A novel design for a low-cost compact high-resolution tactile sensor with application to in-hand manipulation}.
\bjtitle{IEEE Robotics and Automation Letters}
\bvolume{5}(\bissue{3}),
\bfpage{3838}--\blpage{3845}
(\byear{2020}).
\doiurl{10.1109/LRA.2020.2977257}
\end{barticle}
\endbibitem

\bibitem{chandra2021review}
\begin{bchapter}
\bauthor{\bsnm{Chandra}, \binits{S.S.}},
\bauthor{\bsnm{Kulshreshtha}, \binits{M.}},
\bauthor{\bsnm{Randhawa}, \binits{P.}}:
\bctitle{A review of trash collecting and cleaning robots}.
In: \bbtitle{9th Int. Conf. on Reliability, Infocom Technologies and Optimization (Trends and Future Directions) (ICRITO)},
pp. \bfpage{1}--\blpage{5}
(\byear{2021}).
\doiurl{10.1109/ICRITO51393.2021.9596551}
\end{bchapter}
\endbibitem

\bibitem{bai2018deep}
\begin{barticle}
\bauthor{\bsnm{Bai}, \binits{J.}},
\bauthor{\bsnm{Lian}, \binits{S.}},
\bauthor{\bsnm{Liu}, \binits{Z.}},
\bauthor{\bsnm{Wang}, \binits{K.}},
\bauthor{\bsnm{Liu}, \binits{D.}}:
\batitle{Deep learning based robot for automatically picking up garbage on the grass}.
\bjtitle{IEEE Transactions on Consumer Electronics}
\bvolume{64}(\bissue{3}),
\bfpage{382}--\blpage{389}
(\byear{2018}).
\doiurl{10.1109/TCE.2018.2859629}
\end{barticle}
\endbibitem

\bibitem{liu2021garbage}
\begin{bchapter}
\bauthor{\bsnm{Liu}, \binits{J.}},
\bauthor{\bsnm{Balatti}, \binits{P.}},
\bauthor{\bsnm{Ellis}, \binits{K.}},
\bauthor{\bsnm{Hadjivelichkov}, \binits{D.}},
\bauthor{\bsnm{Stoyanov}, \binits{D.}},
\bauthor{\bsnm{Ajoudani}, \binits{A.}},
\bauthor{\bsnm{Kanoulas}, \binits{D.}}:
\bctitle{Garbage collection and sorting with a mobile manipulator using deep learning and whole-body control}.
In: \bbtitle{IEEE 20th Int. Conf. on Humanoid Robots (Humanoids)},
pp. \bfpage{408}--\blpage{414}
(\byear{2021}).
\doiurl{10.1109/HUMANOIDS47582.2021.9555800}
\end{bchapter}
\endbibitem

\bibitem{mnyussiwalla2022evaluation}
\begin{barticle}
\bauthor{\bsnm{Mnyussiwalla}, \binits{H.}},
\bauthor{\bsnm{Seguin}, \binits{P.}},
\bauthor{\bsnm{Vulliez}, \binits{P.}},
\bauthor{\bsnm{Gazeau}, \binits{J.}}:
\batitle{Evaluation and selection of grasp quality criteria for dexterous manipulation}.
\bjtitle{Journal of Intelligent \& Robotic Systems}
\bvolume{104},
\bfpage{20}
(\byear{2022}).
\doiurl{10.1007/s10846-021-01554-4}
\end{barticle}
\endbibitem

\bibitem{ten2017grasp}
\begin{barticle}
\bauthor{\bparticle{ten} \bsnm{Pas}, \binits{A.}},
\bauthor{\bsnm{Gualtieri}, \binits{M.}},
\bauthor{\bsnm{Saenko}, \binits{K.}},
\bauthor{\bsnm{Platt}, \binits{R.}}:
\batitle{Grasp pose detection in point clouds}.
\bjtitle{The International Journal of Robotics Research}
\bvolume{36}(\bissue{13-14}),
\bfpage{1455}--\blpage{1473}
(\byear{2017}).
\doiurl{10.1177\%2F0278364917735594}
\end{barticle}
\endbibitem

\bibitem{dong2017improved}
\begin{bchapter}
\bauthor{\bsnm{Dong}, \binits{S.}},
\bauthor{\bsnm{Yuan}, \binits{W.}},
\bauthor{\bsnm{Adelson}, \binits{E.H.}}:
\bctitle{Improved gelsight tactile sensor for measuring geometry and slip}.
In: \bbtitle{IEEE/RSJ Int. Conf. on Intelligent Robots and Systems (IROS)},
pp. \bfpage{137}--\blpage{144}
(\byear{2017}).
\doiurl{10.1109/IROS.2017.8202149}
\end{bchapter}
\endbibitem

\bibitem{yuan2017gelsight}
\begin{barticle}
\bauthor{\bsnm{Yuan}, \binits{W.}},
\bauthor{\bsnm{Dong}, \binits{S.}},
\bauthor{\bsnm{Adelson}, \binits{E.H.}}:
\batitle{Gelsight: High-resolution robot tactile sensors for estimating geometry and force}.
\bjtitle{Sensors}
\bvolume{17}(\bissue{12}),
\bfpage{2762}
(\byear{2017}).
\doiurl{10.3390/s17122762}
\end{barticle}
\endbibitem

\bibitem{pmlr-v100-zhang20b}
\begin{bchapter}
\bauthor{\bsnm{Zhang}, \binits{Y.}},
\bauthor{\bsnm{Yuan}, \binits{W.}},
\bauthor{\bsnm{Kan}, \binits{Z.}},
\bauthor{\bsnm{Wang}, \binits{M.Y.}}:
\bctitle{Towards learning to detect and predict contact events on vision-based tactile sensors}.
In: \bbtitle{3rd Conf. on Robot Learning (CoRL)},
pp. \bfpage{1395}--\blpage{1404}
(\byear{2019}).
\doiurl{10.48550/arXiv.1910.03973}
\end{bchapter}
\endbibitem

\bibitem{zhang2018fingervision}
\begin{barticle}
\bauthor{\bsnm{Zhang}, \binits{Y.}},
\bauthor{\bsnm{Kan}, \binits{Z.}},
\bauthor{\bsnm{Tse}, \binits{Y.A.}},
\bauthor{\bsnm{Yang}, \binits{Y.}},
\bauthor{\bsnm{Wang}, \binits{M.Y.}}:
\batitle{Fingervision tactile sensor design and slip detection using convolutional lstm network}.
\bjtitle{arXiv preprint arXiv:1810.02653}
(\byear{2018}).
\doiurl{10.48550/arXiv.1810.02653}
\end{barticle}
\endbibitem

\bibitem{james2018slip}
\begin{barticle}
\bauthor{\bsnm{James}, \binits{J.W.}},
\bauthor{\bsnm{Pestell}, \binits{N.}},
\bauthor{\bsnm{Lepora}, \binits{N.F.}}:
\batitle{Slip detection with a biomimetic tactile sensor}.
\bjtitle{IEEE Robotics and Automation Letters}
\bvolume{3}(\bissue{4}),
\bfpage{3340}--\blpage{3346}
(\byear{2018}).
\doiurl{10.1109/LRA.2018.2852797}
\end{barticle}
\endbibitem

\bibitem{li2018slip}
\begin{bchapter}
\bauthor{\bsnm{Li}, \binits{J.}},
\bauthor{\bsnm{Dong}, \binits{S.}},
\bauthor{\bsnm{Adelson}, \binits{E.}}:
\bctitle{Slip detection with combined tactile and visual information}.
In: \bbtitle{IEEE Int. Conf. on Robotics and Automation (ICRA)},
pp. \bfpage{7772}--\blpage{7777}
(\byear{2018}).
\doiurl{10.1109/ICRA.2018.8460495}
\end{bchapter}
\endbibitem

\bibitem{sliptmo}
\begin{barticle}
\bauthor{\bsnm{James}, \binits{J.W.}},
\bauthor{\bsnm{Lepora}, \binits{N.F.}}:
\batitle{Slip detection for grasp stabilization with a multifingered tactile robot hand}.
\bjtitle{IEEE Transactions on Robotics}
\bvolume{37}(\bissue{2}),
\bfpage{506}--\blpage{519}
(\byear{2021}).
\doiurl{10.1109/TRO.2020.3031245}
\end{barticle}
\endbibitem

\bibitem{Tornero2022}
\begin{bchapter}
\bauthor{\bsnm{Tornero}, \binits{P.}},
\bauthor{\bsnm{Puente}, \binits{S.}},
\bauthor{\bsnm{Gil}, \binits{P.}}:
\bctitle{Detection and location of domestic waste for planning its collection using an autonomous robot}.
In: \bbtitle{IEEE 8th Int. Conf. on Control, Automation and Robotics (ICCAR)},
\bconflocation{Xiamen, China},
pp. \bfpage{138}--\blpage{144}
(\byear{2022}).
\doiurl{10.1109/ICCAR55106.2022.9782609}
\end{bchapter}
\endbibitem

\bibitem{hafiz2020survey}
\begin{barticle}
\bauthor{\bsnm{Hafiz}, \binits{A.M.}},
\bauthor{\bsnm{Bhat}, \binits{G.M.}}:
\batitle{A survey on instance segmentation: state of the art}.
\bjtitle{International Journal of Multimedia Information Retrieval}
\bvolume{9}(\bissue{3}),
\bfpage{171}--\blpage{189}
(\byear{2020}).
\doiurl{10.1007/s13735-020-00195-x}
\end{barticle}
\endbibitem

\bibitem{gu2022review}
\begin{barticle}
\bauthor{\bsnm{Gu}, \binits{W.}},
\bauthor{\bsnm{Bai}, \binits{S.}},
\bauthor{\bsnm{Kong}, \binits{L.}}:
\batitle{A review on 2d instance segmentation based on deep neural networks}.
\bjtitle{Image and Vision Computing}
\bvolume{120},
\bfpage{104401}
(\byear{2022}).
\doiurl{10.1016/j.imavis.2022.104401}
\end{barticle}
\endbibitem

\bibitem{He_2017_ICCV}
\begin{bchapter}
\bauthor{\bsnm{He}, \binits{K.}},
\bauthor{\bsnm{Gkioxari}, \binits{G.}},
\bauthor{\bsnm{Dollar}, \binits{P.}},
\bauthor{\bsnm{Girshick}, \binits{R.}}:
\bctitle{Mask r-cnn}.
In: \bbtitle{IEEE/CVF Int. Conf. on Computer Vision (ICCV)}
(\byear{2017}).
\doiurl{10.48550/arXiv.1703.06870}
\end{bchapter}
\endbibitem

\bibitem{yolact-iccv2019}
\begin{bchapter}
\bauthor{\bsnm{Bolya}, \binits{D.}},
\bauthor{\bsnm{Zhou}, \binits{C.}},
\bauthor{\bsnm{Xiao}, \binits{F.}},
\bauthor{\bsnm{Lee}, \binits{Y.J.}}:
\bctitle{Yolact: Real-time instance segmentation}.
In: \bbtitle{IEEE/CVF Int. Conf. on Computer Vision (ICCV)},
pp. \bfpage{9157}--\blpage{9166}
(\byear{2019}).
\doiurl{10.48550/arXiv.1904.02689}
\end{bchapter}
\endbibitem

\bibitem{yolact-plus-tpami2020}
\begin{barticle}
\bauthor{\bsnm{Bolya}, \binits{D.}},
\bauthor{\bsnm{Zhou}, \binits{C.}},
\bauthor{\bsnm{Xiao}, \binits{F.}},
\bauthor{\bsnm{Lee}, \binits{Y.J.}}:
\batitle{Yolact++ better real-time instance segmentation}.
\bjtitle{IEEE Transactions on Pattern Analysis and Machine Intelligence}
\bvolume{44}(\bissue{2}),
\bfpage{1108}--\blpage{1121}
(\byear{2022}).
\doiurl{10.1109/TPAMI.2020.3014297}
\end{barticle}
\endbibitem

\bibitem{DBLP:journals/corr/RenHG015}
\begin{barticle}
\bauthor{\bsnm{Ren}, \binits{S.}},
\bauthor{\bsnm{He}, \binits{K.}},
\bauthor{\bsnm{Girshick}, \binits{R.}},
\bauthor{\bsnm{Sun}, \binits{J.}}:
\batitle{Faster r-cnn: Towards real-time object detection with region proposal networks}.
\bjtitle{IEEE Transactions on Pattern Analysis and Machine Intelligence}
\bvolume{28}(\bissue{6}),
\bfpage{1137}--\blpage{1149}
(\byear{2015}).
\doiurl{10.1109/TPAMI.2016.2577031}
\end{barticle}
\endbibitem

\bibitem{He_2016_CVPR}
\begin{bchapter}
\bauthor{\bsnm{He}, \binits{K.}},
\bauthor{\bsnm{Zhang}, \binits{X.}},
\bauthor{\bsnm{Ren}, \binits{S.}},
\bauthor{\bsnm{Sun}, \binits{J.}}:
\bctitle{Deep residual learning for image recognition}.
In: \bbtitle{IEEE Conf. on Computer Vision and Pattern Recognition (CVPR)}
(\byear{2016}).
\doiurl{10.1109/CVPR.2016.90}
\end{bchapter}
\endbibitem

\bibitem{darknet13}
\begin{botherref}
\oauthor{\bsnm{Redmon}, \binits{J.}}:
Darknet: Open Source Neural Networks in C.
\url{http://pjreddie.com/darknet/}
(2013--2016)
\end{botherref}
\endbibitem

\bibitem{de2021domestic}
\begin{bchapter}
\bauthor{\bsnm{De~Gea}, \binits{V.}},
\bauthor{\bsnm{Puente}, \binits{S.T.}},
\bauthor{\bsnm{Gil}, \binits{P.}}:
\bctitle{Domestic waste detection and grasping points for robotic picking up}.
In: \bbtitle{IEEE Int. Conf. on Robotics and Automation (ICRA)-Workshop: Emerging Paradigms for Robotic Manipulation: from the Lab to the Productive World}
(\byear{2021}).
\doiurl{10.48550/arXiv.2105.06825}
\end{bchapter}
\endbibitem

\bibitem{garrido2014automatic}
\begin{barticle}
\bauthor{\bsnm{Garrido-Jurado}, \binits{S.}},
\bauthor{\bsnm{Mu{\~n}oz-Salinas}, \binits{R.}},
\bauthor{\bsnm{Madrid-Cuevas}, \binits{F.J.}},
\bauthor{\bsnm{Mar{\'\i}n-Jim{\'e}nez}, \binits{M.J.}}:
\batitle{Automatic generation and detection of highly reliable fiducial markers under occlusion}.
\bjtitle{Pattern Recognition}
\bvolume{47}(\bissue{6}),
\bfpage{2280}--\blpage{2292}
(\byear{2014}).
\doiurl{10.1016/j.patcog.2014.01.005}
\end{barticle}
\endbibitem

\bibitem{coleman_sucan_chitta_correll_2014}
\begin{barticle}
\bauthor{\bsnm{Coleman}, \binits{D.}},
\bauthor{\bsnm{Sucan}, \binits{I.}},
\bauthor{\bsnm{Chitta}, \binits{S.}},
\bauthor{\bsnm{Correll}, \binits{N.}}:
\batitle{Reducing the barrier to entry of complex robotic software: a moveit! case study}.
\bjtitle{Journal of Software Engineering for Robotics}
\bvolume{5}(\bissue{1}),
\bfpage{3}--\blpage{16}
(\byear{2014}).
\doiurl{10.48550/arXiv.1404.3785}
\end{barticle}
\endbibitem

\bibitem{lavalle1998rapidly}
\begin{botherref}
\oauthor{\bsnm{LaValle}, \binits{S.M.}}:
Rapidly-exploring random trees: A new tool for path planning.
Technical Report~11,
Computer Science Dept., Iowa State University
(October 1998).
\url{http://lavalle.pl/papers/Lav98c.pdf}
\end{botherref}
\endbibitem

\bibitem{lavalle2001randomized}
\begin{barticle}
\bauthor{\bsnm{LaValle}, \binits{S.M.}},
\bauthor{\bsnm{Kuffner~Jr}, \binits{J.J.}}:
\batitle{Randomized kinodynamic planning}.
\bjtitle{The International Journal of Robotics Research}
\bvolume{20}(\bissue{5}),
\bfpage{378}--\blpage{400}
(\byear{2001}).
\doiurl{10.1177/02783640122067453}
\end{barticle}
\endbibitem

\bibitem{ward2018tactip}
\begin{barticle}
\bauthor{\bsnm{Ward-Cherrier}, \binits{B.}},
\bauthor{\bsnm{Pestell}, \binits{N.}},
\bauthor{\bsnm{Cramphorn}, \binits{L.}},
\bauthor{\bsnm{Winstone}, \binits{B.}},
\bauthor{\bsnm{Giannaccini}, \binits{M.E.}},
\bauthor{\bsnm{Rossiter}, \binits{J.}},
\bauthor{\bsnm{Lepora}, \binits{N.F.}}:
\batitle{The tactip family: Soft optical tactile sensors with 3d-printed biomimetic morphologies}.
\bjtitle{Soft robotics}
\bvolume{5}(\bissue{2}),
\bfpage{216}--\blpage{227}
(\byear{2018}).
\doiurl{10.1089/soro.2017.0052}
\end{barticle}
\endbibitem

\bibitem{pagoli2022large}
\begin{barticle}
\bauthor{\bsnm{Pagoli}, \binits{A.}},
\bauthor{\bsnm{Chapelle}, \binits{F.}},
\bauthor{\bsnm{Corrales-Ramon}, \binits{J.-A.}},
\bauthor{\bsnm{Mezouar}, \binits{Y.}},
\bauthor{\bsnm{Lapusta}, \binits{Y.}}:
\batitle{Large-area and low-cost force/tactile capacitive sensor for soft robotic applications}.
\bjtitle{Sensors}
\bvolume{22}(\bissue{11}),
\bfpage{4083}
(\byear{2022}).
\doiurl{10.3390/s22114083}
\end{barticle}
\endbibitem

\bibitem{kappassov2020touch}
\begin{barticle}
\bauthor{\bsnm{Kappassov}, \binits{Z.}},
\bauthor{\bsnm{Corrales}, \binits{J.-A.}},
\bauthor{\bsnm{Perdereau}, \binits{V.}}:
\batitle{Touch driven controller and tactile features for physical interactions}.
\bjtitle{Robotics and Autonomous Systems}
\bvolume{123},
\bfpage{103332}
(\byear{2020}).
\doiurl{10.1016/j.robot.2019.103332}
\end{barticle}
\endbibitem

\bibitem{castano2021touch}
\begin{bchapter}
\bauthor{\bsnm{Casta{\~n}o~Amor{\'o}s}, \binits{J.}},
\bauthor{\bsnm{Gil}, \binits{P.}},
\bauthor{\bsnm{Puente~M{\'e}ndez}, \binits{S.T.}}:
\bctitle{Touch detection with low-cost visual-based sensor}.
In: \bbtitle{2nd Int. Conf. on Robotics, Computer Vision and Intelligent Systems (ROBOVIS)},
pp. \bfpage{136}--\blpage{142}
(\byear{2021}).
\doiurl{10.5220/0010699800003061}
\end{bchapter}
\endbibitem

\bibitem{simonyan2014very}
\begin{botherref}
\oauthor{\bsnm{Simonyan}, \binits{K.}},
\oauthor{\bsnm{Zisserman}, \binits{A.}}:
Very Deep Convolutional Networks for Large-Scale Image Recognition.
arXiv
(2014).
\doiurl{10.48550/arXiv.1409.1556}
\end{botherref}
\endbibitem

\bibitem{szegedy2016rethinking}
\begin{bchapter}
\bauthor{\bsnm{Szegedy}, \binits{C.}},
\bauthor{\bsnm{Vanhoucke}, \binits{V.}},
\bauthor{\bsnm{Ioffe}, \binits{S.}},
\bauthor{\bsnm{Shlens}, \binits{J.}},
\bauthor{\bsnm{Wojna}, \binits{Z.}}:
\bctitle{Rethinking the inception architecture}.
In: \bbtitle{Conf. on Computer Vision and Pattern Recognition (CVPR)}
(\byear{2016}).
\doiurl{10.1109/CVPR.2016.308}
\end{bchapter}
\endbibitem

\bibitem{sandler2018mobilenetv2}
\begin{bchapter}
\bauthor{\bsnm{Sandler}, \binits{M.}},
\bauthor{\bsnm{Howard}, \binits{A.}},
\bauthor{\bsnm{Zhu}, \binits{M.}},
\bauthor{\bsnm{Zhmoginov}, \binits{A.}},
\bauthor{\bsnm{Chen}, \binits{L.-C.}}:
\bctitle{Mobilenetv2: Inverted residuals and linear bottlenecks}.
In: \bbtitle{IEEE Conf. on Computer Vision and Pattern Recognition (CVPR)}
(\byear{2018}).
\doiurl{10.48550/arXiv.1801.04381}
\end{bchapter}
\endbibitem

\bibitem{everingham2010pascal}
\begin{barticle}
\bauthor{\bsnm{Everingham}, \binits{M.}},
\bauthor{\bsnm{Van~Gool}, \binits{L.}},
\bauthor{\bsnm{Williams}, \binits{C.K.}},
\bauthor{\bsnm{Winn}, \binits{J.}},
\bauthor{\bsnm{Zisserman}, \binits{A.}}:
\batitle{The pascal visual object classes (voc) challenge}.
\bjtitle{International Journal of Computer Vision}
\bvolume{88}(\bissue{2}),
\bfpage{303}--\blpage{338}
(\byear{2010}).
\doiurl{10.1007/s11263-009-0275-4}
\end{barticle}
\endbibitem

\bibitem{salton1983introduction}
\begin{bbook}
\bauthor{\bsnm{Salton}, \binits{G.}},
\bauthor{\bsnm{McGill}, \binits{M.J.}}:
\bbtitle{Introduction to Modern Information Retrieval}.
\bpublisher{McGraw-Hill, Inc.},
\blocation{USA}
(\byear{1986}).
\burl{https://dl.acm.org/doi/book/10.5555/576628}
\end{bbook}
\endbibitem

\bibitem{hossin2015review}
\begin{barticle}
\bauthor{\bsnm{Hossin}, \binits{M.}},
\bauthor{\bsnm{Sulaiman}, \binits{M.N.}}:
\batitle{A review on evaluation metrics for data classification evaluations}.
\bjtitle{International Journal of Data Mining \& knowledge management process}
\bvolume{5}(\bissue{2}),
\bfpage{1}
(\byear{2015}).
\doiurl{10.5121/ijdkp.2015.5201}
\end{barticle}
\endbibitem

\bibitem{5483185}
\begin{barticle}
\bauthor{\bsnm{Torralba}, \binits{A.}},
\bauthor{\bsnm{Russell}, \binits{B.C.}},
\bauthor{\bsnm{Yuen}, \binits{J.}}:
\batitle{Labelme: Online image annotation and applications}.
\bjtitle{Proceedings of the IEEE}
\bvolume{98}(\bissue{8}),
\bfpage{1467}--\blpage{1484}
(\byear{2010}).
\doiurl{10.1109/JPROC.2010.2050290}
\end{barticle}
\endbibitem

\bibitem{kingma2014adam}
\begin{bchapter}
\bauthor{\bsnm{Kingma}, \binits{D.P.}},
\bauthor{\bsnm{Ba}, \binits{J.}}:
\bctitle{Adam: A method for stochastic optimization}.
In: \bbtitle{3rd Int. Conf. for Learning Representations (ICLR)}
(\byear{2014}).
\doiurl{10.48550/arXiv.1412.6980}
\end{bchapter}
\endbibitem

\end{thebibliography}


\end{document}